\documentclass[twoside]{article}

\usepackage[accepted]{aistats2022}
\usepackage{preamble}
\usepackage{pifont}
\newcommand{\cmark}{\ding{51}}%
\newcommand{\xmark}{\ding{56}}%

\usepackage{hyperref}       
\usepackage{booktabs}       
\usepackage{tablefootnote}
\usepackage{threeparttable}
\usepackage{amsfonts}       
\usepackage{nicefrac}       
\usepackage{microtype}      
\usepackage{xcolor}         

\begin{document}

%
\runningtitle{Analysis of a Target-Based Actor-Critic Algorithm with Linear Function Approximation}

%

\twocolumn[

\aistatstitle{Analysis of a Target-Based Actor-Critic Algorithm\\ with Linear Function Approximation}

\aistatsauthor{ Anas Barakat \And Pascal Bianchi \And  Julien Lehmann}

\aistatsaddress{LTCI, Télécom Paris, Institut Polytechnique de Paris, France} ]

\begin{abstract}
  Actor-critic methods integrating target networks have exhibited a stupendous empirical success in deep reinforcement learning. However, a theoretical understanding of the use of target networks in actor-critic methods is largely missing in the literature.
  In this paper, we reduce this gap between theory and practice by proposing the first theoretical analysis of an online target-based actor-critic algorithm with linear function approximation in the discounted reward setting.
  Our algorithm uses three different timescales: one for the actor and two for the critic. Instead of using the standard single timescale temporal difference (TD) learning algorithm as a critic, we use a two timescales target-based version of TD learning closely inspired from practical actor-critic algorithms implementing target networks.
  First, we establish asymptotic convergence results for both the critic and the actor under Markovian sampling.
  Then, we provide a finite-time analysis showing the impact of incorporating a target network into actor-critic methods.
\end{abstract}

\section{INTRODUCTION}

Actor-critic algorithms \cite{barto-sutton-anderson83,konda-borkar99,konda-tsitsiklis03,peters-schaal08,bha-sut-gha-lee09} are a class of reinforcement learning (RL) \cite{sutton-barto18,bertsekas-tsitsiklis96} methods to find an optimal policy maximizing the total expected reward in a stochastic environment modelled by a Markov Decision Process (MDP) \cite{puterman14}. In this type of algorithms, two main processes interplay: the actor and the critic. The actor updates a parameterized policy in a direction of performance improvement whereas the critic estimates the current policy of the actor by estimating the unknown state-value function. In turn, the critic estimation is used to produce the update rule of the actor. Combined with deep neural networks as function approximators of the value function, actor-critic algorithms witnessed a tremendous success in a range of challenging tasks \cite{heess2015memory,lillicrap-et-al16,mnih-et-al16,fujimoto-et-al18,haarnoja-et-al18}. Apart from using neural networks for function approximation (FA),
one of the main features underlying their remarkable empirical achievements is the use of target networks for the critic estimation of the value function. Introduced by the seminal work of Mnih et al. \cite{mnih15} to stabilize the training process, this target innovation consists in using two neural networks maintaining two copies of the estimated value function: A so-called target network tracking a main network with some delay computes the target values for the value function update.

Despite their resounding empirical success in deep RL,
a theoretical understanding of the use of target networks in actor-critic methods is largely missing in the literature.
Theoretical contributions investigating the use of a target network are very recent and limited to temporal difference (TD) learning for policy evaluation \cite{lee-he19} and critic-only methods such as Q-learning for control \cite{zhang-yao-whiteson21}. In particular, these works are not concerned with actor-critic algorithms and leave the question of the finite-time analysis open.

In the present work, we reduce this gap between theory and practice by proposing the first theoretical analysis of an online target-based actor-critic algorithm in the discounted reward setting.
We consider the linear FA setting where a linear combination of pre-selected feature (or basis) functions estimates the value function in the critic. An analysis of this setting is an insightful first step before tackling the more challenging nonlinear FA setting
aligned with the use of neural networks. We conduct our study in the multiple timescales framework.
In the standard two timescales actor-critic algorithms \cite{konda-tsitsiklis03,bha-sut-gha-lee09}, at each iteration, the actor and the critic are updated simultaneously but the critic evolves faster than the actor which uses smaller stepsizes.
We face two main challenges due to the integration of the target variable mechanism. First, in contrast to standard two timescales actor-critic algorithms, our algorithm uses three different timescales: one for the actor and two for the critic. Instead of using the single timescale TD learning algorithm as a critic, we use a two timescales target-based version of TD learning closely inspired from practical actor-critic algorithms implementing target networks. Second, incorporating a target variable into the critic results in the intricate interplay between three processes evolving on three different timescales. In particular, the use of a target variable significantly modifies the dynamics of the actor-critic algorithm and deserves a careful analysis accordingly.

Our main contributions are summarized as follows. First, we prove asymptotic convergence results for both the critic and the actor. More precisely, as the actor parameter changes slowly compared to the critic one, we show that the critic using a target variable tracks a slowly moving target corresponding to a TD-like solution \cite{tsitsiklis-vanroy97}. Our development is based on the ordinary differential equation (ODE) method of stochastic approximation (see, for e.g., \cite{ben-met-pri90,borkar08}).
Then, we show that the actor parameter visits infinitely often a region of the parameter space where the norm of the policy gradient is dominated by a bias due to linear FA.
Second, we conduct a finite-time analysis of our actor-critic algorithm which shows the impact of using a target variable on the convergence rates and the sample complexity.
Loosely speaking, up to a FA error, we show that our target-based algorithm converges in expectation to an $\epsilon$-approximate stationary point of the non-concave performance function using at most~$\mO(\epsilon^{-3}\ln^3(\frac{1}{\epsilon}))$ samples compared with~$\mO(\epsilon^{-2}\ln(\frac{1}{\epsilon}))$ for the best known complexity for two timescales actor-critic algorithms without a target network.
All the proofs are deferred to the appendix.

\section{RELATED WORK}
\label{sec:related_work}

In this section, we briefly discuss the most relevant related works to ours.
Existing theoretical results in the literature can be divided into two classes.

\noindent\textbf{Asymptotic results.}
Almost sure convergence results are referred to as asymptotic.
Konda \& Tsitsiklis \cite{konda-tsitsiklis03,konda02thesis} provided almost sure (with probability one) convergence results for a two timescales actor-critic algorithm
in which the critic estimates the action-value function
via linear FA. Our algorithm is closer to an actor-critic algorithm introduced by Bhatnagar et al. \cite{bha-sut-gha-lee09} in the average reward setting.
However, unlike \cite{bha-sut-gha-lee09}, we integrate a target variable mechanism into our critic and consider the discounted reward setting.
Moreover, as previously mentioned, the target variable for the critic adds an additional timescale in comparison to \cite{konda-tsitsiklis03,bha-sut-gha-lee09} which only involve
two different timescales.
Regarding theoretical results considering target networks,
Lee \& He \cite{lee-he19} proposed a family of single timescale
target-based TD learning algorithms for policy evaluation.
Our critic corresponds to a two timescales version of the
single timescale target-based TD learning algorithm
of Lee \& He \cite[Algorithm~2]{lee-he19} called Averaging TD.
In \cite[Th.~1]{lee-he19}, this single timescale algorithm is shown to converge with probability one (w.p.1) towards the standard TD solution solving the projected Bellman equation
(see \cite{tsitsiklis-vanroy97} for a precise statement).
Besides the timescales difference with \cite{lee-he19}, in this article, we are concerned with a control setting in which the policy changes at each timestep via the actor update.
Yang et al.~\cite{yang-et-al19}
proposed a bilevel optimization perspective to analyze Q-learning with a target network and an actor-critic algorithm without any target network.
More recently, Zhang et al. \cite{zhang-yao-whiteson21} investigated the use of target networks in Q-learning with
linear FA and a target variable with Ridge regularization.
Their analysis covers the average and discounted reward settings and establishes asymptotic convergence results for policy evaluation and control.
This recent work \cite{zhang-yao-whiteson21} focuses on the critic-only Q-learning method with a target network update rule, showing the role of the target network in the off-policy setting. In particular, this work is not concerned with actor-critic algorithms.

\noindent\textbf{Finite-time analysis.}
  The second type of results consists in establishing time-dependent bounds on some error or performance quantities such as the average expected norm of the gradient of the performance function. These are referred to as finite-time analysis.
  In the last few years, several works proposed finite-time analysis for TD learning \cite{bhandari-russo-singal18,srikant-ying19} for two timescales
  TD methods \cite{xu-zou-lia19} and even more generally for two timescales
  linear stochastic approximation algorithms \cite{gupte-et-al19,dalal-et-al18,kal-mou-nau-tad-wai20}.
  These works opened the way to
  the recent development of a flurry of nonasymptotic results for actor-critic algorithms \cite{yan-zha-hon-bas19,qiu-et-al19,kum-kop-rib19,hon-wai-wan-yan20,xu-wan-lia20b,xu-et-al20,wang-et-al20,wu-zha-xu-gu20,shen-zhang-hong-chen20}.
  Regarding online one-step actor-critic algorithms,
  Wu et al. \cite{wu-zha-xu-gu20} provided a finite-time analysis of the standard
  two timescales actor-critic algorithm  \cite[Algorithm~1]{bha-sut-gha-lee09}
  in the average reward setting with linear FA. Shen et al. \cite{shen-zhang-hong-chen20}
  conducted a similar study for a revisited version of the asynchronous advantage actor-critic (A3C)
  algorithm in the discounted setting. None of the mentioned works uses a target network.
  In this work, we conduct a finite-time analysis of our target-based actor-critic algorithm. Such new results are missing in all theoretical results investigating the use of a target network \cite{lee-he19,zhang-yao-whiteson21}.

The summary table~\ref{table-related-works} compiles some key features of our work
to situate it in the literature and highlights our contributions with respect to (w.r.t.) the closest related works.
We also mention that alternative update rules are also possible for actor-critic algorithms.
Other common variants in practice use different policy gradients estimates based directly on the
critic estimate instead of using it for bootstrapping (see for e.g. a recent discussion in~\cite{wen-et-al21icml}).
Such a modification of the actor would not impact our critic analysis but would induce a different bias for the policy gradient estimate
(impacting namely Th.~\ref{th:actor} and Th.~\ref{th:actor_rate} below). Our analysis can also be adapted to this setting
with a suitable analysis of the induced bias.

\begin{table*}[h]
  \caption{Comparison to closest related works.}
  \label{table-related-works}
  \centering
  \begin{threeparttable}
  \begin{tabular}{lccccccc}
    \toprule
          & Discounted  & Actor  & Markovian & Target   & Asymptotic & Finite-time & Timescales \\
          & reward      & critic &  sampling\tnote{1} & variable & results & analysis   &  \\
    \midrule
    \cite{lillicrap-et-al16}     & \cmark  &   \cmark & \xmark & \cmark &  \xmark & \xmark & $1$\\
    \cite{lee-he19}              & \cmark  &   \xmark & \xmark & \cmark &  \cmark & \cmark\tnote{2} & $1$\\
    \cite{wu-zha-xu-gu20}        & \xmark  &   \cmark & \cmark & \xmark &  \xmark & \cmark & $3$\\
    \cite{shen-zhang-hong-chen20}& \cmark  &   \cmark & \cmark & \xmark &  \xmark & \cmark & $2$\\
    \cite{zhang-yao-whiteson21}  & \cmark  &  \xmark  & \cmark & \cmark &  \cmark & \xmark & $2$\\
    \textbf{This paper}                & \cmark & \cmark    & \cmark & \cmark & \cmark & \cmark  & $3$\\
    \bottomrule
  \end{tabular}
  \begin{tablenotes}\footnotesize
  \item[1] refers to the use of samples generated from the MDP and the acting policy, this excludes experience replay as in~\cite{lillicrap-et-al16} and identically independently distributed (i.i.d.) samples used in theoretical analysis.
  \item[2] \cite{lee-he19} provide a finite-time analysis for a target-based TD-learning algorithm (for policy evaluation) based on the periodic update style of the target variable used in~\cite{mnih15} involving two loops. They highlight that a finite-time analysis of the Polyak-averaging style update rule \cite{lillicrap-et-al16} is an open question. Here, we address this question in the control setting.
  \end{tablenotes}
  \end{threeparttable}
\end{table*}

\section{PRELIMINARIES}

\noindent\textbf{Notation.}
For every finite set $\mX$, we use the notation~$\mP(\mX)$ for the set of probability measures on~$\mX$.
The cardinality of a finite set $\mY$ is denoted by $|\mY|$.
For two sequences of nonnegative reals $(x_n)$ and $(y_n)$, the notation $x_n = \mO(y_n)$ means that there exists
a constant~$C$ independent of~$n$ such that $x_n \leq C y_n$ for all~$n \in \bN$\,.
For any integer $p$, the euclidean space $\bR^p$ is equipped with its usual inner product $\ps{\cdot,\cdot}$ and its corresponding $2$-norm $\|\cdot\|$.
For any integer $d$ and any matrix~$A \in \bR^{d \times p}$, we use the notation~$\|A\|$ for the operator norm induced by the euclidean vector norm.
For a symmetric positive semidefinite matrix $B \in \bR^{p \times p}$ and a vector $x \in \bR^p$, the notation $\|x\|_B^2$ refers to the quantity $\ps{x, Bx}$\,.
The transpose of the vector $x$ is denoted by~$x^T$ and~$I_p$ is the identity matrix.

\subsection{Markov decision process and problem formulation}

Consider the RL setting \cite{sutton-barto18,bertsekas-tsitsiklis96,szepesvari10} where a learning agent interacts with an environment modeled as an infinite horizon discrete-time discounted MDP. We denote by $\mS = \{s_1, \cdots, s_n\}$ the finite set of states and $\mA$ the finite set of actions.
Let $p: \mS \times \mA \to \mP(\mS)$ be the state transition probability kernel
and $R : \mS \times \mA \to \bR$ the immediate reward function.
A randomized stationary policy, which we will simply call a policy in the rest of the paper, is a mapping $\pi: \mS \to \mP(\mA)$ specifying for each $s \in \mS, a \in \mA$ the probability $\pi(a|s)$ of selecting action $a$ in state $s$.
At each time step $t \in \bN$, the RL agent in a state $S_t \in \mS$ executes an action $A_t \in \mA$ with probability~$\pi(A_t|S_t)$, transitions into a state $S_{t+1} \in \mS$ with probability $p(S_{t+1}|S_t,A_t)$ and observes a random reward $R_{t+1}\in [-U_R,U_R]$ where $U_R$ is a positive real. We denote by $\bP_{\rho,\pi}$ the probability distribution of the Markov chain~$(S_t,A_t)$ issued from the MDP controlled by the policy~$\pi$ with initial state distribution~$\rho$. The notation~$\bE_{\rho,\pi}$ refers to the associated expectation. We will use~$\bE_{\pi}$ whenever there is no dependence on~$\rho$.
The sequence~$(R_t)$ is
such that (s.t.) $\bE_{\pi}[R_{t+1} | S_t, A_t] = R(S_t,A_t)$\,.
Let $\gamma \in (0,1)$ be a discount factor. Given a policy $\pi$, the long-term expected cumulative discounted reward is quantified by the state-value function $V_\pi : \mS \to \bR$ and the action-value function $Q_\pi : \mS \times \mA \to \bR$ defined for all $s\in \mS, a \in \mA$ by
$
V_\pi(s) \eqdef \bE_\pi [ \sum_{t=0}^\infty \gamma^t R_{t+1}| S_0 = s]
$
and
$
Q_\pi(s,a) \eqdef \bE_\pi [ \sum_{t=0}^\infty \gamma^t R_{t+1} | S_0 = s, A_0= a ]\,.
$
We also define the advantage function $\Delta_\pi : \mS \times \mA \to \bR$ by $\Delta_\pi(s,a) \eqdef Q_\pi(s,a) - V_\pi(s)$.
Given an initial probability distribution $\rho$ over states for the initial state $S_0$, the goal of the agent is to find a policy $\pi$
maximizing the expected long-term return
$
J(\pi) \eqdef \sum_{s\in \mS} \rho(s) V_\pi(s)\,.
$
For this purpose, the agent has only access to realizations of the
random variables $S_t, A_t$ and $R_t$ whereas the state transition kernel~$p$
and the reward function~$R$ are unknown.

\subsection{Policy Gradient framework}

From now on, we restrict the policy search to the set of policies~$\pi$ parameterized by a vector~$\theta \in \bR^d$ for some integer $d > 0$ and optimize the performance criterion $J$ over this family of parameterized policies $\{ \pi_\theta : \theta \in \bR^d\}$. The policy dependent function $J$ can also be seen as a function of the parameter $\theta$. We use the notation~$J(\theta)$ for~$J(\pi_\theta)$ by abuse of notation. The problem that we are concerned with can be written as:
$
      \max_{\theta \in \bR^d} J(\theta)\,.
$
Whenever it exists, define for every~$\theta \in \bR^d$ the function $\psi_\theta : \mS \times \mA \to \bR^d$  for all $(s,a) \in \mS \times \mA$ by:
\[\psi_\theta(s,a) \eqdef \nabla \ln \pi_\theta(a|s)\,,
\]
where~$\nabla$ denotes the gradient w.r.t. $\theta$.
We introduce an
assumption on the regularity of the parameterized family
of policies which is a standard requirement in policy gradients  (see, for eg., \cite[Assumption~3.1]{zhang-koppel-zhu-basar20}\cite[Assumption~2.1]{konda-tsitsiklis03}).
In particular, it ensures that~$\psi_\theta$ is well defined\,. 
\begin{assumption}
 \label{hyp:grad_pi}
 The following conditions hold true for every~$(s,a) \in \mS \times \mA$\,.
 \begin{enumerate}[{\sl (a)},leftmargin=*,noitemsep,topsep=0pt]
 \item For every $\theta \in \bR^d$, $\pi_\theta(a|s) > 0$\,.
 \item \label{hyp:pi_theta} The function $\theta \mapsto \pi_\theta(a|s)$ is continuously differentiable and $L_\pi$-Lipschitz continuous.
 \item \label{hyp:psi_theta} The function $\theta \mapsto \psi_\theta(s,a)$ is bounded and $L_\psi$-Lipschitz.
 \end{enumerate}
\end{assumption}
Assumption~\ref{hyp:grad_pi} is satisfied for instance by the Gibbs (or softmax) policy and the Gaussian policy (see \cite[Sec.~3]{zhang-koppel-zhu-basar20} and the references therein for details).
Under Assumption~\ref{hyp:grad_pi}\,, the policy gradient theorem \cite{sutton-et-al00}\cite[Th.~2.13]{konda02thesis} with the state-value function as a baseline provides an expression for the gradient of the performance metric~$J$ w.r.t. the policy parameter~$\theta$ given by:
\begin{equation}
   \label{eq:pg}
   \nabla J(\theta) = \frac{1}{1-\gamma} \cdot \bE_{(\tilde{S},\tilde{A}) \,\sim\, \mu_{\rho,\theta}}[\Delta_{\pi_\theta}(\tilde{S},\tilde{A})\, \psi_\theta(\tilde{S},\tilde{A})]\,.
\end{equation}
Here, the couple of random variables~$(\tilde{S},\tilde{A})$ follows the discounted state-action occupancy measure~$\mu_{\rho,\theta} \in \mP(\mS,\mA)$ defined for all $(s,a) \in \mS \times \mA$ by:
\begin{align}
\label{eq:def_d_rho_theta}
\mu_{\rho,\theta}(s,a) &\eqdef d_{\rho,\theta}(s)\, \pi_\theta(a|s)\\
\text{where} \quad
 d_{\rho,\theta}(s) &\eqdef (1-\gamma) \sum_{t=0}^\infty \gamma^t \bP_{\rho,\pi_\theta}(S_t = s)\,
\end{align}
is a probability measure over the state space $\mS$ known as the discounted state-occupancy measure.
Note that under Assumption~\ref{hyp:grad_pi}\,, the policy gradient $\nabla J$ is Lipschitz continuous (see~\cite[Lem.~4.2]{zhang-koppel-zhu-basar20}).

\section{TARGET-BASED ACTOR-CRITIC ALGORITHM}
\label{sec:algo}

In this section, we gradually present our actor-critic algorithm.

\subsection{Actor update}
First, we need an estimate of the policy gradient $\nabla J(\theta)$ of Eq.~\eqref{eq:pg} in view of using stochastic gradient ascent to solve the maximization problem\,.
Given Eq.~\eqref{eq:pg} and following previous works, we recall
how to sample according to the distribution~$\mu_{\rho,\theta}$.
As described in \cite[Sec.~2.4]{konda02thesis}, the distribution $\mu_{\rho,\theta}$ is the stationary distribution of a Markov chain $(\tilde{S}_t,\tilde{A}_t)_{t \in \bN}$
issued from the artificial MDP whose transition kernel $\tilde p : \mS \times \mA \to \mP(\mS)$ is defined for every $(s,a) \in \mS \times \mA$ by
\begin{equation}
\label{eq:tilde_p_def}
\tilde p (\cdot |s, a) \eqdef \gamma \, p(\cdot|s, a) + (1-\gamma)\,\rho(\cdot)\,,
\end{equation}
and which is controlled by the policy~$\pi_\theta$
generating the action sequence~$(\tilde{A}_t)$\,.
We will later state conditions to ensure its existence and uniqueness. Therefore, under suitable conditions, the distribution of the Markov chain~$(\tilde{S}_t,\tilde{A}_t)_{t \in \bN}$ will converge geometrically towards its stationary distribution~$\mu_{\rho,\theta}$. This justifies the following sampling procedure. Given a state $\tilde{S}_t$ and an action~$\tilde{A}_t$, we sample a state~$\tilde{S}_{t+1}$ according to this artificial MDP by sampling from~$p(\cdot|\tilde{S}_t,\tilde{A}_t)$ with probability~$\gamma$
and from~$\rho$ otherwise. For this purpose, at each time step~$t$, we draw a Bernoulli
random variable~$B_t \in \{0,1\}$ with parameter~$\gamma$
which is independent of all the past random variables generated until time~$t$.

Then, using the definition of the advantage function, Eq.~\eqref{eq:pg} becomes:
\begin{equation}
   \label{eq:pg_2}
   \nabla J(\theta) = \frac{1}{1-\gamma} \cdot \bE
   [(R(\tilde{S},\tilde{A}) + \gamma V_{\pi_\theta}(S) - V_{\pi_\theta}(\tilde{S}))
   \, \psi_\theta(\tilde{S},\tilde{A})]\,,
\end{equation}
where $(\tilde{S},\tilde{A})\sim\mu_{\rho,\theta}$ and $S \sim p(\cdot|\tilde{S},\tilde{A})$\,.
From this equation, it is natural to define for every~$V \in \bR^n$ the temporal difference (TD) error
\begin{equation}
   \label{eq:tderror}
   \delta_{t+1}^V = R_{t+1} + \gamma\, V(S_{t+1}) - V(\tilde{S}_t)\,,
\end{equation}
where $S_{t+1}$ is drawn from the distribution $p(\cdot|\tilde{S}_t,\tilde{A}_t)$
and~$(\tilde{S}_t,\tilde{A}_t)_{t \in \bN}$ is the Markov chain induced by the artificial MDP 
described in Eq.~\eqref{eq:tilde_p_def} and controlled by the policy~$\pi_\theta$.
Notice here from Eq.~\ref{eq:pg_2} that we need two different sequences~$(S_t)$ and~$(\tilde{S}_t)$ respectively sampled from the kernels~$p$ and~$\tilde{p}$. In our discounted reward setting, using only the sequence~$(\tilde{S}_t)$ issued from the artificial kernel $\tilde{p}$ would result in a bias with a sampling error of the order $1-\gamma$
(see \cite[Eq.~(14) and Lem.~7]{shen-zhang-hong-chen20}).

Supposing for now that the value function $V_{\pi_\theta}$ is known, it stems from Eq.~\eqref{eq:pg_2} that a natural estimator of the gradient~$\nabla J(\theta)$ is~$\delta_{t+1}^{V_{\pi_\theta}} \psi_{\theta}(\tilde{S}_t,\tilde{A}_t)/(1-\gamma)$. This estimator is only biased because the distribution of our sampled Markov chain~$(\tilde{S}_t,\tilde{A}_t)_t$ is not exactly $\mu_{\rho,\theta}$ but converges geometrically to this one.
However, the state-value function $V_{\pi_\theta}$ is unknown.
Given an estimate~$V_{\omega_t} \in \bR^n$ of $V_{\pi_{\theta_t}}$ and a positive stepsize~$\alpha_t$, the actor updates its parameter as follows:
\begin{equation}
   \label{eq:actor}
   \theta_{t+1} = \theta_t + \alpha_t \frac{1}{1-\gamma}\delta_{t+1}^{V_{\omega_t}} \psi_{\theta_t}(\tilde{S}_t,\tilde{A}_t)\,.
\end{equation}

\subsection{Critic update}
\label{subsec:critic_update}
The state-value function $V_{\pi_\theta}$ is approximated
for every state~$s \in \mS$ by a linear function of carefully chosen feature vectors as follows:
$
V_{\pi_\theta}(s) \approx V_\omega(s) = \omega^T \phi(s) = \sum_{i=1}^m \omega_i \phi^i(s)\,,
$
where $\omega = (\omega_1, \cdots, \omega_m)^T \in \bR^m$
for some integer $m \ll n = |\mS|$
and $\phi(s) = (\phi^1(s), \cdots, \phi^m(s))^T$
is the feature vector of the state~$s \in \mS$.
We compactly represent the feature vectors as a matrix of features~$\Phi$ of size $n \times m$
 whose $i$th row corresponds to the row vector~$\phi(s)^T$ for some $s \in \mS$\,.

Now, before completing the presentation of our algorithm, we motivate the use of a target variable for the critic.
As previously mentioned, instead of a standard TD learning algorithm \cite{sutton88}
for the critic, we use a target-based TD learning algorithm.
We follow a similar exposition to \cite[Secs.~2.3, 2.4 and~3]{lee-he19} to introduce
the target variable for the critic. Let us introduce some additional
notations for this purpose.
Fix $\theta \in \bR^d$. Let $P_\theta$ be the transition matrix over the finite state space associated
to the Markov chain~$(S_t)$, i.e., the matrix of size $n \times n$ defined for every $s,s' \in \mS$ by
$
P_\theta(s'|s) \eqdef \sum_{a \in \mA} p(s'|s,a)\pi_\theta(a|s)\,.
$
Consider the vector $R_\theta = (R_\theta(s_1), \cdots R_\theta(s_n))$ whose $i$th coordinate is provided by
$R_\theta(s_i) = \sum_{a \in \mA} \pi_\theta(a|s_i) R(s_i,a)$\,.
Let $D_{\rho,\theta}$ be the diagonal matrix with elements $d_{\rho,\theta}(s_i), \, i= 1, \cdots n$
along its diagonal. Define also the Bellman operator $T_\theta : \bR^n \mapsto \bR^n$ for every $V \in \bR^n$ by
$T_\theta V \eqdef R_\theta + \gamma P_\theta V$\,. The true value function~$V_{\pi_\theta}$
satisfies the celebrated Bellman equation $V_{\pi_\theta} = T_\theta V_{\pi_\theta}$\,.
This naturally leads to minimize the mean-square Bellman
error (MSBE) \cite[Sec.~3]{sutton-et-al09a} defined for every~$\omega \in \bR^m$ by
$
\mE_{\theta}(\omega) \eqdef \frac 12 \|T_\theta V_\omega - V_\omega\|_{D_{\rho,\theta}}^2\,
$
where $V_\omega = \Phi \omega$\,.
The gradient of the MSBE w.r.t. $\omega$ can be written
as $\nabla_\omega \mE_\theta(\omega) = \bE_{\tilde{S} \sim d_{\rho,\theta}}
[(T_\theta V_\omega(\tilde{S}) - V_\omega(\tilde{S}))
(\bE_{S \sim P_\theta(\cdot|\tilde{S})}[\gamma \nabla_\omega V_\omega(S)]
-\nabla_\omega V_\omega(\tilde{S}))]$\,. As explained in \cite[p.~369]{bertsekas-tsitsiklis96},
omitting the gradient term~$\nabla_\omega T_\theta V_\omega(\tilde{S}) = \bE_{S \sim P_\theta(\cdot|\tilde{S})}[\gamma \nabla_\omega V_\omega(S)]$
in $\nabla_\omega \mE_\theta(\omega)$ yields the standard TD learning update rule
$\omega_{t+1} = \omega_t + \delta_{t+1} \phi(\tilde{S}_t)$.
The TD learning update does not coincide with a stochastic gradient descent on the MSBE
or even any other objective function (see \cite[Appendix~1]{barnard93} for a proof).
The idea of target-based TD learning is to consider a modified version of the MSBE
$\tilde{\mE}_\theta(\omega,\bar{\omega}) \eqdef \frac 12 \|T_\theta V_{\bar{\omega}} - V_\omega\|_{D_{\rho,\theta}}^2$\,.
Observe that the term~$T_\theta V_\omega$ depending on~$\omega$ in the MSBE is now
freezed in~$\tilde{\mE}_\theta(\omega,\bar{\omega})$ thanks to the target variable~$\bar{\omega}$\,.
We now need to introduce a new sequence $\bar{\omega}_t$
to define a sample-based version of $T_\theta V_{\bar{\omega}} - V_\omega$ which
will be a modified version of the standard TD-error
\begin{equation}
   \label{eq:tderror_bar}
   \bar{\delta}_{t+1} = R_{t+1} + \gamma \phi(S_{t+1})^T \bar{\omega}_t - \phi(\tilde{S}_t)^T \omega_t\,.
\end{equation}
Then, a stochastic gradient descent on~$\tilde{\mE}$ w.r.t.~$\omega$ yields the critic update
\begin{equation}
   \label{eq:critic}
   \omega_{t+1} = \omega_t + \beta_t \bar{\delta}_{t+1} \phi(\tilde{S}_t)\,.
\end{equation}
The target variable sequence $\bar{\omega}_t$ needs to be a slowed down version
of the critic parameter~$\omega_t$. For this purpose, instead of
using a periodical synchronization of the target variable~$\bar{\omega}_t$
with~$\omega_t$ through a copy as in DQN, we use
the Polyak-averaging update rule proposed by \cite{lillicrap-et-al16}
\begin{equation}
   \label{eq:critic-bar}
   \bar{\omega}_{t+1} = \bar{\omega}_t + \xi_t (\omega_{t+1} - \bar{\omega}_t)\,,
\end{equation}
where $\xi_t$ is a positive stepsize chosen s.t. the sequence~$(\bar{\omega}_t)$
evolves on a slower timescale than the sequence~$(\omega_t)$ to track it.
The update rules of the actor and the critic collected together
from Eqs.~\eqref{eq:tderror} to~\eqref{eq:critic} give rise
to Algorithm~\ref{algo}. We will use the shorthand notation~$\delta_{t+1} \eqdef \delta_{t+1}^{V_{\omega_t}}$ from now on.
\begin{algorithm}
   \caption{Target-based actor-critic.}
   \label{algo}
\begin{algorithmic}
   \State {\bfseries Initialization:} $\theta_0 \in \bR^d, \omega_0 \in \bR^m$\,.
   \For{$t=0, 1, 2, \cdots, T-1$}
   \State $\tilde{A}_t \sim \pi_{\theta_t}(\cdot|\tilde{S}_t); \, S_{t+1} \sim p(\cdot|\tilde{S}_t, \tilde{A}_t)$
   \State $\delta_{t+1} = R_{t+1} + \gamma\, \phi(S_{t+1})^T \omega_t - \phi(\tilde{S}_t)^T \omega_t$ \\\Comment{classical TD error}
   \State $\bar{\delta}_{t+1} = R_{t+1} + \gamma\, \phi(S_{t+1})^T \bar{\omega}_t - \phi(\tilde{S}_t)^T \omega_t$ \\\Comment{target-based TD error}
   \State $\theta_{t+1} = \theta_t + \alpha_t \frac{1}{1-\gamma} \delta_{t+1}\psi_{\theta_t}(\tilde{S}_t,\tilde{A}_t)$ \Comment{actor}
   \State $\omega_{t+1} = \omega_t + \beta_t \bar{\delta}_{t+1} \phi(\tilde{S}_t)$ \Comment{critic}
   \State $\bar{\omega}_{t+1} = \bar{\omega}_t + \xi_t (\omega_{t+1} - \bar{\omega}_t)$ \Comment{target variable}
   \State $S_{t+1}^{\rho} \sim \rho\,;\,\, B_{t+1} \sim \mathcal{B}(\gamma)$
   \State $\tilde{S}_{t+1} = B_{t+1} S_{t+1} + (1-B_{t+1})S_{t+1}^{\rho}$
   \EndFor
   \State {\bfseries Output:} Policy and value function parameters $\theta_T$ and $\omega_T$.
\end{algorithmic}
\end{algorithm}

\begin{remark}
We can simplify Algorithm~\ref{algo} by using only the target-based TD error~$\bar{\delta}_{t+1}$ instead of maintaining both TD errors~$\bar{\delta}_{t+1}$ and~$\delta_{t+1}$.
The proofs can be easily adapted, note for this that~$(\bar{\omega}_t)$ and~$(\omega_t)$ track the same target~$\bar{\omega}_*(\theta_t)$ (see Prop.~\ref{prop:critic2}, Th.~\ref{th:critic}). For clarity of exposition, we present the algorithm with both TD errors, since the classical TD error stems directly from the policy gradient whereas the target-based TD error comes from the use of the target network.
\end{remark}

\section{CONVERGENCE ANALYSIS}
\label{sec:asymptotic_cv}

In this section, we provide asymptotic convergence
guarantees for the critic and the actor of Algorithm~\ref{algo}
successively\,.
For every~$\theta \in \bR^d$, let~$\tilde{K}_\theta \in  \bR^{|\mS||\mA|  \times |\mS||\mA|}$ be the transition matrix over the state-action pairs defined for every $(s,a), (s^\prime,a^\prime) \in \mS \times \mA$ by $\tilde{K}_\theta(s^\prime,a^\prime|s,a) =  \tilde{p}(s^\prime|s,a)\pi_\theta(a^\prime|s^\prime)$\,.
Let $\mathcal{K} \eqdef \{\tilde{K}_\theta : \theta \in \bR^d \}$\, and let $\bar{\mathcal{K}}$ be its closure. Every element of~$\bar{\mathcal{K}}$ defines a Markov chain on the state-action space.
We make the following assumption (see also \cite{zhang-yao-whiteson21,marbach-tsitsiklis01}).
\begin{assumption}
\label{hyp:markov_chain}
For every~$K \in \bar{\mathcal{K}}$, the Markov chain induced by $K$ is ergodic.
\end{assumption}
In particular, it ensures the existence of a unique invariant distribution $\mu_{\rho,\theta}$ for the kernel~$\tilde{K}_\theta$ for every~$\theta \in \bR^d$.
Note that we can replace $\tilde{p}$ by $p$ in Assumption~\ref{hyp:markov_chain}\,.

Algorithm~\ref{algo} involves three different timescales.
The actor parameter~$\theta_t$ is updated on a slower timescale
(i.e., with smaller stepsizes) than the target variable~$\bar{\omega}_t$
which itself uses smaller stepsizes than the main critic parameter~$\omega_t$.
This is guaranteed by a specific choice of the three stepsize schedules. The following assumption is a three timescales version of the standard assumption used for two timescales stochastic approximation \cite[Chap.~6]{borkar08} and plays a pivotal role in our analysis.

\begin{assumption}[stepsizes]
   \label{hyp:stepsizes}
   The sequences of positive stepsizes $(\alpha_t), (\beta_t)$ and~$(\xi_t)$ satisfy:
   \begin{enumerate}[{\sl (a)},leftmargin=*,noitemsep,topsep=0pt]
   \itemsep0em
   \item $\sum_{t} \alpha_t = \sum_{t} \beta_t = \sum_{t} \xi_t = +\infty$\,,
   \item $\sum_{t} (\alpha_t^2 + \beta_t^2 + \xi_t^2) < \infty$\,,
   \item $\lim_{t \to \infty} \alpha_t/\xi_t  = \lim_{t \to \infty} \xi_t/\beta_t = 0$\,.
   \end{enumerate}
\end{assumption}

We also need the following stability assumption.
\begin{assumption}
\label{hyp:stability}
$\sup_t (\|\omega_t\| +  \|\theta_t\|) < +\infty \,\, w.p.1$\,. 
\end{assumption}
The almost sure boundedness assumption is classical \cite{konda-borkar99,borkar08,bha-sut-gha-lee09,karmakar-bhatnagar18}. The stability question could be addressed in a look up table representation setting (for e.g., $m=n$).
Nevertheless, this question seems out of reach in the FA setting without any modification of the algorithm.
Indeed, as discussed in~\cite[p.~2478-2479]{bha-sut-gha-lee09},
FA makes it hard to find a Lyapunov function to apply the stochastic Lyapunov function method \cite{kus-yin-(livre)03} whereas the function~$J$ can be readily used in the tabular case. Under a modification of the actor update of the algorithm and slightly stronger assumptions inspired from \cite{konda-tsitsiklis03slowMC,konda-tsitsiklis03}, the almost sure boundedness of the sequence~$(\omega_t)$ can be relaxed using a generalization to three timescales of the rescaling technique of~\cite{borkar-meyn00} which was extended by~\cite{lakshminarayanan-bhatnagar17} to two timescales stochastic approximation in the case of i.i.d. samples. For simplicity of exposition, we defer the technical details regarding this question to the appendix (see Appendix~\ref{appendix:stab}).
Concerning the sequence~$(\theta_t)$, as previously mentioned, it seems out of reach without modifying the algorithm, \cite{lakshminarayanan-bhatnagar17} (see their Section~6) propose for example to regularize the objective function~$J$ by adding a quadratic penalty~$\epsilon\, \|\theta\|^2/2$ ($\epsilon$ positive) leading to an additional~$\epsilon\, \theta_t$ term in the actor update of the standard actor-critic algorithm~1 of \cite{bha-sut-gha-lee09}. We do not make use of this trick which modifies the critical points of the performance function.
It is also worth mentioning that several works
enforce the boundedness via a projection of the iterates
on some compact set
\cite{bhandari-russo-singal18,wu-zha-xu-gu20,shen-zhang-hong-chen20,zhang-yao-whiteson21}. The drawback of this procedure is that it modifies the dynamics of the iterates and could possibly introduce spurious equilibria.

First, we will analyze the critic before investigating the convergence properties of the actor.

\subsection{Critic analysis}
\label{subsec:critic_asympt}

The following assumption regarding the family of basis functions
is a standard requirement \cite{bha-sut-gha-lee09,konda-tsitsiklis03,tsitsiklis-vanroy97}.

\begin{assumption}[critic features]
\label{hyp:features}
The matrix $\Phi$ has full column rank.
\end{assumption}

We follow the strategy of \cite[Chap.~6, Lem.~1]{borkar08} for the analysis of
multi-timescale stochastic approximation schemes based on the
ODE method.
We start by analyzing the sequence~$(\omega_t)$ evolving on the fastest timescale, i.e., with the slowly vanishing
stepsizes $\beta_t$ (see Assumption~\ref{hyp:stepsizes}).
The main idea behind the proofs is that~$\theta_t,\bar{\omega_t}$ can be considered as quasi-static in this timescale.
Then, loosely speaking (see Appendix for a rigorous statement and proof),
we can show from its update rule Eq.~\eqref{eq:critic} that~$(\omega_t)$ is associated to the ODE
\begin{equation}
  \label{eq:ode-mu}
  \begin{cases}
    \dot{\omega}(s)&= \bar{h}(\theta(s),\bar{\omega}(s)) - \bar{G}(\theta(s))\,\omega(s)\,, \\
    \dot{\theta}(s)&=  0\,,\\
    \dot{\bar{\omega}}(s) &= 0\,,
  \end{cases}
  \tag{ODE-$\omega$}
\end{equation}
where $\bar{h}: \bR^d \times \bR^m \to \bR^m$
and $\bar{G}: \bR^d \to \bR^{m\times m}$ are defined
for every $\theta \in \bR^d, \bar{\omega} \in \bR^m$ by
\begin{equation}
  \label{eq:barh_barG}
\bar{h}(\theta, \bar{\omega}) \eqdef \Phi^T D_{\rho, \theta}  (R_\theta + \gamma P_\theta\Phi\, \bar{\omega})\,
\text{and}\,
\bar{G}(\theta) \eqdef \Phi^T D_{\rho, \theta} \Phi\,.
\end{equation}
Recall that the matrices~$D_{\rho,\theta}, P_\theta$ and the vector~$R_\theta$ are defined in Sec.~\ref{subsec:critic_update}.
\begin{remark}
Under Assumptions~\ref{hyp:markov_chain} and~\ref{hyp:features}\,, the matrix $-\bar{G}(\theta)$ is
Hurwitz for every $\theta \in \bR^d$\,, i.e., all its eigenvalues
have negative real parts. In particular, it is invertible.
\end{remark}

The matrix $-\bar{G}(\theta)$ being Hurwitz,
it follows from~\eqref{eq:ode-mu} that $\omega_t$ tracks a slowly moving target $\omega_*(\theta_t,\bar{\omega}_t)$ governed by
the slower iterates $\theta_t$ and $\bar{\omega}_t$.
The detailed proof in the appendix makes use of a result from \cite{karmakar-bhatnagar18} to handle the Markovian noise.
\begin{proposition}
\label{prop:critic1}
Under Assumptions~\ref{hyp:grad_pi} and \ref{hyp:markov_chain} to~\ref{hyp:features}\,,
the linear equation~$\bar{G}(\theta)\omega = \bar{h}(\theta, \bar{\omega})$ has
a unique solution $\omega_*(\theta,\bar{\omega})$ for every~$\theta \in \bR^d, \bar{\omega} \in \bR^m$
and $\lim_t \|\omega_t - \omega_*(\theta_t,\bar{\omega}_t)\| = 0\,\, w.p.1.\,$
\end{proposition}

In a second step, we analyze the target variable sequence $(\bar{\omega}_t)$ which is evolving on
a faster timescale than the sequence~$(\theta_t)$ and slower than the sequence~$(\omega_t)$. At the timescale $\xi_t$, everything happens as if the quantity~$\omega_t$ in Eq.~\eqref{eq:critic-bar} could be replaced by $\omega_*(\theta_t,\bar{\omega}_t)$ thanks to Prop.~\ref{prop:critic1}.
Thus, in a sense that is made precise in the appendix,
we can show from Eq.~\eqref{eq:critic-bar} that~$(\bar{\omega}_t)$ is related to the ODE
\begin{equation}
\label{eq:ode-barmu}
  \begin{cases}
    \dot{\bar{\omega}}(s)&= \bar{G}(\theta(s))^{-1} (h(\theta(s)) - G(\theta(s))\bar{\omega}(s))\,, \\
    \dot{\theta}(s)&=  0\,,
  \end{cases}
  \tag{ODE-$\bar{\omega}$}
\end{equation}
where $h: \bR^d \to \bR^n$ and $G: \bR^d \to \bR^{m\times m}$ are defined
for every $\theta \in \bR^d$ by
\begin{equation}
  \label{eq:h_G}
h(\theta) \eqdef \Phi^T D_{\rho, \theta} R_\theta\,\quad \text{and} \quad  
G(\theta) \eqdef \Phi^T D_{\rho, \theta} (I_n - \gamma P_\theta) \Phi\,.
\end{equation}

We show in the appendix that the matrix $-G(\theta)$
is Hurwitz. This result differs from \cite[Lem.~6.6. p.300]{bertsekas-tsitsiklis96} or \cite[Lem.~9]{tsitsiklis-vanroy97} because the matrix~$D_{\rho,\theta}$ corresponds to the stationary distribution associated to the artificial kernel~$\tilde{p}$ and the policy $\pi_\theta$ in lieu of the original transition kernel~$p$. Then, we prove that $-\bar{G}(\theta)^{-1}G(\theta)$ is also stable, which suggests from~\eqref{eq:ode-barmu} that $\bar{\omega}_t$ tracks an other slowly moving target $\bar{\omega}_*(\theta_t)$.
This is established in the next proposition.

\begin{proposition}
\label{prop:critic2}
Under Assumptions~\ref{hyp:grad_pi}\, and \ref{hyp:markov_chain} 
to~\ref{hyp:features}\,, for every $\theta \in \bR^d$,
the linear equation $G(\theta)\bar{\omega}= h(\theta)$
has a unique solution $\bar{\omega}_*(\theta)$ and
$\lim_t \|\bar{\omega}_t - \bar{\omega}_*(\theta_t)\| = 0 \,\, w.p.1\,.$
Moreover, for every~$\theta \in \bR^d$, $\Phi\,\bar{\omega}_*(\theta)$ is a fixed point of the
projected Bellman operator, i.e., $\Pi_{\theta} T_{\theta} (\Phi\,\bar{\omega}_*(\theta)) = \Phi\,\bar{\omega}_*(\theta)$,
where $\Pi_\theta = \Phi (\Phi^T D_{\rho,\theta} \Phi)^{-1} \Phi^T D_{\rho,\theta}$
is the projection matrix on the space $\{\Phi\, \omega \,:\, \omega \in \bR^m  \}$ of
all vectors of the form $\Phi\, \omega$ for $\omega \in \bR^m$
w.r.t. the norm $\|\cdot\|_{D_{\rho,\theta}}$.
\end{proposition}

Combining the results from Props.~\ref{prop:critic1} and~\ref{prop:critic2}\,, we prove
that~$\omega_t$ tracks the same target $\bar{\omega}_*(\theta_t)$.

\begin{theorem}
\label{th:critic}
Let Assumptions~\ref{hyp:grad_pi},\,and \ref{hyp:markov_chain}
to~\ref{hyp:features} hold true\,. Then, we have
\[
\lim_t \|\omega_t - \bar{\omega}_*(\theta_t)\| = 0 \,\, w.p.1\,.
\]
Moreover, this limit implies the following:
$
\lim_t \| \Pi_{\theta_t} T_{\theta_t}(\Phi\, \omega_t) - \Phi\, \omega_t\| = 0 \quad w.p.1\,.
$
\end{theorem}

\begin{remark}
When the actor parameter $\theta_t$ is fixed (i.e., we are back to a policy evaluation problem),
the second part of the above convergence result coincides with the widely known
interpretation of the limit of the TD learning algorithm provided in
\cite{tsitsiklis-vanroy97} (see also \cite[p. 303-304]{bertsekas-tsitsiklis96}).
\end{remark}

\subsection{Actor analysis}

\begin{theorem}
\label{th:actor}
Let Assumptions~\ref{hyp:grad_pi} and~\ref{hyp:markov_chain} 
to~\ref{hyp:features} hold true. Then, w.p.1
\[
\liminf_t \left(\|\nabla J(\theta_t)\| - \|b(\theta_t)\| \right) \leq 0\,,
\]
where for every $\theta \in \bR^d, (s,a) \in \mS \times \mA$\,,
$
b(\theta) \eqdef
 \frac{1}{1-\gamma} \bE_{\mu_{\rho,\theta}}[\psi_\theta(\tilde{S},\tilde{A})(\hat{Q}_\theta(\tilde{S},\tilde{A}) -Q_{\pi_\theta}(\tilde{S},\tilde{A}) )]
$
and
$
\hat{Q}_\theta(s,a) \eqdef R(s,a) + \gamma \sum_{s^\prime \in \mS} p(s^\prime|s,a) \phi(s^\prime)^T \bar{\omega}_*(\theta)\,.
$
\end{theorem}
Th.~\ref{th:actor} is analog to \cite[Th.~5.5]{konda02thesis} which is established for the standard on-policy actor-critic in the average reward setting and~\cite[Th.~3]{zhang-et-al20} for an off-policy actor-critic without any target network.
The result states that the sequence $(\theta_t)$
generated by our actor-critic algorithm visits
any neighborhood of the set $\{ \theta \in \bR^d: \|\nabla J(\theta)\| \leq \|b(\theta)\|\}$ infinitely often.
The bias $b(\theta)$ corresponds to the difference between the gradient $\nabla J(\theta)$ and the steady state expectation of the actor's update direction.
The estimate used to update the actor
in Eq.~\eqref{eq:actor} is only a biased estimate of~$\nabla J(\theta)$ because of linear FA.

\begin{remark}
  \label{remark:b_theta}
  The bias $b(\theta)$ disappears in the tabular setting ($m = |\mS|$ and the features spanning $\bR^{|\mS|}$) when we
  do not use FA and in the linear FA setting when the value
  function belongs to the class of linear functions spanned by the pre-selected feature (or basis) functions.
  Beyond these particular settings,
  considering compatible features as introduced in \cite{sutton-et-al00,konda-tsitsiklis03}
  can be a solution to cancel the bias $b(\theta)$
  incurred by Algorithm~\ref{algo}. We do not investigate this direction in this work.
\end{remark}

\section{FINITE-TIME ANALYSIS}
\label{sec:finite-time_analysis}

Our analysis in this section should be valid for a continuous state space~$\mS$ (and still finite action space) upon supposing that the feature map~$\phi$ defined in Section~\ref{subsec:critic_update} has bounded norm (i.e., $\|\phi(\cdot)\| \leq 1$) and slightly adapting our notations and definitions to this more general setting (see also for e.g., \cite{wu-zha-xu-gu20}). To stay concise and consistent with the first part of our analysis in Section~\ref{sec:asymptotic_cv}, we restrict ourselves to the finite state space setting.

\subsection{Critic analysis}
For every~$\theta \in \bR^d$, we suppose that the Markov chain~$(\tilde{S}_t)$ induced by the policy~$\pi_\theta$ and the transition kernel~$\tilde{p}$ mixes at a geometric rate. 
\begin{assumption}
\label{hyp:geom_mix_mc}
There exist constants $c >0$ and $\sigma \in (0,1)$ s.t.
for every~$t \in \bN, \theta \in \bR^d$,
\[
\sup_{s \in \mS} d_{TV}(\bP(\tilde{S}_t \in \cdot| \tilde{S}_0 = s, \pi_\theta), d_{\rho,\theta}) \leq c \sigma^t\,,
\]
where $d_{TV}(\cdot,\cdot)$ denotes the total-variation distance between two probability measures.
\end{assumption}
This assumption is used to control the Markovian noise induced by sampling transitions from the MDP under a dynamically changing policy.
It was considered first in \cite{bhandari-russo-singal18} in a policy evaluation setting for the finite-time analysis of TD learning. It was later used for instance in~\cite{zou-shou-liang19,wu-zha-xu-gu20,shen-zhang-hong-chen20}.

We have seen in Sec.~\ref{subsec:critic_asympt} that the dynamics of the critic is driven by two key matrices~$- \bar{G}(\theta)$ and~$- \bar{G}(\theta)^{-1} G(\theta)$. While we only need
these matrices to be stable for our asymptotic results, we actually show in the appendix that~$- \bar{G}(\theta)$ is even negative definite uniformly in~$\theta$.
We suppose that the second matrix~$- \bar{G}(\theta)^{-1} G(\theta)$ is also negative definite uniformly in~$\theta$.

\begin{assumption}
\label{hyp:eigenval}
There exists~$\zeta>0$ s.t. for every~$\theta \in \bR^d$,\, $\omega \in \bR^m$, $\omega^T \bar{G}(\theta)^{-1} G(\theta) \omega \geq \zeta \|\omega\|^2$\,.
\end{assumption}
We are now ready to state our critic convergence rate.
\begin{theorem}
\label{th:critic_rate}
Let Assumptions~\ref{hyp:grad_pi},\, \ref{hyp:markov_chain}
and~\ref{hyp:stability} to~\ref{hyp:eigenval} hold.
Let $c_1, c_2, c_3, \alpha, \xi, \beta$ be positive constants s.t. $0 < \beta < \xi < \alpha < 1$\,.
Set $\alpha_t = \frac{c_1}{(1+t)^{\alpha}},\, \xi_t = \frac{c_2}{(1+t)^{\xi}}$ and $\beta_t = \frac{c_3}{(1+t)^{\beta}}$\,.
Then, the sequences~$(\omega_t)$ and~$(\theta_t)$ from Algorithm~\ref{algo} 
satisfy for every integer~$T\geq 1$\,,
\begin{multline*}
\frac 1T \sum_{t=1}^T
\bE[\|\omega_t - \bar{\omega}_*(\theta_t)\|^2] =          \mO\left(\frac{1}{T^{1-\xi}}\right)
  + \mO\left(\frac{\ln T}{T^{\beta}}\right)\\
  + \mO\left(\frac{1}{ T^{2(\alpha-\xi)}}\right)
  + \mO\left(\frac{1}{T^{2(\xi-\beta)}}\right)\,.
\end{multline*}
\end{theorem}

The bound of Th.~\ref{th:critic_rate} shows the impact of using a target variable. First, the last two terms impose the conditions~$\alpha >\xi$ and~$\xi > \beta$.
At least with linear FA, this may provide a theoretical justification to the common practice of updating the target network at a slower rate compared to the main network for the critic.
Second, compared to \cite[Th.~4.7]{wu-zha-xu-gu20} which is concerned with the standard actor-critic in the average reward setting, we have the slower~$\mO(T^{\xi-1})$ instead of~$\mO(T^{\beta-1})$ and our bound comprises four error terms. These are also consequences of the use of a target variable.

\begin{remark}
Although we use similar proof techniques to \cite{wu-zha-xu-gu20} for our finite-time analysis,
notice that our novel asymptotic analysis of the critic (Sec.~\ref{subsec:critic_asympt}) is crucial
for the proof (see Sec.~\ref{subsec:critic_rate} for details).
\end{remark}

\subsection{Actor analysis}

We suppose that the critic approximation error induced by linear FA is uniformly bounded (see also~\cite{qiu-et-al19,wu-zha-xu-gu20,xu-et-al20}).

\begin{assumption}
  \label{hyp:eps_fa}
There exists~$\epsilon_{\text{FA}} \geq 0$ s.t. for every~$\theta \in \bR^d,\, \|V_{\pi_\theta}- \Phi\,\bar{\omega}_*(\theta)\|_{D_{\rho,\theta}} \leq \epsilon_{\text{FA}}$\,.
\end{assumption}

Observe that~$\epsilon_{\text{FA}} = 0$ if the true value function~$V_{\pi_\theta}$ belongs to the linear function space
spanned by the feature functions for every~$\theta \in \bR^d$.

\begin{theorem}
\label{th:actor_rate}
Let Assumptions~\ref{hyp:grad_pi}\,, \ref{hyp:markov_chain}\,, 
\ref{hyp:stability} to~\ref{hyp:geom_mix_mc} and~\ref{hyp:eps_fa} hold.
Let $c_1, c_2, c_3, \alpha, \xi, \beta$ be positive constants s.t. $0 < \beta < \xi < \alpha < 1$\,.
Set $\alpha_t = \frac{c_1}{(1+t)^{\alpha}},\, \xi_t = \frac{c_2}{(1+t)^{\xi}}$ and $\beta_t = \frac{c_3}{(1+t)^{\beta}}$\,.
Then, for every integer $T\geq 1$\,,
\begin{multline*}
\frac 1T \sum_{t=1}^T \bE[\|\nabla J(\theta_t)\|^2] = \mO\left(\frac{1}{T^{1-\alpha}}\right)
    + \mO\left(\frac{\ln^2 T}{T^{\alpha}}\right)\\
    + \mO\left(\frac 1T \sum_{t=1}^T \bE[\|\omega_t - \bar{\omega}_*(\theta_t)\|^2] \right)
    + \mO\left(\epsilon_{\text{FA}}\right)\,.
\end{multline*}
\end{theorem}

Combining Th.~\ref{th:critic_rate} and Th.~\ref{th:actor_rate},
we obtain the following result.

\begin{corollary}
\label{corollary}
Under the setting and the assumptions of Ths.~\ref{th:critic_rate} and~\ref{th:actor_rate}\,, we have for every~$T\geq 1$\,,
\begin{multline*}
\frac 1T \sum_{t=1}^T \bE[\|\nabla J(\theta_t)\|^2] = \mO\left(\frac{1}{T^{1-\alpha}}\right)
    + \mO\left(\frac{\ln T}{T^{\beta}}\right)\\
    + \mO\left(\frac{1}{T^{2(\alpha-\xi)}}\right)
    + \mO\left(\frac{1}{T^{2(\xi-\beta)}}\right)
    + \mO\left(\epsilon_{\text{FA}}\right).
\end{multline*}
Moreover, if we set $\alpha = \frac 23,\, \xi = \frac 12$ and $\beta = \frac 13$
to define the stepsizes $(\alpha_t)$, $(\xi_t)$ and $(\beta_t)$,
the actor parameter sequence $(\theta_t)$ generated by Algorithm~\ref{algo} within $T = \mO(\epsilon^{-3}\ln^3(\frac{1}{\epsilon}))$ steps,
satisfies
\[
\min_{0 \leq t \leq T} \bE[\|\nabla J(\theta_t)\|^2] \leq \mO(\epsilon_{\text{FA}}) + \epsilon\,.
\]
\end{corollary}

As a consequence, since Algorithm~\ref{algo} uses a single sample from the MDP per iteration,
its sample complexity is $\mO(\epsilon^{-3}\ln^3(\frac{1}{\epsilon}))$\,. This is to compare with
the best $\mO(\epsilon^{-2}\ln(\frac{1}{\epsilon}))$ sample complexity known in the literature (to the best of our knowledge)
for actor-critic algorithms up to the linear FA error \cite[Th.~2]{xu-et-al20}.
Although the use of a target
variable seems to deteriorate the sample complexity w.r.t. the best known result for target-free actor-critic methods, note that
it is still aligned with the complexity reported in \cite{qiu-et-al19} (up to logarithmic factors) and better than the~$\mO(\epsilon^{-4})$ sample complexity obtained
in \cite{kum-kop-rib19} with i.i.d. sampling. Notice that we do not make use of mini-batching of samples (even from a single sample path) or nested loops as in \cite{xu-et-al20}. We refer to \cite[Section~4.4]{wu-zha-xu-gu20} and~\cite[Table~1]{xu-et-al20} for further discussion.
We briefly comment on the origin of this deteriorated sample complexity stemming from our finite-time bounds.
Due to the use of a target variable, instead of the $O(T^{2(\alpha-\beta)})$ error term of the standard actor-critic (see~\cite[Cor.~4.9]{wu-zha-xu-gu20} or \cite[Ths.3-4]{shen-zhang-hong-chen20}), we have two error terms $\mO(T^{2(\alpha-\xi)})$ and~$\mO(T^{2(\xi-\beta)})$ slowing down the convergence because of the condition~$\beta < \xi < \alpha$.
Interestingly, at least in the linear FA setting, this corroborates the practical intuition that the use of a target network may slow down learning as formulated for instance in~\cite[Section~3]{lillicrap-et-al16} (even if constant stepsizes are used in practice).

\begin{remark}
  \label{remark:eps_FA}
 Remark~\ref{remark:b_theta} also applies to the function approximation error~$\epsilon_{\text{FA}}$.
\end{remark}

\section{CONCLUSION AND FUTURE WORK}
\label{sec:future_work}

This paper provides the first convergence analysis of an actor-critic algorithm incorporating a target network, establishing  both asymptotic and finite-time results under Markovian sampling.
Motivated by the success of actor-critic methods using target networks in deep RL,
our analysis shows that this target network mechanism is theoretically sound in
the linear FA setting. Although our analysis does not demonstrate a particular advantage
of target-based actor-critic methods over non-target based counterpart in the linear FA setting,
our results pave the road for the nonlinear FA setting.
There are several interesting directions for future research.
A theoretical justification of the use of a target network in the nonlinear FA setting
beyond linear FA is a challenging problem that merit further investigation. In particular, as practical algorithms
in deep RL seem to indicate, it would be interesting
to see if such a trick can be a theoretically grounded alternative to the failure of temporal difference learning with nonlinear FA. Another possible avenue for future work to close the gap between theory and practice is to address the case of \textit{off-policy} target-based actor-critic algorithms which have enjoyed great empirical success \cite{fujimoto-et-al18,haarnoja-et-al18}.

\subsubsection*{Acknowledgements}

The authors would like to thank the anonymous referees for their useful feedback.
Anas Barakat was supported by the “Futur \& Ruptures” research program which is jointly funded
by the IMT, the Mines-Télécom Foundation and the Carnot TSN Institute.


\bibliography{biblio}


\clearpage
\appendix

\thispagestyle{empty}

\onecolumn \makesupplementtitle

\section{Proofs for Sec.~\ref{sec:asymptotic_cv}: asymptotic convergence results}

\subsection{Critic analysis}

The objective of this section is to prove Th.~\ref{th:critic}.
First, we recall the outline of the proof.
Our actor-critic algorithm
features three different timescales associated to three different stepsizes converging to zero with different rates,
each one associated to one of the sequences $(\theta_t), (\bar{\omega}_t)$ and $(\omega_t)$.
In spirit, we follow the strategy of \cite[Chap.~6, Lem.~1]{borkar08} for the analysis of two timescales stochastic approximation schemes.
We make use of the results of \cite{karmakar-bhatnagar18} which handles controlled Markov noise.
The proof is divided into three main steps:

\begin{enumerate}[{\sl (i)}]
   \item We start by analyzing the sequence $(\omega_t)$ evolving on the fastest timescale, i.e., with the
stepsizes $\beta_t$ which are converging the slowest to zero (see Assumption~\ref{hyp:stepsizes}). We rewrite the slower sequences $(\theta_t), (\bar{\omega}_t)$
with the stepsizes $\beta_t$. In this timescale, $(\theta_t), (\bar{\omega}_t)$ are quasi-static from the point of view of the
evolution of the sequence $(\omega_t)$. We deduce from this first step that $\omega_t$ tracks
a slowly moving target $\omega_*(\theta_t,\bar{\omega}_t)$ governed by
the slower iterates $\theta_t$ and $\bar{\omega}_t$. This is the purpose of
Prop.~\ref{prop:critic1} which is proved in Sec.~\ref{subsubsec:proof_critic1} below.

   \item In a second step, we analyze the sequence $(\bar{\omega}_t)$ which is evolving in a faster timescale than the sequence $(\theta_t)$ and slower than the sequence $(\omega_t)$.
Similarly, we show that $\bar{\omega}_t$ tracks an other slowly moving target $\bar{\omega}_*(\theta_t)$.
This is established in the proof of Prop.~\ref{prop:critic2} in Sec.~\ref{subsubsec:proof_critic2}.

   \item We conclude in Sec.\ref{subsubsec:proof_critic} by combining the results from the first two steps, proving that the sequence~$\omega_t$ tracks the same target $\bar{\omega}_*(\theta_t)$.
\end{enumerate}

\subsubsection{Proof of Prop.~\ref{prop:critic1}}
\label{subsubsec:proof_critic1}

Let $\cF_t$ be the $\sigma$-field
generated by the random variables $S_l, \tilde{S}_l, \tilde{A}_l, \theta_l, \bar{\omega}_l,\omega_l$ for $l \leq t$.
For each time step $t$, let $Z_t= (\tilde{S}_t,\tilde{A}_t)$.
Our objective here is to show that the critic sequence~$(\omega_t)$ tracks the slowly moving target~$\omega_*(\theta_t, \bar{\omega}_t)$ defined in Prop.~\ref{prop:critic1}.
From the update rule of the sequence~$(\omega_t)$, we have
\begin{align}
  \label{eq:omega_decomp}
  \omega_{t+1} &= \omega_t + \beta_t \bar{\delta}_{t+1} \phi(\tilde{S}_t)\nonumber\\
  &= \omega_t + \beta_t (R_{t+1} + \gamma \phi(S_{t+1})^T \bar{\omega}_t - \phi(\tilde{S}_t)^T \omega_t) \phi(\tilde{S}_t)\nonumber\\
  &= \omega_t + \beta_t w(\bar{\omega}_t, \omega_t, Z_t) + \beta_t \eta_{t+1}^{(1)}\,,
\end{align}
where for every~$\bar{\omega}, \omega \in \bR^m, z = (s,a) \in \mS \times \mA,$
\begin{equation}
  \label{eq:w}
w(\bar{\omega},\omega,z) \eqdef \left(R(s,a) + \gamma \sum_{s^\prime \in \mS} p(s^\prime|s,a) \phi(s^\prime)^T \bar{\omega}\right)\phi(s) - \phi(s)\phi(s)^T\omega
\end{equation}
and $\eta_{t+1}^{(1)}$ is a martingale difference sequence defined as
\begin{equation}
\eta_{t+1}^{(1)} = (R_{t+1} - R(\tilde{S}_t,\tilde{A}_t))\phi(\tilde{S}_t) + \gamma \bar{\omega}_t^T (\phi(S_{t+1}) - \bE[\phi(S_{t+1})|\cF_t])\,\phi(\tilde{S}_t)\,.
\end{equation}

As can be seen in Eq.~\eqref{eq:omega_decomp}, the sequence $(\omega_t)$ can be written as a linear stochastic approximation scheme controlled by the slowly varying Markov chains $(\theta_t)$ and $(\bar{\omega}_t)$. In view of characterizing its asymptotic behavior, we compute for fixed $\bar{\omega}, \omega \in \bR^m$ the expectation of the quantity $w(\bar{\omega},\omega,Z)$ (see Eq.~\eqref{eq:w}) where~$Z = (\tilde{S},\tilde{A})$ is a random variable (on $\mS \times \mA$) following the stationary distribution~$\mu_{\rho,\theta}$ (see Eq.~\eqref{eq:def_d_rho_theta}) of the Markov chain $(Z_t)$.
Recall the definitions of $\bar{h}: \bR^d \times \bR^m \to \bR^m$ and $\bar{G}: \bR^d \to \bR^{m\times m}$ from~Eq.~\eqref{eq:barh_barG},
for every $\theta \in \bR^d, \bar{\omega} \in \bR^m$
\begin{equation*}
\bar{h}(\theta, \bar{\omega}) \eqdef \Phi^T D_{\rho, \theta}  (R_\theta + \gamma P_\theta\Phi\, \bar{\omega})\,\quad \text{and}\quad 
\bar{G}(\theta) \eqdef \Phi^T D_{\rho, \theta} \Phi\,.
\end{equation*}

\begin{lemma}
   \label{lem:h_theta_barmu_G_theta}
   Under Assumption~\ref{hyp:markov_chain}\,, for every $\bar{\omega}, \omega \in \bR^m$, we have
   \[
   \bE_{Z \sim \mu_{\rho,\theta}}[w(\bar{\omega},\omega,Z)]
   = \bar{h}(\theta,\bar{\omega}) - \bar{G}(\theta)\omega\,.
   \]
\end{lemma}

\begin{proof}
We obtain from the definitions of~$w$ in Eq.~\eqref{eq:w} and~$\mu_{\rho,\theta}$ in Eq.~\eqref{eq:def_d_rho_theta}  that
\begin{align*}
\bE_{Z \sim \mu_{\rho,\theta}}[w(\bar{\omega},\omega,Z)]
&= \bE_{Z \sim \mu_{\rho,\theta}}\left[\left(R(\tilde{S},\tilde{A}) + \gamma \sum_{s^\prime \in \mS} p(s^\prime|\tilde{S},\tilde{A}) \phi(s^\prime)^T \bar{\omega}\right) \phi(\tilde{S}) - \phi(\tilde{S})\phi(\tilde{S})^T \omega \right]\nonumber\\
    &= \sum_{s \in \mS, a\in \mA} \mu_{\rho,\theta}(s,a) \left(R(s,a) + \gamma \sum_{s^\prime \in \mS} p(s^\prime|s,a) \phi(s^\prime)^T \bar{\omega}\right) \phi(s) - \phi(s)\phi(s)^T \omega  \nonumber\\
    &= \sum_{s\in \mS} d_{\rho,\theta}(s) \left(R_\theta(s) \phi(s)
      + \gamma \sum_{s^\prime \in \mS} p_\theta(s^\prime|s) \phi(s^\prime)^T \bar{\omega} \phi(s) -\phi(s)\phi(s)^T \omega  \right) \nonumber\\
    &= \bar{h}(\theta,\bar{\omega}) - \bar{G}(\theta)\omega\,,
\end{align*}
where the penultimate equation stems from recalling that $R_\theta(s) = \sum_{a \in \mA} R(s,a)\pi_\theta(a|s)$ and~$p_\theta(s^\prime|s) = \sum_{a \in \mA} p(s^\prime|s,a) \pi_\theta(a|s)$ for every~$s \in \mS$.
\end{proof}

Defining $\chi_t = (\theta_t, \bar{\omega}_t)$, we obtain from the update rules of~$(\theta_t)$ and~$(\bar{\omega}_t)$ that
\begin{equation}
   \label{eq:decomp_theta_barmu}
   \chi_{t+1} = \chi_t + \beta_t \varepsilon_t\,,
\end{equation}
where
$
\varepsilon_t = \left(\frac{\alpha_t}{\beta_t} \frac {1}{1-\gamma} \delta_{t+1} \psi_{\theta_t}(Z_t),\frac{\xi_t}{\beta_t}(\omega_{t+1} - \bar{\omega}_t)\right)\,.
$
Notice that $\epsilon_t \to 0$ as $t \to \infty$. This is because $\frac{\alpha_t}{\beta_t} \to 0$, $\frac{\xi_t}{\beta_t} \to 0$ by Assumption~\ref{hyp:stepsizes}\,, $(\omega_t)$ and (hence)~$(\bar{\omega}_t)$ are a.s. bounded by Assumption~\ref{hyp:stability}\,, $(R_t)$ is bounded by~$U_R$, $\theta \mapsto \psi_\theta(s,a)$ is bounded by Assumption~\ref{hyp:grad_pi}
and~$\mS, \mA$ are finite.

Let $\zeta_t = (\chi_t, \omega_t),\, \zeta = (\theta, \bar{\omega}, \omega) \in \bR^{d+2m},\, W(\zeta,z)= (0,w(\bar{\omega},\omega,z)),\, \varepsilon^{\prime}_t = (\varepsilon_t, 0)$ and~$\tilde{\eta}_{t+1}^{(1)} = (0,\eta_{t+1}^{(1)})$. Then, we can write Eqs.~\eqref{eq:decomp_theta_barmu} and~\eqref{eq:omega_decomp} in the framework of \cite[Sec.~3, Eq.(14), Lem.~9]{karmakar-bhatnagar18}, i.e., as a single timescale controlled Markov noise stochastic approximation scheme:
\begin{equation}
   \zeta_{t+1} = \zeta_t + \beta_t [ W(\zeta_t,Z_t) + \varepsilon_t^{\prime} + \tilde{\eta}_{t+1}^{(1)}]\,,
\end{equation}
with $\epsilon_t^{\prime} \to 0$\,.
Under the assumptions of \cite{karmakar-bhatnagar18} that we will verify at the end of the proof, we obtain that the sequence~$(\zeta_t)$ converges to an internally chain transitive set (i.e., a compact invariant set which has no proper attractor, see definition in \cite[Sec.~2.1]{karmakar-bhatnagar18} or \cite[Sec.~1 p.~439]{benaim96}) of the ODE
\[
\frac{d}{ds} \zeta(s) = \bar{W}(\zeta(s))\quad \text{where} \quad \bar{W}(\zeta) = (0, \bar{h}(\chi) - \bar{G}(\theta)\omega)\,,
\]
i.e.,
\begin{equation}
  \label{ode1}
\begin{cases}
  \frac{d}{ds}\chi(s) &= 0 \,,\\
  \frac{d}{ds}\omega(s) &= \bar{h}(\chi(s)) -  \bar{G}(\theta(s))\omega(s)\,.
\end{cases}
\end{equation}
As we will show that the second ODE governing $\omega$ has a unique asymptotically stable equilibrium~$\omega_*(\theta,\bar{\omega})$ for every constant function $\chi(t) = \chi = (\theta,\bar{\omega})$, it follows that
$(\chi_t,\omega_t)$ converges a.s. towards the set $\{(\chi,\omega_*(\chi)): \chi \in \bR^{d+m}\}$. In other words,
$\lim_t \|\omega_t - \omega_*(\theta_t,\bar{\omega}_t)\| = 0$, which is the desired result.

We now conclude the proof by verifying among (A1) to (A7) of \cite{karmakar-bhatnagar18} the assumptions under which \cite[Lemmas~9 and~10]{karmakar-bhatnagar18} hold.
\begin{enumerate}[{\sl (i)}]
   \item (A1): $(Z_t)$ takes values in a compact metric space. Note that it is a finite state-action Markov chain controlled by the sequence $(\theta_t)$.

   \item (A2): It is easy to see from~Eq.~\eqref{eq:w} that the drift function~$w$ is Lipschitz continuous w.r.t. the variables $\bar{\omega}, \omega$
   uniformly w.r.t. the last variable $z$ because $p$ is a probability kernel and the set of states~$\mS$ is finite.

   \item \label{mds_control} (A3): $(\tilde{\eta}_{t+1}^{(1)})$ is a martingale difference sequence w.r.t. the filtration~$(\cF_t)$\,.
    Moreover, since $(R_t)$ is bounded, there exists $K>0$ s.t. $\bE[\|\tilde{\eta}_{n+1}^{(1)}\|^2|\cF_t] \leq K (1+ \|\omega_t\|^2 + \|\bar{\omega}_t\|^2)$.

   \item \label{justif_eps_to_zero} (A4): The stepsizes~$(\beta_t)$ satisfy $\sum_t \beta_t = +\infty$ and $\sum_t \beta_t^2 < \infty$ as formulated in Assumption~\ref{hyp:stepsizes}.

   \item (A5): The transition kernel associated to the controlled Markov process $(Z_t)$ is continuous w.r.t. the variables $z \in \mS \times \mA,\, \chi \in \bR^{d+m},\, \omega \in \bR^m$. Continuity (w.r.t. to the metric of the weak convergence of probability measures) is a consequence of the fact that we have a finite-state MDP.

   \item (A6'): We first note that the inverse of the matrix $\bar{G}(\theta)$ exists thanks to Assumptions~\ref{hyp:markov_chain} and~\ref{hyp:features}\,.
   For all $\chi = (\theta, \bar{\omega}) \in \bR^{d+m}$, we now show that the ODE $\frac{d}{ds}\omega(s) = \bar{h}(\chi) - \bar{G}(\theta)\omega(s)$ has a unique globally asymptotically stable equilibrium $\omega_*(\chi) = \bar{G}(\theta)^{-1}\bar{h}(\chi)$.
   The aforementioned ODE is stable if and only if
   the matrix $\bar{G}(\theta)$ is Hurwitz. We actually show that we have a stronger result  in Lem.~\ref{lem:bar_G_unif_posdef} under Assumptions~\ref{hyp:markov_chain} and~\ref{hyp:features}\,. We briefly explicit why the assumption as formulated in the rest of (A6') holds.

   Define the function $L(\chi,\omega) = \frac 12 \|\bar{G}(\theta) \omega - \bar{h}(\chi)\|^2$\,.
   For every~$\chi = (\theta, \bar{\omega}) \in \bR^{d+m}$, the function $L(\chi,\cdot)$ is a Lyapunov function for ODE~\eqref{ode1}. Indeed, using Lem.~\ref{lem:bar_G_unif_posdef} below, we can write
   \[
  \frac{d}{ds}L(\chi,\omega(s))
  = - \ps{\bar{h}(\chi) - \bar{G}(\theta) \omega(s),
  \bar{G}(\theta) (\bar{h}(\chi) - \bar{G}(\theta) \omega(s))}
  \leq - \varepsilon \|\bar{G}(\theta)\omega(s) - \bar{h}(\chi)\|^2\,.
   \]

   \item (A7): The stability Assumption~\ref{hyp:stability} ensures that $\sup_t (\|\omega_t\| + \|\theta_t\| )< +\infty$  w.p.1\,. As a consequence, it also follows from the update rule of~$(\bar{\omega}_t)$ that~$\sup_t \|\bar{\omega}_t\|< +\infty$\,.
\end{enumerate}

\begin{lemma}
\label{lem:bar_G_unif_posdef}
Under Assumptions~\ref{hyp:markov_chain} and~\ref{hyp:features}\,, there exists~$\varepsilon > 0$ s.t. for all $\theta \in \bR^d, \omega \in \bR^m$,
\[
\omega^T\bar{G}(\theta)\omega \geq \varepsilon \|\omega\|^2\,.
\]
In particular, it holds that $\sup_{\theta \in \bR^d} \|\bar{G}(\theta)^{-1}\| < \infty$\,.
\end{lemma}
\begin{proof}
  Recall that $\mathcal{K} \eqdef \{\tilde{K}_\theta : \theta \in \bR^d \}$ where for every~$\theta \in \bR^d$, $\tilde{K}_\theta \in  \bR^{|\mS||\mA|  \times |\mS||\mA|}$ is the transition matrix over the state-action pairs defined for every $(s,a), (s^\prime,a^\prime) \in \mS \times \mA$ by $\tilde{K}_\theta(s^\prime,a^\prime|s,a) =  \tilde{p}(s^\prime|s,a)\pi_\theta(a^\prime|s^\prime)$\,.
  We also denoted by $\bar{\mathcal{K}}$ the closure of~$\mathcal{K}$.
  Under Assumption~\ref{hyp:markov_chain}\,, there exists a unique stationary distribution~$\mu_K \in \bR^{\mS \times \mA}$ for every $K \in \bar{\mathcal{K}}$.

  We first show that the map $K \mapsto \mu_K$ is continuous over the set~$\bar{\mathcal{K}}$\,. The proof of this fact is similar to the proofs of \cite[Lem.~9]{zhang-yao-whiteson21} and~\cite[Lem.~1]{marbach-tsitsiklis01}. We reproduce a similar argument here for completeness.
  Observe first that $\mu_K$ satisfies:
  \[
  M(K) \mu_K = \begin{bmatrix}
    0 \\
    1
  \end{bmatrix}\quad
  \text{where} \quad
  M(K) \eqdef
  \begin{bmatrix}
    K^T - I \\
    \1
  \end{bmatrix}\,.
  \]
  As a consequence, since~$M(K)$ has full column rank thanks to Assumption~\ref{hyp:markov_chain}\,, the matrix~$M(K)^T M(K)$ is invertible and we obtain a closed form expression for~$\mu_K$ given by:
  \[
  \mu_K = (M(K)^T M(K))^{-1} M(K)^T \begin{bmatrix}
    0 \\
    1
  \end{bmatrix}
  = \frac{\text{com}(M(K)^T M(K))^T}{\det{(M(K)^T M(K))}} M(K)^T \begin{bmatrix}
    0 \\
    1
  \end{bmatrix}\,,
  \]
  where~$\text{com}(A)$ stands for the comatrix of the matrix~$A$.
  Then, it can be seen from this expression that the map $K \mapsto \mu_K$ is continuous.
  Note for this that the entries of the comatrix are polynomial functions of the entries of~$M(K)^T M(K)$, and the determinant operator is continuous.

  It follows from Assumption~\ref{hyp:markov_chain} that for every~$K \in \bar{\mathcal{K}}$ and every~$(s,a) \in \mS \times \mA$,
  $\mu_K(s,a) > 0$\,.  We deduce from the continuity of the map $K \mapsto \mu_K$ over the compact set~$\bar{\mathcal{K}}$ that~$
  \inf_{K \in \bar{\mathcal{K}}} \mu_K(s,a)~>~0\,.\,
  $
  Since~$\tilde{K}_\theta \in \bar{K}$ for every~$\theta \in \bR^d$\,, we obtain that
  $
  \inf_{\theta} \mu_{\rho,\theta}(s,a)~>~0\,
  $
  where we recall that~$\mu_{\rho,\theta}$ is the unique stationary distribution of the Markov chain induced by~$\tilde{K}_\theta$\,. As a consequence, since~$d_{\rho,\theta}(s) = \sum_{a \in \mA} \mu_{\rho,\theta}(s,a)$, it also holds that
  \[
  \inf_{\theta} d_{\rho,\theta}(s) > 0\,.
  \]
  Therefore, for every~$\theta \in \bR^d,\, \omega \in \bR^m$:
  \[
  \omega^T \bar{G}(\theta) \omega = (\Phi \,\omega)^T D_{\rho,\theta} (\Phi \,\omega) \geq \min_{s \in \mS} \inf_{\theta} d_{\rho,\theta}(s) \|\Phi \omega\|^2
  \geq \min_{s \in \mS} \inf_{\theta} d_{\rho,\theta}(s) \lambda_{\min}(\Phi^T \Phi)\|\omega\|^2\,,
  \]
  where $\lambda_{\min}(\Phi^T \Phi) > 0$ corresponds to the smallest eigenvalue of the symmetric positive definite matrix~$\Phi^T \Phi$ which is invertible thanks to Assumption~\ref{hyp:features}\,.
  The proof is concluded by setting~$\varepsilon \eqdef \lambda_{\min}(\Phi^T \Phi) \cdot \min_{s \in \mS} \inf_{\theta} d_{\rho,\theta}(s) > 0$ which is independent of~$\theta$.
\end{proof}

\subsubsection{Proof of Prop.~\ref{prop:critic2}}
\label{subsubsec:proof_critic2}

Recall the definitions of the vector $h(\theta)$ and the matrix $G(\theta)$ from Eq.~\eqref{eq:h_G}:
\begin{equation}
h(\theta) \eqdef \Phi^T D_{\rho, \theta} R_\theta\,\quad \text{and} \quad  
G(\theta) \eqdef \Phi^T D_{\rho, \theta} (I_n - \gamma P_\theta) \Phi\,.
\end{equation}

We begin the proof by showing the existence of a unique solution $\bar{\omega}_*(\theta)$ to
the linear system $G(\theta)\bar{\omega} = h(\theta)$\,.
The following lemma establishes the uniform positive definiteness of the matrix~$G(\theta)$
Note that we do not include symmetry in our definition of positive definiteness as in \cite{bertsekas-tsitsiklis96}.
As a matter of fact, the matrix~$G(\theta)$ is not symmetric in general.
\begin{lemma}
\label{lem:G_posdef}
If Assumptions~\ref{hyp:markov_chain} and~\ref{hyp:features} hold, there exists $\kappa > 0$ s.t. for all $\theta \in \bR^d$ and $\omega \in \bR^m$,
\[
\omega^T G(\theta)\omega \geq \kappa \|\omega\|^2\,.
\]
In particular, the matrix $G(\theta)$ is invertible.
\end{lemma}

\begin{proof}
First, we have for every $\theta \in \bR^d,\, \omega \in \bR^m$,
\begin{equation}
\label{eq:G_pos_def_proof}
\omega^T G(\theta)\omega = (\Phi \omega)^T D_{\rho,\theta}(I_n - \gamma P_\theta)\Phi \omega
= (\Phi \omega)^T D_{\rho,\theta}(\Phi \omega) - \gamma (\Phi \omega)^T D_{\rho,\theta} P_\theta (\Phi \omega)\,.
\end{equation}
Then, the Cauchy-Schwarz inequality yields
\begin{equation}
\label{eq:interm_proof_lem1}
(\Phi \omega)^T D_{\rho,\theta} P_\theta (\Phi \omega)
= (\Phi \omega)^T D_{\rho,\theta}^{\frac 12} D_{\rho,\theta}^{\frac 12} P_\theta (\Phi \omega)
\leq \|\Phi \omega\|_{D_{\rho,\theta}} \|P_\theta \Phi \omega \|_{D_{\rho,\theta}}\,.
\end{equation}
Notice now that we cannot use the classical result \cite[Lem.~1]{tsitsiklis-vanroy97} to obtain that $\|P_\theta V\|_{D_{\rho,\theta}} \leq \|V\|_{D_{\rho,\theta}}$ for any $V \in \bR^n$ because $D_{\rho,\theta}$ is not the stationary distribution of the kernel $P_\theta$ but it is instead associated to the artificial kernel $\tilde{P}_\theta$. Nevertheless, the following lemma provides an analogous result with a similar proof.

\begin{lemma}
  \label{lem:G_posdef2}
  For every $\theta \in \bR^d$,\, $V \in \bR^n$, we have
  \[
  \|P_\theta V\|_{D_{\rho,\theta}}^2
  \leq \frac{1}{\gamma}\| V\|_{D_{\rho,\theta}}^2
      - \frac{1-\gamma}{\gamma} \| V\|_{\rho}^2 \leq \frac{1}{\gamma}\| V\|_{D_{\rho,\theta}}^2\,.
  \]
\end{lemma}

\begin{proof}
It follows from Jensen's inequality that
  \[
  \|P_\theta V\|_{D_{\rho,\theta}}^2 = \sum_{i=1}^n d_{\rho,\theta}(s_i) \biggl(\sum_{j = 1}^n P_\theta (s_j|s_i) V_j\biggr)^2
  \leq \sum_{i=1}^n d_{\rho,\theta}(s_i) \sum_{j = 1}^n P_\theta(s_j|s_i) V_j^2.
  \]
  Then, observe that $\tilde{P}_\theta = \gamma P_\theta + (1-\gamma)\1\rho^T$ as a consequence of Eq.~\eqref{eq:tilde_p_def}.
  By plugging this formula and then using the fact that $d_{\rho,\theta}^T\tilde{P}_\theta = d_{\rho,\theta}^T$, we obtain
  \begin{align*}
  \sum_{i=1}^n d_{\rho,\theta}(s_i) \sum_{j = 1}^n P_\theta(s_j|s_i) V_j^2 &= \frac{1}{\gamma}\biggl[\biggl(\sum_{j=1}^n \sum_{i=1}^n d_{\rho,\theta}(s_i)\tilde{P}_\theta(s_j|s_i) V_j^2 \biggr) - (1-\gamma)\sum_{j=1}^n \rho (s_j)V_j^2 \biggr]\\
  &= \frac{1}{\gamma} \biggl[\sum_{j=1}^n  d_{\rho,\theta}(s_j) V_j^2 - (1-\gamma)\sum_{j=1}^n \rho(s_j) V_j^2 \biggr]\\
  &= \frac{1}{\gamma}\| V\|_{D_{\rho,\theta}}^2
      - \frac{1-\gamma}{\gamma} \| V\|_{\rho}^2\,,
  \end{align*}
  which concludes the proof of Lem.~\ref{lem:G_posdef2}\,.
\end{proof}
We now complete the proof of Lem.~\ref{lem:G_posdef}\,. From Eq.~\eqref{eq:interm_proof_lem1}, Lem.~\ref{lem:G_posdef2} with
$V = \Phi \omega$ yields
\[
(\Phi \omega)^T D_{\rho,\theta} P_\theta (\Phi \omega) \leq \frac{1}{\sqrt{\gamma}} \|\Phi\omega\|_{D_{\rho,\theta}}^2 = \frac{1}{\sqrt{\gamma}} (\Phi \omega)^T D_{\rho,\theta} (\Phi \omega)\,.
\]
Whence, we obtain from Eq.~\eqref{eq:G_pos_def_proof} that
\[
\omega^T G(\theta)\omega \geq (1-\sqrt{\gamma}) (\Phi \omega)^T D_{\rho,\theta} (\Phi \omega) \geq \varepsilon(1-\sqrt{\gamma}) \|\omega\|^2\,,
\]
where the last inequality stems from Lem.~\ref{lem:bar_G_unif_posdef}.
\end{proof}

We now prove the remaining convergence results. We start with the first result showing that the sequence $(\bar{\omega}_t)$ tracks $\bar{\omega}_*(\theta_t)$\,.
From the update rules of the sequences $(\bar \omega_t)$ and~$(\omega_t)$ (Eqs.~\eqref{eq:critic}-\eqref{eq:critic-bar}), we can introduce the quantity $\omega_*(\theta_t, \bar{\omega}_t)$ as defined in Prop.~\ref{prop:critic1} to obtain
\begin{align}
\bar{\omega}_{t+1} &= \bar{\omega}_t + \xi_t(\omega_{t+1} - \bar{\omega}_t)\nonumber\\
&= \bar{\omega}_t + \xi_t(\omega_t + \beta_t w(\bar{\omega}_t, \omega_t, Z_t) + \beta_t \eta_{t+1}^{(1)} - \bar{\omega}_t)\nonumber\\
&= \bar{\omega}_t + \xi_t (\omega_*(\theta_t, \bar{\omega}_t) - \bar{\omega}_t) + \xi_t (\omega_t - \omega_*(\theta_t,\bar{\omega}_t) + \beta_t w(\bar{\omega}_t, \omega_t, Z_t)) + \xi_t \beta_t \eta_{t+1}^{(1)}\,.
\end{align}

Then, using the expressions of $\bar{h}, \bar{G}$ in Eq.~\eqref{eq:barh_barG} and $h, G$ in Eq.~\eqref{eq:h_G}, we can write
\begin{equation*}
   \omega_*(\theta_t,\bar{\omega}_t) - \bar{\omega}_t = \bar{G}(\theta_t)^{-1}(\bar{h}(\theta_t,\bar{\omega}_t) - \bar{G}(\theta_t)\bar{\omega}_t)
                                             = \bar{G}(\theta_t)^{-1}(h(\theta_t) - G(\theta_t)\bar{\omega}_t)\,.
\end{equation*}
As a consequence,
\begin{equation}
\label{eq:barmu_decomp_bis}
\bar{\omega}_{t+1} = \bar{\omega}_t + \xi_t \bar{G}(\theta_t)^{-1}(h(\theta_t) - G(\theta_t)\bar{\omega}_t)
                              + \xi_t (\omega_t - \omega_*(\theta_t,\bar{\omega}_t) + \beta_t w(\bar{\omega}_t, \omega_t, Z_t)) + \xi_t \beta_t \eta_{t+1}^{(1)}\,.
\end{equation}
Therefore, the sequence~$(\bar{\omega}_t)$
satisfies a linear stochastic approximation scheme driven by the slowly varying Markov chain~$(\theta_t)$ evolving on a slower timescale than the iterates $(\bar{\omega}_t)$. We proceed similarly to the proof of Prop.~\ref{prop:critic1}\,.

Recall the notation $\chi_t = (\theta_t, \bar{\omega}_t)$. Let $\chi = (\theta,\bar{\omega}) \in \bR^{d+m},\, U(\chi) = (0,\bar{G}(\theta)^{-1}(h(\theta) - G(\theta)\bar{\omega}))$. Then,
\begin{equation}
\chi_{t+1} = \chi_t + \xi_t [U(\chi_t) + \tilde{\varepsilon}_t]\,,
\end{equation}
where $\tilde{\varepsilon}_t = (\frac{\alpha_t}{\xi_t}\frac{1}{1-\gamma} \delta_{t+1}\psi_{\theta_t}(\tilde{S}_t, \tilde{A}_t) , \omega_t - \omega_*(\theta_t,\bar{\omega}_t) + \beta_t w(\bar{\omega}_t,\omega_t,Z_t) + \beta_t \eta_{t+1}^{(1)})$\,. 

It can be shown that $\tilde{\varepsilon}_t \to 0$ as $t \to +\infty$.
Note for this that $\alpha_t/\xi_t \to 0$ and $\beta_t \to 0$ by Assumption~\ref{hyp:stepsizes}\,, $\omega_t - \omega_*(\theta_t,\bar{\omega}_t) \to 0$ as proved in Prop.~\ref{prop:critic1} and $\delta_{t+1}\psi_{\theta_t}(\tilde{S}_t, \tilde{A}_t), w(\bar{\omega}_t,\omega_t,Z_t)$ are bounded by Assumptions~\ref{hyp:grad_pi}-\ref{hyp:psi_theta}\,, \ref{hyp:stability}, the boundedness of the reward function~$R$ and the fact that the sets $\mS, \mA$ are finite.
Moreover, Assumption~\ref{hyp:stepsizes} ensures that $\sum_t \xi_t = +\infty$ and $\sum_t \xi_t^2 <+\infty$\,.

Furthermore, one can show that the function~$U$ is Lipschitz continuous. For this, remark that:
\begin{enumerate}[{\sl (a)}]
  \item The function~$U$ is affine in~$\bar{\omega}$.
  \item The functions $\theta \mapsto R_\theta$ and $\theta \mapsto P_\theta$ are Lipschitz continuous as $P_\theta(s^\prime|s)= p(s^\prime|s,a)\pi_\theta(a|s)$\,, $R_\theta(s) = \sum_{a \in \mA} R(s,a)\pi_\theta(a|s)$\, and Assumption~\ref{hyp:grad_pi}-\ref{hyp:pi_theta} guarantees that $\theta \mapsto \pi_\theta(a|s)$ is Lipschitz continuous for every $(s,a) \in \mS \times \mA$\,.
  \item The function~$\theta \mapsto D_{\rho,\theta}$ is Lipschitz continuous. We refer to~\cite[Lem.~9]{zhang-yao-whiteson21} for a proof.
  \item The function~$\theta \mapsto \bar{G}(\theta)^{-1}$ is Lipschitz continuous. Observe for this that for every~$\theta, \theta^\prime \in \bR^d\,, \bar{G}(\theta)^{-1} - \bar{G}(\theta^\prime)^{-1} = \bar{G}(\theta)^{-1}(\bar{G}(\theta^\prime) - \bar{G}(\theta))\bar{G}(\theta^\prime)^{-1}$ and that  $\sup_\theta \|\bar{G}(\theta)^{-1}\| < \infty$ using Lem.~\ref{lem:bar_G_unif_posdef}.
  \item The reward function~$R$ is bounded and the entries of the matrices~$D_{\rho,\theta}$ and $P_\theta$ are bounded by one.
\end{enumerate}

Using classical stochastic approximation results (see, for e.g., \cite[Th.1.2]{benaim96}),
we obtain that the sequence~$(\chi_t)$ converges a.s. towards an internally chain transitive set of the ODE
$\frac{d}{ds}\chi(s) = U(\chi(s))\,,$
i.e.,
\begin{equation}
\begin{cases}
  \frac{d}{ds}\theta(s) &= 0 \,,\\
  \frac{d}{ds}\bar{\omega}(s) &= \bar{G}(\theta(s))^{-1}(h(\theta(s)) - G(\theta(s))\bar{\omega}(s))\,.
\end{cases}
\end{equation}
We conclude by showing that for every $\theta \in \bR^d$, the ODE $\frac{d}{ds}\bar{\omega}(s) = \bar{G}(\theta)^{-1}(h(\theta) - G(\theta)\bar{\omega}(s))$ has a globally asymptotically stable equilibrium~$\bar{\omega}_*(\theta)$. This result holds if the matrix $- \bar{G}(\theta)^{-1}G(\theta)$ is Hurwitz,
i.e., all its eigenvalues have negative real parts. We show this result in Lem.~\ref{lem:hurwitz_critic2} below.

Then, it follows that $\chi_t = (\theta_t,\bar{\omega}_t)$ converges a.s.
towards the set $\{(\theta,\bar{\omega}_*(\theta)): \theta \in \bR^d\}$. This yields the desired result
$\lim_t \|\bar{\omega}_t - \bar{\omega}_*(\theta_t)\| = 0$.

\begin{lemma}
\label{lem:hurwitz_critic2}
For every $\theta \in \bR^d,$ the matrix $- \bar{G}(\theta)^{-1}G(\theta)$ is Hurwitz\,.
\end{lemma}
\begin{proof}
We first recall Lyapunov's theorem which characterizes Hurwitz matrices (see, for e.g., \cite[Th.2.2.1 p. 96]{hor-joh-(livre)-topics}). A complex matrix $A$ is Hurwitz if and only if there exists a positive definite matrix $M = M^*$ s.t.~$A^*M + MA$ is negative definite, where $M^*$ and $A^*$ are the complex conjugate transposes of $M$ and~$A$\,. We use this theorem with $A = -\bar{G}(\theta)^{-1}G(\theta)$ and $M = \bar{G}(\theta)$ which is symmetric by definition and positive definite thanks to~Lem.~\ref{lem:bar_G_unif_posdef}. Then, we obtain that
\[
A^*M + MA = - G(\theta)^T \bar{G}(\theta)^{-1}\bar{G}(\theta) - \bar{G}(\theta)\bar{G}(\theta)^{-1}G(\theta) = - (G(\theta)^T + G(\theta))\,.
\]
We conclude the proof by showing that $G(\theta)^T + G(\theta)$ is a (symmetric) positive definite matrix. For that, observe that for every nonzero vector $\omega \in \bR^m$, it holds that $\omega^T (G(\theta)^T + G(\theta))\omega = 2 \omega^T G(\theta) \omega > 0$ where the positivity stems from Lem.~\ref{lem:G_posdef}.
\end{proof}

The last result states that for every~$\theta \in \bR^d,$ $\Phi \bar{\omega}_*(\theta)$ is a fixed point of the projected Bellman operator~$\Pi_\theta T_\theta$. This is a consequence of the following derivations:
\begin{align}
\Pi_\theta T_\theta (\Phi \bar{\omega}_*(\theta)) &=
\Phi \bar{G}(\theta)^{-1} \Phi^T D_{\rho,\theta} T_\theta(\Phi \bar{\omega}_*(\theta))\nonumber\\
&= \Phi \bar{G}(\theta)^{-1} \Phi^T D_{\rho,\theta} (R_\theta + \gamma P_\theta \Phi \bar{\omega}_*(\theta))\nonumber\\
&= \Phi \bar{G}(\theta)^{-1} h(\theta) + \Phi \bar{G}(\theta)^{-1}(\bar{G}(\theta) - G(\theta))G(\theta)^{-1}h(\theta)\nonumber\\
&= \Phi \bar{G}(\theta)^{-1} h(\theta) + \Phi G(\theta)^{-1} h(\theta) - \Phi \bar{G}(\theta)^{-1} h(\theta)\nonumber\\
&= \Phi G(\theta)^{-1} h(\theta)\nonumber\\
&= \Phi\bar{\omega}_*(\theta)\,,
\end{align}
where the first equality uses the expression of the projection~$\Pi_\theta$, the second one uses the definition of the Bellman operator~$T_\theta$ and the third one stems from the definitions of the matrices~$\bar{G}(\theta)$ and $G(\theta)$ (see Eqs.~\eqref{eq:barh_barG} and~\eqref{eq:h_G}).

\subsubsection{Proof of Th.~\ref{th:critic}}
\label{subsubsec:proof_critic}

The proof of Th.~\ref{th:critic} uses both Prop.~\ref{prop:critic1} and Prop.~\ref{prop:critic2}\,.

In order to show that
$
\lim_t \|\omega_t - \bar{\omega}_*(\theta_t)\| = 0 \,\, w.p.1\,,
$
we prove the two following results:
\begin{enumerate}[{(a)}]
   \item $\lim_t \|\omega_t - \omega_*(\theta_t, \bar{\omega}_*(\theta_t))\| = 0\,\, w.p.1$\,.
   \item $\omega_*(\theta, \bar{\omega}_*(\theta)) = \bar{\omega}_*(\theta)$ for all $\theta \in \bR^d$\,.
\end{enumerate}
(a) We have the decomposition
\begin{align}
\omega_t - \omega_*(\theta_t, \bar{\omega}_*(\theta_t))
        &= [\omega_t- \omega_*(\theta_t,\bar{\omega}_t)] + [\omega_*(\theta_t,\bar{\omega}_t) - \omega_*(\theta_t,\bar{\omega}_*(\theta_t))],\nonumber\\
        &= [\omega_t- \omega_*(\theta_t,\bar{\omega}_t)] + \bar{G}(\theta_t)^{-1}(\bar{h}(\theta_t,\bar{\omega}_t)- \bar{h}(\theta_t,\bar{\omega}_*(\theta_t)))\nonumber\\
        &= [\omega_t- \omega_*(\theta_t,\bar{\omega}_t)] +
        \bar{G}(\theta_t)^{-1} \Phi^T D_{\rho,\theta_t}P_{\theta_t} \Phi (\bar{\omega}_t - \bar{\omega}_*(\theta_t))\nonumber\\
        &= [\omega_t- \omega_*(\theta_t,\bar{\omega}_t)] +
        \bar{G}(\theta_t)^{-1} (\bar{G}(\theta_t) - G(\theta_t)) (\bar{\omega}_t - \bar{\omega}_*(\theta_t))\nonumber\\
        &= [\omega_t- \omega_*(\theta_t,\bar{\omega}_t)] +
        (I_m - \bar{G}(\theta_t)^{-1}G(\theta_t)) (\bar{\omega}_t - \bar{\omega}_*(\theta_t))\,.
\end{align}
It follows from Prop.~\ref{prop:critic1} that the first term in the above decomposition goes to zero. Then, observe that $\sup_\theta \|\bar{G}(\theta)^{-1}\| < \infty$ given Lem.~\ref{lem:bar_G_unif_posdef} and $\sup_\theta \|G(\theta)\| < \infty$ thanks to the boundedness of the matrices $P_\theta$ and~$D_{\rho,\theta}$ uniformly in~$\theta$. As a consequence, the second term also converges to zero using Prop.~\ref{prop:critic2}\,.

(b) Using the definitions of the functions $\omega_*$ and $\bar{\omega}_*$, we can write for every $\theta \in \bR^d$,
\begin{align*}
\omega_*(\theta, \bar{\omega}_*(\theta)) &= \bar{G}(\theta)^{-1} \bar{h}(\theta,\bar{\omega}_*(\theta))\\
                                   &= \bar{G}(\theta)^{-1} \Phi^T D_{\rho,\theta} (R_\theta + \gamma P_\theta \Phi G(\theta)^{-1}h(\theta))\\
                                   &= \bar{G}(\theta)^{-1} ( h(\theta) + \gamma \Phi^{T} D_{\rho,\theta} P_\theta \Phi G(\theta)^{-1}h(\theta) )\\
                                   &= \bar{G}(\theta)^{-1} (I_n + \gamma \Phi^{T} D_{\rho,\theta} P_\theta \Phi G(\theta)^{-1}) h(\theta)\\
                                   &= \bar{G}(\theta)^{-1} (G(\theta) + \gamma \Phi^{T} D_{\rho,\theta} P_\theta \Phi)G(\theta)^{-1} h(\theta)\\
                                   &=  \bar{G}(\theta)^{-1}  \bar{G}(\theta) G(\theta)^{-1} h(\theta)\\
                                   &= \bar{\omega}_*(\theta)\,.
\end{align*}

For the last result, we write
\begin{align}
\| \Pi_{\theta_t} T_{\theta_t}(\Phi \omega_t) - \Phi \omega_t\|
&= \|\Phi \left( \bar{G}(\theta_t)^{-1}\Phi^T D_{\rho,\theta_t}T_{\theta_t}(\Phi \omega_t) - \omega_t\right) \| \nonumber\\
&= \|\Phi \left(\bar{G}(\theta_t)^{-1}\Phi^T D_{\rho,\theta_t} (T_{\theta_t}(\Phi \omega_t) - \Phi \omega_t) \right) \|\nonumber\\
&= \|\Phi \left(\bar{G}(\theta_t)^{-1} (h(\theta_t) - G(\theta_t)\omega_t) \right) \|\nonumber\\
&=  \|\Phi \bar{G}(\theta_t)^{-1} G(\theta_t) (\omega_t - \bar{\omega}_*(\theta_t)) \|\nonumber\\
&\leq \|\Phi\| \|\bar{G}(\theta_t)^{-1}\| \|G(\theta_t)\| \|\omega_t - \bar{\omega}_*(\theta_t)\|\,.
\end{align}

Then, as previously mentioned in the proof, observe that $\sup_\theta \|\bar{G}(\theta)^{-1}\| < \infty$ and $\sup_\theta \|G(\theta)\| < \infty$.
Since $\bar{\omega}_t -\bar{\omega}_*(\theta_t) \to 0$ as $t \to \infty$, the result follows.

\subsection{Proof of Th.~\ref{th:actor}: actor analysis}

In this subsection, we present a proof of Th.~\ref{th:actor} which is similar in spirit to the proof in \cite[Sec.~6]{konda-tsitsiklis03}. Recall the notation $Z_t = (\tilde{S}_t,\tilde{A}_t)$. Note that $(Z_t)$ is a Markov chain.
The actor parameter $\theta_t$ iterates as follows:
\begin{align}
\theta_{t+1} &= \theta_t + \alpha_t \frac{1}{1-\gamma} \delta_{t+1} \psi_{\theta_t}(\tilde{S}_t,\tilde{A }_t)\nonumber\\
             &= \theta_t + \alpha_t  \frac{1}{1-\gamma} (R_{t+1} + (\gamma \phi(S_{t+1}) - \phi(\tilde{S_t}))^T \omega_t) \psi_{\theta_t}(\tilde{S}_t,\tilde{A }_t)\nonumber\\
             &= \theta_t + \alpha_t \frac{1}{1-\gamma} (R(\tilde{S}_t,\tilde{A}_t) \psi_{\theta_t}(\tilde{S}_t,\tilde{A }_t) + H_{\theta_t}(Z_t) \omega_t) + \alpha_t \frac{1}{1-\gamma} \tilde{\eta}_{t+1}\nonumber\,,
\end{align}
where for every $\theta \in \bR^d, z = (s,a) \in \mS \times \mA$,
\[
H_\theta(z) = \psi_\theta(s,a)
\left(\gamma \sum_{s^\prime \in \mS} p(s^\prime|s,a) \phi(s^\prime) - \phi(s)\right)^T\,,
\]
and $(\tilde{\eta}_{t+1})$ is an $\bR^d$-valued $\cF_t$-martingale difference sequence defined by
\begin{equation}
\tilde{\eta}_{t+1} = (R_{t+1} - \bE[R_{t+1}|\cF_t])\psi_{\theta_t}(\tilde{S}_t,\tilde{A }_t) \ + \gamma \psi_{\theta_t}(\tilde{S}_t,\tilde{A }_t) (\phi(S_{t+1}) - \bE[\phi(S_{t+1})|\cF_t])^T\omega_t\,.
\end{equation}

We now introduce the steady-state expectation of the main term $H_\theta(Z_t)\omega_t + R(\tilde{S}_t,\tilde{A}_t) \psi_{\theta_t}(\tilde{S}_t,\tilde{A }_t)$. Recall that~$\mu_{\rho,\theta}$ is the stationary distribution of the Markov chain~$(Z_t)$.
Define the functions $\bar{H}: \bR^d \to \bR^{d \times m}$ and $u: \bR^d \to \bR^d$ for every $\theta \in \bR^d$ by
\begin{align}
\bar{H}(\theta) &= \bE_{Z \sim \mu_{\rho,\theta}}[H_\theta(Z)]\,,\label{eq:barH}
\\
u(\theta) &= \bE_{Z \sim \mu_{\rho,\theta}}[R(\tilde{S},\tilde{A}) \psi_{\theta}(\tilde{S},\tilde{A})]\,,\label{eq:u}
\end{align}
where $Z = (\tilde{S},\tilde{A})$ is a random variable following the distribution~$\mu_{\rho,\theta}$\,.

Then, we introduce the quantity $\bar{\omega}_*(\theta_t)$ which approximates well $\omega_t$ for large $t$ (in the sense of Th.~\ref{th:critic}) and only depends on the actor parameter $\theta_t$. We obtain the following decomposition
\begin{equation}
\label{eq:decomp_theta}
\theta_{t+1} = \theta_t + \alpha_t f(\theta_t)
                        + \alpha_t \frac{1}{1-\gamma}(\tilde{\eta}_{t+1} + e_t^{(1)} + e_t^{(2)})\,,
\end{equation}
where the function $f: \bR^d \to \bR^d$ and
the error terms $e_t^{(1)}$ and $e_t^{(2)}$ are defined as follows
\begin{align}
f(\theta) &= \frac{1}{1-\gamma}(\bar{H}(\theta)\,\bar{\omega}_*(\theta) + u(\theta))\,,\label{eq:f}\\
e_t^{(1)} &= (R(\tilde{S}_t,\tilde{A}_t) \psi_{\theta_t}(\tilde{S}_t,\tilde{A}_t) + H_{\theta_t}(Z_t) \bar {\omega}_*(\theta_t))
 - (\bar{H}(\theta_t)\,\bar {\omega}_*(\theta_t) + u(\theta_t))\,,\\
e_t^{(2)} &= H_{\theta_t}(Z_t) (\omega_t - \bar{\omega}_*(\theta_t))\,.
\end{align}

The bias induced by the approximation of~$\nabla J(\theta)$ by our actor-critic algorithm is defined for every~$\theta \in \bR^d$ by
\begin{equation}
\label{eq:bias}
b(\theta) \eqdef f(\theta) - \nabla J(\theta)\,.
\end{equation}

This bias is due to the linear FA of the true state-value function. It is defined as the difference between the steady-state expectation of the actor update given by the function~$f$ defined in Eq.~\eqref{eq:f} and the gradient $\nabla J(\theta)$ we are interested in\,. The following lemma provides a more explicit and interpretable expression for the bias $b(\theta)$. The state-value function $V_{\pi_\theta}$ will be seen as a vector of $\bR^{|\mS|}$.

\begin{lemma}
\label{lem:bias_explicit}
For every $\theta \in \bR^d$,
\[
b(\theta) = \frac{\gamma}{1-\gamma}
\sum_{s \in \mS, a \in \mA} \mu_{\rho,\theta}(s,a)\psi_{\theta}(s,a) \sum_{s^\prime \in \mS} p(s^\prime|s,a) (\phi(s^\prime)^T \bar{\omega}_*(\theta) - V_{\pi_\theta}(s^\prime) )\,.
\]
\end{lemma}

\begin{proof}
The expression follows from using the definition of~$b(\theta)$
and computing both the function $\bar{H}$ defined in Eq.~\eqref{eq:barH} and the gradient of the function $J$.

First, we explicit the function $\bar{H}$, writing
\begin{align}
  \label{eq:computing_barH}
\bar{H}(\theta) = \bE_{Z \sim \mu_{\rho,\theta}}[H_\theta(Z)]
&= \bE_{Z \sim \mu_{\rho,\theta}}\left[\psi_{\theta}(\tilde{S},\tilde{A}) \left(\gamma \sum_{s^\prime \in \mS} p(s^\prime|\tilde{S},\tilde{A}) \phi(s^\prime) - \phi(\tilde{S})\right)^T\right]\nonumber\\
&= \sum_{s \in \mS, a \in \mA} \mu_{\rho,\theta}(s,a) \psi_{\theta}(s,a)  \left(\gamma \sum_{s^\prime \in \mS} p(s^\prime|s,a) \phi(s^\prime)^T - \phi(s)^T\right)\nonumber\\
&= \gamma \sum_{s \in \mS, a \in \mA} \mu_{\rho,\theta}(s,a)\psi_{\theta}(s,a) \sum_{s^\prime \in \mS} p(s^\prime|s,a) \phi(s^\prime)^T\,,
\end{align}
where the last equality stems from remarking that~$\sum_{a \in \mA} \mu_{\rho,\theta}(s,a) \psi_{\theta}(s,a) = 0$.

Then, the policy gradient theorem as formulated in Eq.~\eqref{eq:pg} and the definition of the advantage function provide
\begin{align}
  \label{eq:computing_nablaJ}
(1-\gamma) \nabla J(\theta) &= \bE_{Z \sim \mu_{\rho,\theta}}[\Delta_{\pi_\theta}(\tilde{S},\tilde{A})\psi_{\theta}(\tilde{S},\tilde{A})]\nonumber\\
      &= \bE_{Z \sim \mu_{\rho,\theta}}[ (R(\tilde{S},\tilde{A})
      + \gamma \sum_{s^\prime \in \mS} p(s^\prime|\tilde{S}_t,\tilde{A }_t) V_{\pi_\theta}(s^\prime) - V_{\pi_\theta}(\tilde{S}))\psi_{\theta}(\tilde{S},\tilde{A})   ] \nonumber\\
      &= \sum_{s,a} \mu_{\rho,\theta}(s,a) (R(s,a) +
      \gamma \sum_{s^\prime \in \mS} p(s^\prime|s,a) V_{\pi_\theta}(s^\prime)
      - V_{\pi_\theta}(s))\psi_{\theta}(s,a)\nonumber\\
      &= u(\theta) + \gamma \sum_{s \in \mS, a \in \mA} \mu_{\rho,\theta}(s,a) \psi_{\theta}(s,a) \sum_{s^\prime \in \mS} p(s^\prime|s,a) V_{\pi_\theta}(s^\prime)\,.
\end{align}

The result stems from using the definition of~$b(\theta)$ together with Eqs.~\eqref{eq:computing_barH}
and~\eqref{eq:computing_nablaJ}.
\end{proof}

Using a second-order Taylor expansion of the $\tilde{L}$-Lipschitz function $\nabla J$ (again see~\cite[Lem.~4.2]{zhang-koppel-zhu-basar20}) together with Eq.~\eqref{eq:decomp_theta}, we can derive the following inequalities
\begin{align}
\label{eq:taylor_J}
J(\theta_{t+1}) &\geq J(\theta_t) + \ps{\nabla{J}(\theta_t),\theta_{t+1}-\theta_t} - L \|\theta_{t+1} - \theta_t\|^2 \nonumber\,,\\
                &\geq J(\theta_t) + \alpha_t \ps{\nabla{J}(\theta_t), f(\theta_t)}\nonumber\\
                &+ \frac{\alpha_t }{1-\gamma}\ps{\nabla{J}(\theta_t),\tilde{\eta}_{t+1}
                + e_t^{(1)} + e_t^{(2)}} - \tilde{L} \frac{\alpha_t^2}{(1-\gamma)^2} \|\delta_{t+1}\psi_{\theta_t}(\tilde{S}_t, \tilde{A}_t)\|^2\,.
\end{align}

The above inequality consists of a main term involving the function~$f$ and noise terms. The following lemma controls these noise terms which are shown to be negligible.
\begin{lemma}
\label{lem:noise_terms}
\begin{enumerate}[{\sl (a)}]
   \item \label{noise_e13} $\sum_{t=0}^\infty \alpha_t \ps{\nabla J(\theta_t), e_t^{(1)}} < \infty$\,w.p.$1$\,,
   \item \label{noise_grad_eta} $\sum_{t=0}^\infty \alpha_t \ps{\nabla J(\theta_t), \tilde{\eta}_{t+1}} < \infty$\,w.p.$1$\,,
   \item \label{noise_e2} $\lim_{t \to \infty} e_t^{(2)} = 0$\,, w.p.$1$\,,
   \item \label{noise_alpha2rest} $\sum_{t=0}^\infty \alpha_t^2 \|\delta_{t+1} \psi_{\theta_t}(\tilde{S}_t,\tilde{A}_t)\|^2 < \infty$\, w.p.$1$\,.
\end{enumerate}
\end{lemma}

\begin{proof}

\begin{enumerate}[{\sl (a)}]

\item The proof is based on the classical decomposition of the Markov noise term~$e_t^{(1)}$ using the Poisson equation \cite[p.~222-229]{ben-met-pri90}.
We refer to \cite[Lem.~7 and Sec.~A.8.3]{zhang-et-al20} for a detailed proof using this technique. The proof of our result here follows the same line. For conciseness, we only describe the necessary tools, pointing out the differences with~\cite[Lem.~7 and Sec.~A.8.3]{zhang-et-al20} which is concerned with a different algorithm.

Let $\mZ \eqdef \mS \times \mA$.
First, define the functions~$g_\theta^* : \mZ \to \bR^d$ and~$\bar{g}: \bR^d \to \bR^d$ by:
\begin{align}
\label{eq:reward_funs_mrp}
g_\theta^*(z) &\eqdef R(z) \psi_\theta(z) + H_\theta(z) \bar{\omega}_*(\theta)\,,\\
\bar{g}(\theta) &\eqdef u(\theta) + \bar{H}(\theta) \bar{\omega}_*(\theta)\,,
\end{align}
for every~$z = (s,a) \in \mZ, \theta \in \bR^d$.
Observe in particular that $e_t^{(1)} = g_{\theta_t}^*(\tilde{S}_t,\tilde{A}_t) - \bar{g}(\theta_t)$\,.
Recall that for every~$\theta \in \bR^d$, the kernel transition~$\tilde{K}_\theta$ is defined for every
$(s,a), (s^\prime,a^\prime) \in \mS \times \mA$ by~$\tilde{K}_\theta(s^\prime,a^\prime) = \tilde{p}(s^\prime|s,a)\pi_\theta(a^\prime|s^\prime)$\,
(see Assumption~\ref{hyp:markov_chain}).
The idea of the proof is to introduce for each integer~$i = 1, \cdots, d$ a
Markov Reward Process (MRP) \cite[Sec.~8.2]{puterman14} on the space~$\mZ$
induced by the transition kernel~$\tilde{K}_\theta$ and the reward function~$g_{\theta,i}^*$
($i$th coordinate of the function~$g_\theta^*$). As a consequence, the corresponding average reward
is given by~$\bar{g}_i(\theta)$ ($i$th coordinate of~$\bar{g}(\theta$)).
Then, the differential value function of the MRP is provided
by~$v_{\theta,i} \eqdef (I - \tilde{K}_\theta + \1 \mu_{\rho,\theta}^T)^{-1}
(I - \1 \mu_{\rho,\theta}^T)g_{\theta,i}^*$ as shown for instance in \cite[Sec.~8.2]{puterman14}.
The functions~$v_{\theta,i}$ for~$i = 1, \cdots, d$ define together
a vector valued function~$v_\theta : \mZ \to \bR^d$\,.
Under Assumption~\ref{hyp:markov_chain}, using similar arguments
to the proof of Lem.~\ref{lem:bar_G_unif_posdef}
(see also~\cite[Proof of Lem.~4, p.~26]{zhang-yao-whiteson21}),
we can show that the function~$K \in \bar{\mathcal{K}} \mapsto (I - K + \1 \mu_K^T)^{-1}
(I - \1 \mu_K^T)$ is continuous on the compact set~$\bar{\mathcal{K}}$. It follows
that~$\sup_{\theta,z} \|v_\theta(z)\| < \infty$ because~$\tilde{K}_\theta \in \bar{\mathcal{K}}$
for every~$\theta \in \bR^d$ and~$g_{\theta,i}^*$ is uniformly bounded w.r.t.~$\theta$ under our assumptions\,.
Moreover, the differential value function satisfies the crucial Bellman equation:
\[
v_\theta(z) = g_\theta^*(z) - \bar{g}(\theta) + \sum_{z^\prime \in \mZ} \tilde{K}_\theta(z^\prime|z) v_\theta(z)\,,
\]
for every~$z \in \mZ$\,.
We use the above Poisson equation to
express~$e_t^{(1)} = g_{\theta_t}^*(\tilde{S}_t,\tilde{A}_t) - \bar{g}(\theta_t)$
using~$v_\theta$.
The rest of the proof follows the same line
as~\cite[Lem.~7 and Sec.~A.8.3]{zhang-et-al20}.

\item First, recall that $(\tilde{\eta}_t)$ is a martingale difference sequence adapted to $\cF_t$ and so is $(\ps{\nabla J(\theta_t), \tilde{\eta}_{t+1}})$. Using the boundedness of the function $\theta \to \psi_\theta(s,a)$ guaranteed by Assumption~\ref{hyp:grad_pi}-\ref{hyp:psi_theta} with the boundedness of the rewards sequence $(R_t)$, the sequence~$(\omega_t)$ (Assumption~\ref{hyp:stability}\,) and the gradient $\nabla J$\,, one can show by Cauchy-Schwarz inequality that there exists a constant $C >0$ s.t.
$\bE[|\ps{\nabla J(\theta_t), \tilde{\eta}_{t+1}}|^2|\cF_t] \leq C$ a.s. Then, using that $\sum_t \alpha_t^2 < \infty$ (Assumption~\ref{hyp:stepsizes}\,), it follows that $\sum_t \bE[|\alpha_t\ps{\nabla J(\theta_t), \tilde{\eta}_{t+1}}|^2|\cF_t] < \infty$ a.s. We deduce from Doob's convergence theorem that item~\ref{noise_grad_eta} holds.

\item As for item~\ref{noise_e2}, we first observe that $\bar{H}(\theta_t)$ is bounded since $\theta \mapsto \psi_\theta(s,a)$ is bounded for every $(s,a) \in \mS \times \mA$ thanks again to Assumption~\ref{hyp:grad_pi}-\ref{hyp:psi_theta}.
Then, item~\ref{noise_e2} stems from the fact that $\omega_t - \bar{\omega}_*(\theta_t) \to 0$ as shown in Th.~\ref{th:critic}\,.

\item Similarly to $\bar{H}(\theta_t)$, upon noticing that the reward sequence $(R_t)$ is bounded by $U_R$ and the sequence $(\omega_t)$ is a.s. bounded by Assumption~\ref{hyp:stability}\,, the quantity $\delta_{t+1} \psi_{\theta_t}(\tilde{S}_t,\tilde{A}_t)$ is also a.s. bounded. Then, item~\ref{noise_alpha2rest} is a consequence of the square summability of the stepsizes $\alpha_t$ ($\sum_t \alpha_t^2 < \infty$) as guaranteed by Assumption~\ref{hyp:stepsizes}.
\end{enumerate}
\end{proof}

The end of the proof follows the same line as \cite[p.~1163]{konda-tsitsiklis03} (see also~\cite[p.~86]{konda02thesis}). We reproduce the argument here for completeness.
Let~$T>0$. Define a sequence $k_t$ by
\[
k_0 = 0\,, \quad k_{t+1} = \min \left\{ k \geq k_t : \sum_{i=k_t}^k \alpha_i \geq T \right\} \quad \text{for}\,\, t>0\,.
\]
Using Eq.~\eqref{eq:taylor_J} together with the Cauchy-Schwarz inequality and Eq.~\eqref{eq:bias}, we can write
\[
J(\theta_{k_{t+1}}) \geq J(\theta_{k_t}) + \sum_{k=k_t}^{k_{t+1}-1} \alpha_k (\|\nabla J(\theta_k)\|^2 - \|b(\theta_k)\| \cdot \|\nabla J(\theta_k)\|) + \upsilon_t\,,
\]
where $\upsilon_t$ is defined by
\[
\upsilon_t = \sum_{k=k_t}^{k_{t+1}-1} \left(
               \frac{\alpha_k}{1-\gamma} \ps{\nabla{J}(\theta_k),\tilde{\eta}_{k+1} + e_k^{(1)} + e_k^{(2)}}
               - \tilde{L} \frac{\alpha_k^2}{(1-\gamma)^2} \|\delta_{k+1}\psi_{\theta_k}(\tilde{S}_k, \tilde{A}_k)\|^2 \right)\,.
\]
It stems from Lem.~\ref{lem:noise_terms} that $\upsilon_t \to 0$ as $t \to +\infty$\,.
By contradiction,
if the result does not hold, the sequence~$J(\theta_k)$ would increase indefinitely. This contradicts the
boundedness of the function $J$ (note that $\theta \mapsto V_{\pi_\theta}$ is bounded since the rewards are bounded).

\section{Proofs for Sec.~\ref{sec:finite-time_analysis}: finite-time analysis}

Throughout our finite-time analysis, we will not track all the constants although these can be precisely determined. We will in particular explicit the dependence on the effective horizon~$1/(1-\gamma)$ and the cardinal~$|\mA|$ of the action space. The universal constant $C$ may change from line to line and from inequality to inequality.
It may depend on constants of the problem s.t. the Lipschitz constants of the functions~$J , \theta \mapsto \psi_\theta$, $\theta \mapsto \pi_\theta$, upperbounds of the rewards and the score function~$\psi_\theta$.

\subsection{Proof of Th.~\ref{th:critic_rate}: finite-time analysis of the critic}
\label{subsec:critic_rate}

The proof is inspired from the recent works \cite{wu-zha-xu-gu20,shen-zhang-hong-chen20}.
However, it significantly deviates from these works because of the use of a target variable $\bar{\omega}$ in Algorithm~\ref{algo}\,.
 In particular, as previously mentioned, Algorithm~\ref{algo}
involves three different timescales whereas the actor-critic algorithms considered in \cite{wu-zha-xu-gu20,shen-zhang-hong-chen20} only use two different timescales respectively associated to the critic and the actor.

We follow a similar strategy to our asymptotic analysis of the critic.
Indeed, our non-asymptotic analysis consists of two main steps based on the following decomposition:
\begin{align}
\omega_{t} - \bar{\omega}_*(\theta_t) &= \omega_{t} - \omega_*(\theta_t,\bar{\omega}_t) + \omega_*(\theta_t,\bar{\omega}_t) - \bar{\omega}_*(\theta_t)\nonumber\\
                                &= \omega_{t} - \omega_*(\theta_t,\bar{\omega}_t) + \omega_*(\theta_t,\bar{\omega}_t) - \omega_*(\theta_t,\bar{\omega}_*(\theta_t))\nonumber\\
                                &= \omega_{t} - \omega_*(\theta_t,\bar{\omega}_t)
                                + \bar{G}(\theta_t)^{-1} (\bar{h}(\theta_t,\bar{\omega}_t)- \bar{h}(\theta_t,\bar{\omega}_*(\theta_t)))\,.
\end{align}
Hence, it is sufficient to obtain a control of the convergence rates of the quantities $\omega_{t} - \omega_*(\theta_t,\bar{\omega}_t)$ and $\bar{\omega}_t - \bar{\omega}_*(\theta_t)$\,. We already know that these quantities converge a.s. to zero thanks to Props.~\ref{prop:critic1} and~\ref{prop:critic2}\,. We conduct a finite-time analysis of each of the terms separately in the subsections below and combine the obtained results to conclude the proof.

We start by introducing a few useful shorthand notations. Let $\tilde{x}_t \eqdef (\tilde{S}_t, \tilde{A}_t, S_{t+1})$\,. Define for every $\tilde{x} = (\tilde{s},\tilde{a},s) \in \mS \times \mA \times \mS$ and every~$\bar{\omega}, \omega \in \bR^m$:
\begin{align}
\bar{\delta}(\tilde x, \bar{\omega}, \omega) &=  R(\tilde s,\tilde a) + \gamma \phi(s)^T \bar{\omega} - \phi(\tilde s)^T \omega\,,\\
g(\tilde x,\bar{\omega},\omega) &= \bar{\delta}(\tilde x, \bar{\omega}, \omega) \phi(\tilde s)\,.
\end{align}
Finally, define for every~$\theta \in \bR^d$ the steady-state expectation:
\begin{equation}
\bar{g}(\theta,\bar{\omega}, \omega) = \bE_{\tilde s \sim d_{\rho,\theta}, \tilde a \sim \pi_\theta, s \sim p(\cdot|\tilde s, \tilde a)}[g(\tilde x,\bar{\omega},\omega)] = \bar{h}(\theta, \bar{\omega}) - \bar{G}(\theta)\omega\,.
\end{equation}

\subsubsection{Control of the first error term $\omega_{t} - \omega_*(\theta_t,\bar{\omega}_t)$}
\label{sec:1st-error-term}

We introduce an additional shorthand notation for brevity:
\[
\nu_t \eqdef \omega_t - \omega_*(\theta_t,\bar{\omega}_t)\,.
\]

\paragraph{Decomposition of the error.}
Using the update rule of the critic gives
\begin{align}
\|\nu_{t+1}\|^2 &= \|\omega_t + \beta_t g(\tilde{x}_t,\bar{\omega}_t,\omega_t) - \omega_*(\theta_{t+1},\bar{\omega}_{t+1})\|^2\nonumber\\
                &= \|\nu_t + \beta_t g(\tilde{x}_t,\bar{\omega}_t,\omega_t) + \omega_*(\theta_t,\bar{\omega}_t) - \omega_*(\theta_{t+1},\bar{\omega}_{t+1})\|^2\nonumber\,.
\end{align}
Then, we develop the squared norm and use the classical inequality $\|a+b\|^2 \leq 2 \|a\| + 2 \|b\|^2$ to obtain
\begin{multline}
\label{eq:decomp_critic_rate1}
\|\nu_{t+1}\|^2 \leq \|\nu_t\|^2 + 2 \beta_t \ps{\nu_t, g(\tilde{x}_t,\bar{\omega}_t,\omega_t)}+ 2 \ps{\nu_t, \omega_*(\theta_t,\bar{\omega}_t) - \omega_*(\theta_{t+1},\bar{\omega}_{t+1})}\\
 + 2 \|\omega_*(\theta_t,\bar{\omega}_t) - \omega_*(\theta_{t+1},\bar{\omega}_{t+1})\|^2 + 2 C \beta_t^2\,.
\end{multline}

Now, we decompose the first inner product into a main term generating a repelling effect and a second Markov noise term as follows
\begin{equation}
\label{eq:main_term_plus markov_noise}
\ps{\nu_t,g(\tilde{x}_t,\bar{\omega}_t,\omega_t)} = \ps{\nu_t, \bar{g}(\theta_t, \bar{\omega}_t, \omega_t)} + \Lambda(\theta_t,\bar{\omega}_t,\omega_t, \tilde{x}_t)\,,
\end{equation}
where we used the shorthand notation
\begin{equation}
\label{eq:markov_noise}
\Lambda(\theta,\bar{\omega},\omega, \tilde{x}) \eqdef \ps{\omega - \omega_*(\theta,\bar{\omega}), g(\tilde{x},\bar{\omega},\omega) - \bar{g}(\theta, \bar{\omega}, \omega)}\,.
\end{equation}

We control the first term in Eq.~\eqref{eq:main_term_plus markov_noise} as follows
\begin{equation}
\ps{\nu_t, \bar{g}(\theta_t, \bar{\omega}_t, \omega_t)} = \ps{\nu_t,\bar{g}(\theta_t, \bar{\omega}_t, \omega_t) - \bar{g}(\theta_t, \bar{\omega}_t, \omega_*(\theta_t,\bar{\omega}_t))}
= - \ps{\nu_t, \bar{G}(\theta_t) \nu_t}
\leq - \varepsilon \|\nu_t\|^2\,.
\end{equation}
We used the fact that $ \bar{g}(\theta_t, \bar{\omega}_t, \omega_*(\theta_t,\bar{\omega}_t)) = 0$ for the first equality and Lem.~\ref{lem:bar_G_unif_posdef} for the inequality.
Then, it can be shown that
\begin{equation}
\label{eq:mu_star_lip}
\|\omega_*(\theta_t,\bar{\omega}_t) - \omega_*(\theta_{t+1},\bar{\omega}_{t+1})\| \leq C (\|\theta_t - \theta_{t+1}\| + \|\bar{\omega}_t - \bar{\omega}_{t+1}\|) \leq C \left(\frac{\alpha_t}{1-\gamma} + \xi_t\right)\,.
\end{equation}

Combining Eqs.~\eqref{eq:decomp_critic_rate1} to~\eqref{eq:mu_star_lip} leads to
\begin{equation}
\label{eq:decomp_rate_critic1}
\|\nu_{t+1}\|^2 \leq (1-2\varepsilon \beta_t) \|\nu_t\|^2 + 2 \beta_t \Lambda(\theta_t,\bar{\omega}_t,\omega_t, \tilde{x}_t) + C \left(\frac{\alpha_t}{1-\gamma} + \xi_t \right) \|\nu_t\| + C\left(\frac{\alpha_t^2}{(1-\gamma)^2} + \xi_t^2 + \beta_t^2\right)\,.
\end{equation}

\paragraph{Control of the Markov noise term $\Lambda(\theta_t,\bar{\omega}_t,\omega_t, \tilde{x}_t)$\,.}

We decompose the noise term using a similar technique to~\cite{zou-shou-liang19} which was then used in \cite{wu-zha-xu-gu20,shen-zhang-hong-chen20}. Let $T > 0$. Define the mixing time
\begin{equation}
\label{eq:def_tau}
\tau_T \eqdef \min \{ t \in \bN, t \geq 1 : c \sigma^{t-1} \leq \min\{\alpha_T, \xi_T, \beta_T\} \}\,.
\end{equation}
In the remainder of the proof, we will use the notation $\tau$ for $\tau_T$ (interchangeably).
In order to control the difference between the update rule of the critic and its steady-state expectation, we introduce an auxiliary chain which coincides with $\tilde{x}_t$ except for the $\tau$ last steps where the policy is fixed to~$\pi_{\theta_{t-\tau}}$. The auxiliary chain will be denoted by~$\check{x}_t \eqdef (\check{S}_t,\check{A}_t,S_{t+1})$ where $S_{t+1} \sim p(\cdot|\check{S}_t,\check{A}_t)$ and~$(\check{S}_t,\check{A}_t)$ is generated as follows:
\[
\tilde{S}_{t-\tau} \xrightarrow{\theta_{t-\tau}} \tilde{A}_{t-\tau}
\xrightarrow{\tilde{p}} \tilde{S}_{t-\tau+1}
\xrightarrow{\theta_{t-\tau}} \check{A}_{t-\tau+1}
\xrightarrow{\tilde{p}} \check{S}_{t-\tau+2}
\xrightarrow{\theta_{t-\tau}} \check{A}_{t-\tau+2}
\xrightarrow{\tilde{p}} \cdots
\xrightarrow{\tilde{p}} \check{S}_t
\xrightarrow{\theta_{t-\tau}} \check{A}_{t}
\xrightarrow{\tilde{p}} \check{S}_{t+1}\,.
\]
Compared to this chain, the original chain has a drifting policy, i.e., at each time step, the actor parameter $\theta_t$ is updated and so is the policy $\pi_{\theta_t}$ and we recall that it is given by:
\[
\tilde{S}_{t-\tau} \xrightarrow{\theta_{t-\tau}} \tilde{A}_{t-\tau}
\xrightarrow{\tilde{p}} \tilde{S}_{t-\tau+1}
\xrightarrow{\theta_{t-\tau+1}} \tilde{A}_{t-\tau+1}
\xrightarrow{\tilde{p}} \tilde{S}_{t-\tau+2}
\xrightarrow{\theta_{t-\tau+2}} \tilde{A}_{t-\tau+2}
\xrightarrow{\tilde{p}} \cdots
\xrightarrow{\tilde{p}} \tilde{S}_t
\xrightarrow{\theta_{t}} \tilde{A}_{t}
\xrightarrow{\tilde{p}} \tilde{S}_{t+1}\,.
\]

Using the shorthand notation $z_t \eqdef (\bar{\omega}_t, \omega_t)$, the Markov noise term can be decomposed as follows:
\begin{multline}
\label{eq:markov_noise_decomp}
\Lambda(\theta_t,\bar{\omega}_t,\omega_t, \tilde{x}_t) = (\Lambda(\theta_t,z_t, \tilde{x}_t) - \Lambda(\theta_{t-\tau},z_{t-\tau}, \tilde{x}_t))
+ (\Lambda(\theta_{t-\tau},z_{t-\tau}, \tilde{x}_t)
- \Lambda(\theta_{t-\tau},z_{t-\tau}, \check{x}_t))\\
+ \Lambda(\theta_{t-\tau},z_{t-\tau}, \check{x}_t)\,.
\end{multline}

We control each one of the terms successively.
\begin{enumerate}[{\sl (a)}, leftmargin=*]
\item\underline{\textbf{Control of $\Lambda(\theta_t,z_t, \tilde{x}_t) - \Lambda(\theta_{t-\tau},z_{t-\tau}, \tilde{x}_t)$:}}
Using that $\omega_*$ and $\bar{g}$ are Lipschitz in all their arguments, $g$ is Lipschitz in its two last arguments and $\omega_t, \omega_*, g$ and $\bar{g}$ are all bounded, one can show after tedious decompositions that
\begin{equation}
\label{eq:control_noise1}
|\Lambda(\theta_t,z_t, \tilde{x}_t) - \Lambda(\theta_{t-\tau},z_{t-\tau}, \tilde{x}_t)| \leq
                        C (\|\theta_t - \theta_{t-\tau}\| + \|\bar{\omega}_t - \bar{\omega}_{t-\tau} \| + \|\omega_t - \omega_{t-\tau} \|)\,.
\end{equation}
Then, recalling that the sequence $(\alpha_t)$ is nonincreasing, remark that
\[
\|\theta_t - \theta_{t-\tau}\| \leq \sum_{t-\tau}^{t-1} \|\theta_{j+1}-\theta_j\| \leq \frac{C}{1-\gamma} \sum_{t-\tau}^{t-1} \alpha_j \leq \frac{C}{1-\gamma} \tau \alpha_{t-\tau}\,.
\]
Similarly, we have $\|\bar{\omega}_t - \bar{\omega}_{t-\tau} \| \leq C \tau \xi_{t-\tau}$, $\|\omega_t - \omega_{t-\tau}\| \leq C \tau \beta_{t-\tau}$ and we can therefore deduce from Eq.~\eqref{eq:control_noise1} that
\begin{equation}
\label{eq:control_noise1bis}
|\Lambda(\theta_t,z_t, \tilde{x}_t) - \Lambda(\theta_{t-\tau},z_{t-\tau}, \tilde{x}_t)| \leq
                        C \tau \left( \frac{\alpha_{t-\tau}}{1-\gamma} + \beta_{t-\tau} + \xi_{t-\tau}\right)\,.
\end{equation}

\item \underline{\textbf{Control of $\Lambda(\theta_{t-\tau},z_{t-\tau}, \tilde{x}_t) - \Lambda(\theta_{t-\tau},z_{t-\tau}, \check{x}_t)$:}} following similar arguments to \cite{wu-zha-xu-gu20,shen-zhang-hong-chen20}, we upperbound the conditional expectation of this error term w.r.t. $\tilde{S}_{t-\tau + 1}, \bar{\omega}_{t-\tau},\omega_{t-\tau}$ and $\theta_ {t-\tau}$. Note that our definition of $\check{x}_t$ is slightly different from the ones used in the two aforementioned references because of the third component of $\check{x}_t$ (and also $\tilde{x}_t$) which is generated according to the original kernel $p$ instead of the artificial kernel $\tilde{p}$. We have
\begin{align}
\label{eq:control_noise2}
\bE[\Lambda(\theta_{t-\tau},z_{t-\tau}, \tilde{x}_t) - \Lambda(\theta_{t-\tau},z_{t-\tau}, \check{x}_t)|\tilde{S}_{t-\tau+1}, \theta_{t-\tau}]
&= \bE[\ps{\nu_{t-\tau},g(\tilde{x}_t,z_{t-\tau}) - g(\check{x}_t,z_{t-\tau})}| \tilde{S}_{t-\tau+1}, \theta_{t-\tau}]\nonumber\\
&\leq C d_{TV}(\bP(\tilde{x}_t \in \cdot|\tilde{S}_{t-\tau+1}, \theta_{t-\tau}), \bP(\check{x}_t \in \cdot|\tilde{S}_{t-\tau+1}, \theta_{t-\tau}))\nonumber\\
&\leq \frac C2 |\mA| L_\pi \sum_{i=t-\tau}^t \bE[\|\theta_i - \theta_{t-\tau}\|| \tilde{S}_{t-\tau+1}, \theta_{t-\tau}]\,,
\end{align}
where the first equality stems from the definition of $\Lambda$, the first inequality uses the definition of the total variation distance $d_{TV}$ between two probability measures and the last inequality is a consequence of \cite[Lem.~B.2, p.17]{wu-zha-xu-gu20} (see also \cite[Lem.~2 p.12]{shen-zhang-hong-chen20}).

Then, we have
\begin{multline*}
\sum_{i= t-\tau}^t \bE[\|\theta_i - \theta_{t-\tau}\|| \tilde{S}_{t-\tau+1}, \theta_{t-\tau}] \leq \sum_{i=t-\tau}^{t} \sum_{j=t-\tau}^{i-1} \bE[\|\theta_{j+1}-\theta_j\||\tilde{S}_{t-\tau+1}, \theta_{t-\tau}]\\
\leq \frac{C}{1-\gamma} \sum_{i=t-\tau}^{t} \sum_{j=t-\tau}^{i-1} \alpha_j
\leq \frac{C}{1-\gamma} \alpha_{t-\tau} \sum_{i=0}^\tau i
\leq \frac{C}{1-\gamma} \alpha_{t-\tau} (\tau +1)^2\,.
\end{multline*}
As a consequence of these derivations, Eq.~\eqref{eq:control_noise2} yields
\begin{equation}
\label{eq:control_noise2bis}
\bE[\Lambda(\theta_{t-\tau},z_{t-\tau}, \tilde{x}_t) - \Lambda(\theta_{t-\tau},z_{t-\tau}, \check{x}_t)| \tilde{S}_{t-\tau+1}, \theta_{t-\tau}]
\leq \frac{C}{1-\gamma} |\mA| \alpha_{t-\tau} (\tau +1)^2\,,
\end{equation}

\item \underline{\textbf{Control of $\Lambda(\theta_{t-\tau},z_{t-\tau}, \check{x}_t):$}}
Define $\bar{x}_t \eqdef (\bar{S}_t, \bar{A}_t, S_{t+1})$ where $\bar{S}_t \sim d_{\rho,\theta_{t-\tau}},\, \bar{A}_t \sim \pi_{\theta_{t-\tau}}$ and $S_{t+1} \sim p(\cdot|\bar{S}_t, \bar{A}_t)$. Observing that $\bE[\Lambda(\theta_{t-\tau},z_{t-\tau}, \bar{x}_t)|\tilde{S}_{t-\tau+1},\theta_{t-\tau}] = 0$, we obtain
\begin{align}
\label{eq:control_noise3}
\bE[\Lambda(\theta_{t-\tau},z_{t-\tau}, \check{x}_t)|\tilde{S}_{t-\tau+1},\theta_{t-\tau}] &=
              \bE[\Lambda(\theta_{t-\tau},z_{t-\tau}, \check{x}_t) - \Lambda(\theta_{t-\tau},z_{t-\tau}, \bar{x}_t)|\tilde{S}_{t-\tau+1},\theta_{t-\tau}]\nonumber\\
              &= \bE[\ps{\nu_{t-\tau},g(\check{x}_t,z_{t-\tau}) - g(\bar{x}_t,z_{t-\tau})}|\tilde{S}_{t-\tau+1},\theta_{t-\tau}]\nonumber\\
              &\leq C d_{TV}(\bP(\check{x}_t \in \cdot|\tilde{S}_{t-\tau+1},\theta_{t-\tau}), \bP(\bar{x}_t \in \cdot|\tilde{S}_{t-\tau+1},\theta_{t-\tau}))\nonumber\\
              &= C d_{TV}(\bP(\tilde{S}_t \in \cdot|\tilde{S}_{t-\tau+1},\theta_{t-\tau}), d_{\rho,\theta_{t-\tau}})\nonumber\\
              &\leq C \sigma^{\tau-1}\nonumber\\
              &\leq C \alpha_T\,,
\end{align}
where the first inequality stems again from the definition of the total variation norm and the last two ones follow from Assumption~\ref{hyp:geom_mix_mc} and the definition of the mixing time~$\tau = \tau_T$ (see Eq.~\eqref{eq:def_tau}).
\end{enumerate}

Given the decomposition of Eq.~\eqref{eq:markov_noise_decomp}, collecting Eqs.\eqref{eq:control_noise1bis}, \eqref{eq:control_noise2bis}, \eqref{eq:control_noise3} and taking total expectation leads to the conclusion of this subsection
\begin{equation}
\label{eq:markov_noise_bound}
\bE[\Lambda(\theta_t,z_t,\tilde{x}_t)] \leq C \left(\tau \left( \frac{\alpha_{t-\tau}}{1-\gamma} + \beta_{t-\tau} + \xi_{t-\tau}\right)  + |\mA|\frac{\alpha_{t-\tau}}{1-\gamma} (\tau +1)^2 + \alpha_T \right)\,.
\end{equation}

\paragraph{Derivation of the convergence rate of the mean error term $\frac 1T \sum_{t=1}^T \|\nu_t\|^2$\,.}
We obtain from taking the total expectation in Eq.~\eqref{eq:decomp_rate_critic1} together with Eq.~\eqref{eq:markov_noise_bound} that
\begin{multline}
\label{eq:decomp_rate_critic1bis}
\bE[\|\nu_{t+1}\|^2] \leq (1-2\varepsilon \beta_t) \bE[\|\nu_t\|^2]
         + 2 C \beta_t \left(\tau \left( \frac{\alpha_{t-\tau}}{1-\gamma} + \beta_{t-\tau} + \xi_{t-\tau}\right)  + |\mA|\frac{\alpha_{t-\tau}}{1-\gamma} (\tau +1)^2 + \alpha_T\right)\\
         + C \left(\frac{\alpha_t}{1-\gamma} + \xi_t\right) \bE[\|\nu_t\|] + C\left(\frac{\alpha_t^2}{(1-\gamma)^2} + \xi_t^2 + \beta_t^2\right)\,.
\end{multline}

Rearranging the inequality and summing for $t$ between $\tau_T$ and T, we get
\begin{equation}
  \label{eq:sum_I}
2 \varepsilon \sum_{t=\tau_T}^T \bE[\|\nu_t\|^2] \leq I_1(T) + I_2(T) + I_3(T) + I_4(T)\,,
\end{equation}
where
\begin{align}
I_1(T) &\eqdef \sum_{t=\tau_T}^T \frac{1}{\beta_t} (\bE[\|\nu_t\|^2] - \bE[\|\nu_{t+1}\|^2])\,,\\
I_2(T) &\eqdef \sum_{t=\tau_T}^T 2C \left(\tau \left( \frac{\alpha_{t-\tau}}{1-\gamma} + \beta_{t-\tau} + \xi_{t-\tau}\right)  + |\mA|\frac{\alpha_{t-\tau}}{1-\gamma} (\tau +1)^2 + \alpha_T\right)\\
I_3(T) &\eqdef C \sum_{t=\tau_T}^T \left(\frac{\alpha_t}{(1-\gamma)\beta_t} + \frac{\xi_t}{\beta_t}\right) \bE[\|\nu_t\|]  \\
I_4(T) &\eqdef C \sum_{t=\tau_T}^T \frac{\alpha_t^2}{(1-\gamma)^2 \beta_t} + \frac{\xi_t^2}{\beta_t} + \beta_t \,.
\end{align}

We derive estimates of each one of the terms $I_i(T)$ for $i=1,2,3,4$.
\begin{enumerate}[{\sl (1)}]
\item Since $(\nu_t)$ is a bounded sequence,
\begin{multline}
I_1(T) = \sum_{t=\tau_T}^T \left(\frac{1}{\beta_t} - \frac{1}{\beta_{t-1}}\right) \bE[\|\nu_t\|^2]
                     + \frac{1}{\beta_{\tau_T - 1}} \bE[\|\nu_{\tau_T}\|^2]
                     - \frac{1}{\beta_{\tau_T}} \bE[\|\nu_{T+1}\|^2] \\
      \leq C \left[\sum_{t=\tau_T}^T \left(\frac{1}{\beta_t} - \frac{1}{\beta_{t-1}}\right) + \frac{1}{\beta_{\tau_T-1}} \right]
      = \frac{C}{\beta_T} = \mO(T^{\beta})\,.
\end{multline}
Then, since~$\tau_T = \mO(\ln T)$, it follows that
\[
\frac{1}{1+T-\tau_T} I_1(T) \leq \frac{1}{1+T-\tau_T} \frac{C}{\beta_T} = \frac{1}{T(\frac{1}{T} +1 - \frac{\tau_T}{T})} \frac{C}{\beta_T} = \mO(T^{\beta -1})\,.
\]

\item Using the inequality $\sum_{k=l}^{p} k^{-\beta} \leq \frac{p^{1-\beta}}{1-\beta}$ for $1 \leq l < p$ and the fact that $\tau_T = \mO(\ln T)$, we have
\begin{align}
I_2(T) &\leq  C \left( \tau_T \sum_{t=0}^{T-\tau} \left(\frac{\alpha_t}{1-\gamma} + \beta_t + \xi_t\right) + |\mA| \frac{(\tau +1)^2}{1-\gamma} \sum_{t=0}^{T-\tau} \alpha_t + (1+T-\tau)\alpha_T \right)\nonumber\\
    &\leq \frac{C}{1-\gamma}(\tau(1+T)^{1-\beta} + (\tau +1)^2|\mA|(1+T)^{1-\alpha})\nonumber\\
    &= \mO\left(\frac{\ln T}{1-\gamma}T^{1-\beta}\right) + \mO\left(\frac{|\mA|}{1-\gamma}\ln^2(T) T^{1-\alpha}\right) = \mO\left(  \frac{|\mA|}{1-\gamma}\ln(T) T^{1-\beta} \right)\,,
\end{align}
where we recall for the second inequality that $0 < \beta < \xi < \alpha < 1$\, and for the last equality, we recall that $|\mA|$ is finite.
As a consequence,
\[
\frac{1}{1+T-\tau_T} I_2(T) = \mO\left(\frac{|\mA|}{1-\gamma} \ln(T) T^{-\beta}\right)\,.
\]

\item Using the Cauchy-Schwarz inequality, we can write:
\begin{align}
I_3(T) &= \sum_{t= \tau_T}^T C \left( \frac{\alpha_t}{(1-\gamma)\beta_t} + \frac{\xi_t}{\beta_t} \right) \bE[\|\nu_t\|]\nonumber\\
&\leq C \sqrt{\sum_{t=\tau_T}^T \left( \frac{\alpha_t}{(1-\gamma)\beta_t} + \frac{\xi_t}{\beta_t}\right)^2} \sqrt{\sum_{t=\tau_T}^T \bE[\|\nu_t\|^2]}\,.
\end{align}
Then, observing that the sequences~$(\frac{\alpha_t}{\beta_t})$ and~$(\frac{\xi_t}{\beta_t})$ are nonincreasing, we have:
\begin{align}
\frac{1}{1+T-\tau_T} \sum_{t=\tau_T}^T \left( \frac{\alpha_t}{(1-\gamma)\beta_t} + \frac{\xi_t}{\beta_t} \right)^2
&\leq \frac{2}{1+T-\tau_T} \sum_{t=\tau_T}^T
\left( \left(\frac{\alpha_t}{(1-\gamma)\beta_t}\right)^2
+ \left(\frac{\xi_t}{\beta_t}\right)^2 \right)\nonumber\\
&=  \frac{2}{1+T-\tau_T} \sum_{t=0}^{T-\tau_T} \left( \left(\frac{\alpha_{t+\tau_T}}{(1-\gamma)\beta_{t+\tau_T}}\right)^2
+ \left(\frac{\xi_{t+\tau_T}}{\beta_{t+\tau_T}}\right)^2 \right)\nonumber\\
&\leq \frac{2}{T-\tau_T+1} \sum_{t=0}^{T-\tau_T}
\left( \left(\frac{\alpha_t}{(1-\gamma)\beta_t}\right)^2
+ \left(\frac{\xi_t}{\beta_t}\right)^2 \right)\nonumber\\
&\leq \frac{(T-\tau_T +1)^{-2(\alpha -\beta)}}{(1-\gamma)^2(1-2(\alpha-\beta))} + \frac{(T-\tau_T +1)^{-2(\xi -\beta)}}{1-2(\xi-\beta)}\nonumber\\
&= \mO\left(\frac{T^{-2(\alpha -\beta)}}{(1-\gamma)^2}  + T^{-2(\xi -\beta)}\right)\,.
\end{align}

\item Similarly to item~(3), to control the fourth term, we write:
\begin{align}
\frac{1}{1+T-\tau_T} \sum_{t=\tau_T}^T \left(\frac{\alpha_t^2}{(1-\gamma)^2\beta_t}
+ \frac{\xi_t^2}{\beta_t}  + \beta_t  \right)
&\leq \frac{1}{1+T-\tau_T} \sum_{t=0}^{T-\tau_T}
\left(\frac{\alpha_t^2}{(1-\gamma)^2\beta_t}
+ \frac{\xi_t^2}{\beta_t}  + \beta_t  \right)\nonumber\\
&\leq \frac{(1+T-\tau_T)^{-(2\alpha- \beta)}}{(1-\gamma)^2 (1- (2\alpha -\beta))} + \frac{(1+T-\tau_T)^{-(2\xi- \beta)}}{1- (2\xi -\beta)}\nonumber\\
&+ \frac{(1+T-\tau_T)^{-\beta}}{1-\beta}\nonumber\\
&= \mO\left(\frac{T^{-(2\alpha- \beta)}}{(1-\gamma)^2} + T^{-(2\xi- \beta)} + T^{-\beta}\right)\,.
\end{align}
Hence,
\begin{equation}
\frac{1}{1+T-\tau_T} I_4(T) =  \mO\left(\frac{T^{-(2\alpha- \beta)}}{(1-\gamma)^2} + T^{-(2\xi- \beta)} + T^{-\beta}\right)\,.
\end{equation}
\end{enumerate}

Define:
\begin{align}
N(T) &\eqdef \frac{1}{1+T-\tau_T} \sum_{t=\tau_T}^T \bE[\|\nu_t\|^2]\,,\\
F(T) &\eqdef \frac{1}{1+T-\tau_T} \sum_{t=\tau_T}^T \left( \left(\frac{\alpha_t}{(1-\gamma)\beta_t}\right)^2 + \left(\frac{\xi_t}{\beta_t}\right)^2 \right)\,,\\
G(T) &\eqdef \frac{1}{1+T-\tau_T} (I_1(T) + I_2(T) + I_4(T))\,.
\end{align}

Using items~(1) to~(4), we have:
\begin{align}
F(T)  &= \mO\left(\frac{T^{-2(\alpha -\beta)}}{(1-\gamma)^2}  + T^{-2(\xi -\beta)}\right) \,,\\
G(T)  &= \mO(T^{\beta -1}) +  \mO\left(\frac{|\mA|}{1-\gamma} \ln(T) T^{-\beta}\right) + \mO\left(\frac{T^{-(2\alpha- \beta)}}{(1-\gamma)^2} + T^{-(2\xi- \beta)} + T^{-\beta}\right)\,.
\end{align}

From Eq.~\eqref{eq:sum_I} and items~(1) to~(4) above, we have:
\[
2 \varepsilon N(T) \leq C \sqrt{F(T)} \sqrt{N(T)} + G(T)\,.
\]
Solving this inequality yields:
\[
N(T) = \mO(F(T) + G(T))\,.
\]
Remarking that $0< 2(\alpha -\beta) < 2\alpha -\beta$ and $0< 2(\xi -\beta)  < 2\xi -\beta$, we obtain:
\[
N(T) = \mO(T^{\beta -1}) + \mO\left(\frac{|\mA|}{1-\gamma} \ln(T) T^{-\beta}\right) + \mO\left(\frac{T^{-2(\alpha -\beta)}}{(1-\gamma)^2}\right) + \mO(T^{-2(\xi-\beta)})\,.
\]

Then, we conclude that:
\[
\frac{1}{T} \sum_{t=1}^T \bE[\|\nu_t\|^2] = \mO(\ln(T) T^{-1}) + \mO(N(T)) = \mO(N(T))\,.
\]

\subsubsection{Control of the second error term $\bar{\omega}_t - \bar{\omega}_*(\theta_t)$}
\label{sec:2nd-error-term}

Consider the shorthand notation~$\bar{\nu}_t \eqdef \bar{\omega}_t - \bar{\omega}_*(\theta_t)$\,.

Using the update rules of~$(\bar{\omega}_t)$, $(\omega_t)$ and developing the squared norm gives:
\begin{align}
  \label{eq:barnu}
\|\bar{\nu}_{t+1}\|^2 &= \|\bar{\omega}_t + \xi_t (\omega_{t+1} - \bar{\omega}_t) - \bar{\omega}_*(\theta_{t+1})\|^2\nonumber\\
&= \|\bar{\nu}_t + \xi_t ( \omega_t + \beta_t g(\tilde{x}_t,\bar{\omega}_t,\omega_t) - \bar{\omega}_t)
+ \bar{\omega}_*(\theta_t) - \bar{\omega}_*(\theta_{t+1}) \|^2\nonumber\\
&= \| \bar{\nu}_t + \left( \xi_t (\nu_t + \beta_t g(\tilde{x}_t,\bar{\omega}_t,\omega_t) + \omega_*(\theta_t,\bar{\omega}_t) - \bar{\omega}_t) + \bar{\omega}_*(\theta_t) - \bar{\omega}_*(\theta_{t+1})\right) \|^2\nonumber\\
&= \|\bar{\nu}_t\|^2 + 2 \ps{\bar{\nu}_t,\xi_t (\nu_t + \beta_t g(\tilde{x}_t,\bar{\omega}_t,\omega_t) + \omega_*(\theta_t,\bar{\omega}_t) - \bar{\omega}_t) + \bar{\omega}_*(\theta_t) - \bar{\omega}_*(\theta_{t+1})}\nonumber\\
&+ \|\xi_t (\nu_t + \beta_t g(\tilde{x}_t,\bar{\omega}_t,\omega_t) + \omega_*(\theta_t,\bar{\omega}_t) - \bar{\omega}_t) + \bar{\omega}_*(\theta_t) - \bar{\omega}_*(\theta_{t+1})\|^2 \,.
\end{align}

Since the sequences~$(\nu_t), (\bar{\omega}_t)$ and the functions~$g, \omega_*$ are bounded and the function~$\bar{\omega}_*$ is Lipschitz continuous, the last squared norm term can be bounded by:
$
C (\xi_t^2 \beta_t^2 + \xi_t^2 + \frac{\alpha_t^2}{(1-\gamma)^2})\,.
$

We now control the scalar product in Eq.~\eqref{eq:barnu}. We  decompose this term into four different terms:

\begin{enumerate}[{\sl (a)},leftmargin=*]
  \item Using Assumption~\ref{hyp:eigenval}, it holds that:
  \[
  2 \xi_t \ps{\bar{\nu}_t,\omega_*(\theta_t,\bar{\omega}_t) - \bar{\omega}_t} = - 2 \xi_t \ps{\bar{\nu}_t, \bar{G}(\theta_t)^{-1} G(\theta_t) \bar{\nu}_t} \leq -2 \zeta \xi_t \|\bar{\nu}_t\|^2\,.
  \]
  \item The boundedness of the function~$g$ implies that:
  \[
  2 \xi_t \beta_t \ps{\bar{\nu}_t,g(\tilde{x}_t, \bar{\omega}_t, \omega_t)} \leq C \xi_t \beta_t \|\bar{\nu}_t\|\,.
  \]

  \item Applying the Cauchy-Schwarz inequality gives:
  \[
  2 \xi_t \ps{\bar{\nu}_t,\nu_t} \leq 2 \xi_t \|\bar{\nu}_t\| \cdot \|\nu_t\|\,.
  \]

  \item Since $\bar{\omega}_*$ is Lipschitz continuous, we can write:
  \[
  2 \ps{\bar{\nu}_t,\bar{\omega}_*(\theta_t) - \bar{\omega}_*(\theta_{t+1})} \leq C \frac{\alpha_t}{1-\gamma} \|\bar{\nu}_t\|\,.
  \]
\end{enumerate}

Collecting the bounds from items~(a) to~(d) and incorporating them into Eq.~\eqref{eq:barnu}, we obtain:
\begin{equation}
  \label{eq:barnu_ineq}
\|\bar{\nu}_{t+1}\|^2 \leq (1-2\zeta\xi_t)\|\bar{\nu}_t\|^2
            + C \left(\xi_t \beta_t + \frac{\alpha_t}{1-\gamma}\right) \|\bar{\nu}_t\|
            + 2 \xi_t \|\bar{\nu}_t\| \cdot \|\nu_t\|
            + C \left(\xi_t^2 \beta_t^2 + \xi_t^2 + \frac{\alpha_t^2}{(1-\gamma)^2}\right)\,.
\end{equation}
Rearranging Ineq.~\eqref{eq:barnu_ineq} leads to:
\begin{equation}
  \label{eq:barnu_ineq2}
2 \zeta \|\bar{\nu}_t\|^2
\leq \frac{1}{\xi_t}(\|\bar{\nu}_t\|^2 - \|\bar{\nu}_{t+1}\|^2)
+ C \left(\beta_t + \frac{\alpha_t}{(1-\gamma)\xi_t}\right) \|\bar{\nu}_t\|
+ 2 \|\bar{\nu}_t\|\cdot \|\nu_t\|
+ C \left(\xi_t \beta_t^2 + \xi_t + \frac{\alpha_t^2}{(1-\gamma)^2 \xi_t}\right)\,.
\end{equation}
Summing this inequality for $t$ between $1$ and~$T$ and taking total expectation yield:
\begin{equation}
  \label{eq:ineq_rate}
\frac{2 \zeta}{T}\sum_{t=1}^T \bE[\|\bar{\nu}_t\|^2] \leq \Sigma_1(T) + \Sigma_2(T) + \Sigma_3(T) + \Sigma_4(T)\,,
\end{equation}
where
\begin{align}
\Sigma_1(T) &\eqdef \frac{1}{T}\sum_{t=1}^T \frac{1}{\xi_t}(\bE[\|\bar{\nu}_t\|^2] - \bE[\|\bar{\nu}_{t+1}\|^2])\,,\\
\Sigma_2(T) &\eqdef \frac{C}{T}\sum_{t=1}^T \left(\beta_t + \frac{\alpha_t}{(1-\gamma)\xi_t}\right) \bE[\|\bar{\nu}_t\|] \,,\\
\Sigma_3(T) &\eqdef \frac{2}{T}\sum_{t=1}^T \bE[\|\bar{\nu}_t\|\cdot \|\nu_t\|] \,,\\
\Sigma_4(T) &\eqdef \frac{C}{T}\sum_{t=1}^T \left(\xi_t \beta_t^2 + \xi_t + \frac{\alpha_t^2}{(1-\gamma)^2\xi_t}\right)\,.
\end{align}

Similarly to Sec.~\ref{sec:1st-error-term}, we control each one of the terms $\Sigma_i, i=1,2,3,4$ successively.

\begin{enumerate}[{\sl (i)},leftmargin=*]
\item First, using the boundedness of $(\bar{\nu}_t)$, we estimate~$\Sigma_1$ as follows:
\[
\Sigma_1(T) = \frac{1}{T} \left[ \sum_{t=1}^T \left(\frac{1}{\xi_t} - \frac{1}{\xi_{t-1}}\right) \bE[\|\bar{\nu}_t\|^2] + \frac{1}{\xi_0}\bE[\|\bar {\nu}_1\|^2] - \frac{1}{\xi_T}\bE[\|\bar {\nu}_{T+1}\|^2] \right]
\leq \frac{C}{T\xi_T}
= \mO(T^{\xi-1})\,.
\]

\item Cauchy-Schwarz inequality implies:
\begin{align*}
\Sigma_2(T) &\leq \frac{C}{T} \sqrt{\sum_{t=1}^T \left( \beta_t + \frac{\alpha_t}{(1-\gamma)\xi_t} \right)^2}\sqrt{\sum_{t=1}^T \bE[\|\bar{\nu}_t\|^2]}\\
&\leq C \sqrt{\frac{1}{T}\sum_{t=1}^T \left( \beta_t^2 + \left(\frac{\alpha_t}{(1-\gamma)\xi_t}\right)^2 \right)}\sqrt{\frac{1}{T}\sum_{t=1}^T \bE[\|\bar{\nu}_t\|^2]}\,.
\end{align*}
Moreover,
\begin{align*}
\frac{1}{T}\sum_{t=1}^T \left( \beta_t^2 + \left(\frac{\alpha_t}{(1-\gamma)\xi_t}\right)^2 \right)
&\leq \frac{1}{T}\left( \frac{(T+1)^{1-2\beta}}{1-2\beta} + \frac{(T+1)^{1-2(\alpha-\xi)}}{(1-\gamma)^2 (1-2(\alpha-\xi))} \right)\\
&= \mO(T^{-2\beta}) + \mO\left(\frac{T^{-2(\alpha - \xi)}}{(1-\gamma)^2}\right)\,.
\end{align*}

\item Invoking the Cauchy-Schwarz inequality again yields:
\[
\Sigma_3(T) \leq 2\sqrt{\frac{1}{T}\sum_{t=1}^T \bE[\|\bar{\nu}_t\|^2]} \sqrt{\frac{1}{T}\sum_{t=1}^T \bE[\|\nu_t\|^2]}
\]

\item Similarly to item~(ii), we obtain
\[
\Sigma_4(T)
= \mO(T^{-\xi-2\beta})+ \mO(T^{-\xi}) + \mO\left(\frac{T^{\xi-2\alpha}}{(1-\gamma)^2}\right)\,.
\]
\end{enumerate}

Define for every~$T >0$ the following quantities:
\begin{align}
W(T)&\eqdef  \frac{1}{T} \sum_{t=1}^T \bE[\|\nu_t\|^2]\,,\\
X(T)&\eqdef \frac{1}{T} \sum_{t=1}^T \bE[\|\bar{\nu}_t\|^2]\,,\\
Y(T)&\eqdef \frac{1}{T}\sum_{t=1}^T \left( \beta_t^2 + \left(\frac{\alpha_t}{(1-\gamma)\xi_t}\right)^2 \right)\,,\\
Z(T)&\eqdef \Sigma_1(T) + \Sigma_4(T)\,.
\end{align}

It follows from items (i) to (iv) and Sec.~\ref{sec:1st-error-term} (for the last estimate) that
\begin{align}
Y(T)&= \mO(T^{-2\beta}) + \mO\left(\frac{T^{-2(\alpha - \xi)}}{(1-\gamma)^2}\right)\,,\\
Z(T)&= \mO(T^{\xi-1}) + \mO(T^{-\xi-2\beta})+ \mO(T^{-\xi}) + \mO\left(\frac{T^{\xi-2\alpha}}{(1-\gamma)^2}\right)\,,\\
W(T) &= \mO(T^{\beta -1}) + \mO\left(\frac{|\mA|}{1-\gamma} \ln(T)\, T^{-\beta}\right) + \mO\left(\frac{T^{2(\beta- \alpha)}}{(1-\gamma)^2}\right) + \mO(T^{2(\beta- \xi)})\,.
\end{align}

Eq.~\eqref{eq:ineq_rate} can be written:
\[
2 \zeta  X(T) \leq C \left(\sqrt{Y(T)} + \sqrt{W(T)}\right) \sqrt{X(T)} + Z(T)\,.
\]
Solving this inequality implies:
\begin{equation}
  \label{eq:mO_X}
X(T) = \mO(Y(T) + W(T) + Z(T))\,.
\end{equation}
Since $0 < \beta < \xi < \alpha < 1$, we obtain:
\begin{equation}
  \label{eq:mO_X_rate}
X(T) = \mO(T^{\xi-1}) + \mO\left(\frac{|\mA|}{1-\gamma} \ln(T)\, T^{-\beta}\right)
        + \mO\left(\frac{T^{-2(\alpha - \xi)}}{(1-\gamma)^2}\right)  + \mO(T^{-2(\xi-\beta)})\,.
\end{equation}

\subsubsection{End of Proof of Th.~\ref{th:critic_rate}}

We conclude our finite-time analysis of the critic by combining both previous sections (\ref{sec:1st-error-term} and~\ref{sec:2nd-error-term}):
\begin{align}
\frac{1}{T} \sum_{t=1}^T \bE[\|\omega_t- \bar{\omega}_*(\theta_t)\|^2]
&= \frac{1}{T} \sum_{t=1}^T \bE[\|\nu_t + \omega_*(\theta_t,\bar{\omega}_t) - \bar{\omega}_*(\theta_t)\|^2]\nonumber\\
&= \frac{1}{T} \sum_{t=1}^T \bE[\|\nu_t + \omega_*(\theta_t,\bar{\omega}_t) - \omega_*(\theta_t,\bar{\omega}_*(\theta_t))\|^2]\nonumber\\
&\leq 2 W(T) + C X(T)\nonumber\\
&= \mO(X(T))\nonumber\\
&= \mO(T^{\xi-1}) + \mO\left(\frac{|\mA|}{1-\gamma} \ln(T)\, T^{-\beta}\right)
+ \mO\left(\frac{T^{-2(\alpha - \xi)}}{(1-\gamma)^2}\right)  + \mO(T^{-2(\xi-\beta)})\,,
\end{align}
where the second equality follows from using the identity $w_*(\theta,\bar{\omega}_*(\theta)) = \bar{\omega}_*(\theta)$ for every~$\theta \in \bR^d$, the inequality stems from using the classical inequality $\|a+b\|^2 \leq 2 (\|a\|^2+ \|b\|^2)$ together with the fact that~$\omega_*$ is Lipschitz continuous, the penultimate equality is a consequence of Eq.~\eqref{eq:mO_X} and the last equality is the result of the previous section (see Eq.~\eqref{eq:mO_X_rate}).

\subsection{Proof of Th.~\ref{th:actor_rate}: finite-time analysis of the actor}

Recall the notation $\tilde{x}_t \eqdef (\tilde{S}_t,\tilde{A}_t,S_{t+1})$. In this section,
we overload this notation with the reward sequence~$(R_t)$, i.e., $\tilde{x}_t \eqdef (\tilde{S}_t,\tilde{A}_t,S_{t+1}, R_{t+1})$\,.
Let us fix some additional convenient notations. Define for every~$\tilde x = (\tilde s, \tilde a, s, r) \in \mS \times \mA \times \mS \times [-U_R,U_R]$,\, and every~$\omega \in \bR^m, \theta \in \bR^d$:
\begin{align}
\hat{\delta}(\tilde{x},\omega) &\eqdef r + \gamma \phi(s)^T \omega - \phi(\tilde s)^T \omega\\
\delta(\tilde x,\theta) &= r + \gamma V_{\pi_\theta}(s) - V_{\pi_\theta}(\tilde s)\,.
\end{align}

Note that the TD error~$\delta_{t+1}$ used in Algorithm~\ref{algo} coincides with~$\hat{\delta}(\tilde{x}_t,\omega_t)$.

Recall that $\theta \mapsto \nabla J(\theta)$ and $\theta \mapsto V_{\pi_\theta}(s)$ (for every~$s \in \mS$) are Lipschitz continuous. Throughout the proof, $L_{\nabla J}$ (resp. $L_V$) stands for the Lipschitz constant of $\theta \mapsto \nabla J(\theta)$ (resp. $\theta \mapsto V_{\pi_\theta}(s)$ for every~$s \in \mS$) and $C_{\nabla J}$ (resp.~$C_V$) denotes the upperbound of $\theta \mapsto \|\nabla J(\theta)\|$ (resp. $\theta \mapsto V_{\pi_\theta}(s)$ for every~$s \in \mS$).
Since the function~$\nabla J$ is $L_{\nabla J}$-Lipschitz continuous, a classical Taylor inequality combined with the update rule of~$(\theta_t)$ yields:
\begin{equation}
  \label{eq:taylor1}
J(\theta_{t+1}) \geq J(\theta_t) + \frac{\alpha_t}{1-\gamma} \ps{\nabla J(\theta_t),\hat{\delta}(\tilde{x}_t,\omega_t) \psi_{\theta_t}(\tilde{S}_t,\tilde{A}_t)} - \frac{L_{\nabla J}}{2} \frac{\alpha_t^2}{(1-\gamma)^2} \|\hat{\delta}(\tilde{x}_t,\omega_t) \psi_{\theta_t}(\tilde{S}_t,\tilde{A}_t)\|^2\,.
\end{equation}
Recalling that $\theta \mapsto \psi_\theta(s,a)$ is bounded by Assumption~\ref{hyp:grad_pi}-\ref{hyp:psi_theta}, $(R_t)$ and~$(\omega_t)$ are bounded (see Assumption~\ref{hyp:stability}) and $\mS, \mA$ are finite, we obtain from Eq.~\eqref{eq:taylor1} that there exists a constant~$C$ s.t.:
\begin{equation}
\label{eq:taylor2}
J(\theta_{t+1}) \geq J(\theta_t) + \frac{\alpha_t}{1-\gamma} \ps{\nabla J(\theta_t),\hat{\delta}(\tilde{x}_t,\omega_t) \psi_{\theta_t}(\tilde{S}_t,\tilde{A}_t)} - C L_{\nabla J} \frac{\alpha_t^2}{(1-\gamma)^2}\,.
\end{equation}

Now, we decompose the TD error by introducing both the moving target~$\bar{\omega}_*(\theta_t)$ and the TD error~$\delta(\tilde{x}_t,\theta_t)$ associated to the true value function~$V_{\pi_{\theta_t}}$:
\begin{equation}
  \label{eq:decomp_delta}
\hat{\delta}(\tilde{x}_t,\omega_t) =
[\hat{\delta}(\tilde{x}_t,\omega_t) - \hat{\delta}(\tilde{x}_t,\bar{\omega}_*(\theta_t))]
+ [\hat{\delta}(\tilde{x}_t,\bar{\omega}_*(\theta_t))
- \delta(\tilde{x}_t,\theta_t)]
+ \delta(\tilde{x}_t,\theta_t)\,.
\end{equation}
Incorporating this decomposition~\eqref{eq:decomp_delta} into Eq.~\eqref{eq:taylor2} gives:
\begin{align}
  \label{eq:taylor}
J(\theta_{t+1}) &\geq J(\theta_t)
+ \frac{\alpha_t}{1-\gamma} \ps{\nabla J(\theta_t),(\hat{\delta}(\tilde{x}_t,\omega_t) - \hat{\delta}(\tilde{x}_t,\bar{\omega}_*(\theta_t))) \psi_{\theta_t}(\tilde{S}_t,\tilde{A}_t)}\nonumber\\
&+ \frac{\alpha_t}{1-\gamma} \ps{\nabla J(\theta_t),(\hat{\delta}(\tilde{x}_t,\bar{\omega}_*(\theta_t))
- \delta(\tilde{x}_t,\theta_t)) \psi_{\theta_t}(\tilde{S}_t,\tilde{A}_t)}\nonumber\\
&+ \frac{\alpha_t}{1-\gamma} \ps{\nabla J(\theta_t),\delta(\tilde{x}_t,\theta_t) \psi_{\theta_t}(\tilde{S}_t,\tilde{A}_t) - \nabla J(\theta_t)}
+ \frac{\alpha_t}{1-\gamma} \|\nabla J(\theta_t)\|^2
- C L_{\nabla J} \frac{\alpha_t^2}{(1-\gamma)^2}\,.
\end{align}

In Eq.~\eqref{eq:taylor}, the first inner product corresponds to the bias introduced by the critic. The second one represents the linear FA error and the third translates the Markovian noise.
Our task now is to control each one of these error terms in Eq.~\eqref{eq:taylor}.

For the first term, observing that~$\hat{\delta}(\tilde{x}_t,\omega_t) - \hat{\delta}(\tilde{x}_t,\bar{\omega}_*(\theta_t)) = (\gamma \phi(S_{t+1}) -  \phi(\tilde{S}_t))^T (\omega_t - \bar{\omega}_*(\theta_t))$, the Cauchy-Schwarz inequality leads to:
\begin{equation}
  \label{eq:taylor_term1}
  \bE[\ps{\nabla J(\theta_t),\hat{\delta}(\tilde{x}_t,\omega_t) - \hat{\delta}(\tilde{x}_t,\bar{\omega}_*(\theta_t))\psi_{\theta_t}(\tilde{S}_t, \tilde{A}_t)}]
  \geq - C \sqrt{\bE[\|\nabla J(\theta_t)\|^2]}\sqrt{\bE[\|\omega_t - \bar{\omega}_*(\theta_t)\|^2]}\,.
\end{equation}

Then, we control each one of the second and third terms in Eq.~\eqref{eq:taylor} in the following sections successively.

\subsubsection{Control of the Markovian bias term}

We introduce a specific convenient notation for the second term, for every~$\tilde x = (\tilde s, \tilde a, s, r) \in \mS \times \mA \times \mS \times [-U_R,U_R]$,\, and every~$\theta \in \bR^d$:
\[
\Gamma(\tilde{x},\theta) \eqdef \ps{\nabla J(\theta),\delta(\tilde{x},\theta) \psi_{\theta}(\tilde{s},\tilde{a}) - \nabla J(\theta)}\,.
\]
Recall from Sec.~\ref{sec:1st-error-term} the auxiliary Markov chain~$(\check{x}_t)$, the Markov chain~$(\bar{x}_t)$ induced by the stationary distribution and the mixing time~$\tau$ defined in Eq.~\eqref{eq:def_tau}.

Similarly to Sec.~\ref{sec:1st-error-term}, we introduce the following decomposition:
\begin{multline}
  \label{eq:decomp_Gamma}
  \bE[\Gamma(\tilde{x}_t,\theta_t)]
  = \bE[\Gamma(\tilde{x}_t,\theta_t) - \Gamma(\tilde{x}_t,\theta_{t-\tau})]
  + \bE[\Gamma(\tilde{x}_t,\theta_{t-\tau}) -
  \Gamma(\check{x}_t,\theta_{t-\tau})]\\
  + \bE[\Gamma(\check{x}_t,\theta_{t-\tau}) -
  \Gamma(\bar{x}_t,\theta_{t-\tau})]
  + \bE[\Gamma(\bar{x}_t,\theta_{t-\tau})]\,.
\end{multline}

We address each term of this decomposition successively.
\begin{enumerate}[{\sl (a)},leftmargin=*]
  \item For this first term, we write:
  \begin{align*}
    \label{eq:Gamma1}
    \Gamma(\tilde{x}_t,\theta_t) - \Gamma(\tilde{x}_t,\theta_{t-\tau})
    &=  \ps{\nabla J(\theta_t) - \nabla J(\theta_{t-\tau}),\delta(\tilde{x}_t,\theta_t) \psi_{\theta_t}(\tilde{S}_t,\tilde{A}_t) - \nabla J(\theta_t)}\nonumber\\
    &+ \ps{\nabla J(\theta_{t-\tau}), (\delta(\tilde{x}_t,\theta_t) - \delta(\tilde{x}_t,\theta_{t-\tau})) \psi_{\theta_t}(\tilde{S}_t,\tilde{A}_t)}\nonumber\\
    &+ \ps{\nabla J(\theta_{t-\tau}),\delta(\tilde{x}_t,\theta_{t-\tau}) (\psi_{\theta_t}(\tilde{S}_t,\tilde{A}_t) - \psi_{\theta_{t-\tau}}(\tilde{S}_t,\tilde{A}_t))}\nonumber\\
    &+ \ps{\nabla J(\theta_{t-\tau}),\nabla J(\theta_{t-\tau}) - \nabla J(\theta_t)}\,.
  \end{align*}
  Moreover, note that:
  \begin{equation*}
    \label{eq:diff_delta}
  \delta(\tilde{x}_t,\theta_t) - \delta(\tilde{x}_t,\theta_{t-\tau}) = \gamma ( V_{\pi_{\theta_t}}(S_{t+1}) - V_{\pi_{\theta_{t-\tau}}}(S_{t+1})) + V_{\pi_{\theta_{t-\tau}}}(\tilde{S}_t) - V_{\pi_{\theta_t}}(\tilde{S}_t)\,.
\end{equation*}

Remark that $\nabla J, \theta \mapsto \psi_\theta$ and~$\theta \mapsto V_{\pi_\theta}$ are bounded functions under Assumption~\ref{hyp:grad_pi}\,.
Since~$\nabla J, V_{\pi_\theta}, \psi_\theta$ are in addition Lipschitz continuous as functions of~$\theta$ (see, for e.g., \cite[Lem.~3]{shen-zhang-hong-chen20} for a proof for~$V_{\pi_\theta}$) under Assumption~\ref{hyp:grad_pi}\,, one can show after tedious inequalities that:
\begin{align}
\label{eq:control_Gamma1}
\nonumber |\Gamma(\tilde{x}_t,\theta_t) - \Gamma(\tilde{x}_t,\theta_{t-\tau})|
&\leq (L_{\nabla J} (C(1+C_V) + C_{\nabla J}) + CC_{\nabla J} L_V + C(1+C_V)C_{\nabla J} + C_{\nabla J} L_{\nabla J}) \|\theta_t -\theta_{t-\tau}\| \\
&\leq C C_{1-\gamma}\|\theta_t - \theta_{t-\tau}\|  \,,
\end{align}
where~$C_{1-\gamma} \eqdef \max(L_{\nabla J} C_V, L_{\nabla J} C_{\nabla J}, L_V C_{\nabla J}, C_V C_{\nabla J})$\,. Note here that the last notation highlights that the constant depends on~$1-\gamma$ due to the dependence on~$1-\gamma$ of the constants defining~$C_{1-\gamma}$. We will explicit this dependence later on in the proof.

  \item For the second term, we have:
  \begin{align}
    \label{eq:Gamma2}
    &|\bE[\Gamma(\tilde{x}_t,\theta_{t-\tau}) -
    \Gamma(\check{x}_t,\theta_{t-\tau})]|\nonumber\\
    &=
    |\bE[\ps{\nabla J(\theta_{t-\tau}),\delta(\tilde{x}_t,\theta_{t-\tau})\psi_{\theta_{t-\tau}}(\tilde{S}_t,\tilde{A}_t) - \delta(\check{x}_t,\theta_{t-\tau})\psi_{\theta_{t-\tau}}(\check{S}_t,\check{A}_t) }]|\nonumber\\
    &=|\bE[\ps{\nabla J(\theta_{t-\tau}),\delta(\tilde{x}_t,\theta_{t-\tau})\psi_{\theta_{t-\tau}}(\tilde{S}_t,\tilde{A}_t) - \delta(\check{x}_t,\theta_{t-\tau})\psi_{\theta_{t-\tau}}(\check{S}_t,\check{A}_t) }|\tilde{S}_{t-\tau+1},\theta_{t-\tau}]|\nonumber\\
    &\leq C C_V C_{\nabla J}\bE[d_{TV}(\bP(\tilde{x}_t \in \cdot |\tilde{S}_{t-\tau+1},\theta_{t-\tau}),\bP(\check{x}_t \in \cdot |\tilde{S}_{t-\tau+1},\theta_{t-\tau}))]\nonumber\\
    &\leq C C_V C_{\nabla J} |\mA| \sum_{i=t-\tau}^t \bE[\|\theta_i -\theta_{t-\tau}\|]\,.
  \end{align}
Here, the first inequality is a consequence of the definition of the total variation distance whereas the second inequality follows from applying \cite[Lem.~B.2]{wu-zha-xu-gu20}. Indeed, using this last lemma, to show the last inequality, it is sufficient to write:
\begin{align*}
&d_{TV}(\bP(\tilde{x}_t \in \cdot |\tilde{S}_{t-\tau+1},\theta_{t-\tau}),\bP(\check{x}_t \in \cdot |\tilde{S}_{t-\tau+1},\theta_{t-\tau}))\nonumber\\
&= d_{TV}(\bP((\tilde{S}_t,\tilde{A}_t) \in \cdot |\tilde{S}_{t-\tau+1},\theta_{t-\tau}),\bP((\check{S}_t,\check{A}_t) \in \cdot |\tilde{S}_{t-\tau+1},\theta_{t-\tau}))\nonumber\\
&\leq d_{TV}(\bP(\tilde{S}_t \in \cdot |\tilde{S}_{t-\tau+1},\theta_{t-\tau}),\bP(\check{S}_t \in \cdot |\tilde{S}_{t-\tau+1},\theta_{t-\tau})) + \frac{1}{2} |\mA| L_\pi \bE[\|\theta_t - \theta_{t- \tau}\|]\,.
\end{align*}
Iterating this inequality gives the desired result of Eq.~\eqref{eq:Gamma2}\,. We conclude from this item that:
\begin{equation*}
\label{eq:control_Gamma2}
\bE[\Gamma(\tilde{x}_t,\theta_{t-\tau}) -
\Gamma(\check{x}_t,\theta_{t-\tau})]
\geq - C C_V C_{\nabla J} |\mA|\sum_{i=t-\tau}^t \bE[\|\theta_i -\theta_{t-\tau}\|]\,.
\end{equation*}

  \item Regarding the third term, similarly to item~(b), we can write:
  \begin{align}
    \label{eq:control_Gamma3}
    \bE[\Gamma(\check{x}_t,\theta_{t-\tau}) -
    \Gamma(\bar{x}_t,\theta_{t-\tau})]
    &\geq - C C_V C_{\nabla J} \bE[d_{TV}(\bP(\check{x}_t \in \cdot|\tilde{S}_{t-\tau+1},\theta_{t-\tau}),\bP(\bar{x}_t \in \cdot|\tilde{S}_{t-\tau+1},\theta_{t-\tau}))]\nonumber\\
    &= -C C_V C_{\nabla J} \bE[d_{TV}(\bP(\check{x}_t \in \cdot|\tilde{S}_{t-\tau+1},\theta_{t-\tau}),d_{\rho,\theta_{t-\tau}} \otimes \pi_{\theta_{t-\tau}} \otimes p)]\nonumber\\
    &= -C C_V C_{\nabla J} \bE[d_{TV}(\bP(\check{S}_t \in \cdot|\tilde{S}_{t-\tau+1},\theta_{t-\tau}),d_{\rho,\theta_{t-\tau}})]\nonumber\\
    &\geq -C C_V C_{\nabla J}\sigma^{\tau-1}\,,
  \end{align}
where the equalities follow from the definitions of~$\check{x}_t, \bar{x}_t$ and the last inequality stems from Assumption~\ref{hyp:geom_mix_mc}\,.

  \item Since the Markov chain~$\bar{x}_t$ is built s.t. $\bar{S}_t \sim d_{\rho,\theta_{t-\tau}}, \bar{A}_t \sim \pi_{\theta_{t-\tau}}, S_{t+1} \sim p(\cdot|\bar{S}_t,\bar{A}_t)$, one can see that
  $\bE[\Gamma(\bar{x}_t, \theta_{t-\tau})] = 0$\,.
\end{enumerate}

We conlude this section from Eq.~\eqref{eq:decomp_Gamma} by collecting Eqs.~\eqref{eq:control_Gamma1} to~\eqref{eq:control_Gamma3} (items (a) to (d)) to obtain:
\begin{align}
\label{eq:control_Gamma}
\bE[\Gamma(\tilde{x}_t,\theta_t)]
&\geq  - C C_{1-\gamma} \bE[\|\theta_t - \theta_{t-\tau}\|]
- C C_V C_{\nabla J} \sum_{i=t-\tau+1}^t \bE[\|\theta_i - \theta_{t-\tau}\|]
- C C_V C_{\nabla J} \sigma^{\tau-1}\nonumber\\
&\geq - C C_{1-\gamma} \sum_{i= t-\tau+1}^t \bE[\|\theta_i - \theta_{i-1}\|] - C C_V C_{\nabla J} \sum_{i= t-\tau+1}^t \sum_{j=t-\tau+1}^i \bE[\|\theta_j - \theta_{j-1}\|]
- C C_V C_{\nabla J}\sigma^{\tau-1} \nonumber\\
&\geq - C C_{1-\gamma} \sum_{i= t-\tau+1}^t \bE[\|\theta_i - \theta_{i-1}\|] - C C_V C_{\nabla J} \sum_{i= t-\tau+1}^t \sum_{j=t-\tau+1}^t \bE[\|\theta_j - \theta_{j-1}\|]
- C C_V C_{\nabla J}\sigma^{\tau-1} \nonumber\\
&\geq - C (C_{1-\gamma} + C_V C_{\nabla J} \tau) \sum_{i= t-\tau+1}^t \bE[\|\theta_i - \theta_{i-1}\|] - C C_V C_{\nabla J} \sigma^{\tau-1} \nonumber\\
&\geq  - C \left( (C_{1-\gamma}\, \tau + C_V C_{\nabla J} \tau^2) \frac{\alpha_{t-\tau}}{1-\gamma} + C_V C_{\nabla J} \alpha_T\right)\,,
\end{align}
where the last inequality uses the definition of the mixing time~$\tau$ and the fact that the sequence~$(\alpha_t)$ is nonincreasing.

\subsubsection{Control of the linear FA error term}

Recall that $\theta \mapsto \psi_\theta$ is Lipschitz continuous, $\nabla J$ is bounded and remark that the quantity~$\hat{\delta}(\tilde{x}_t,\bar{\omega}_*(\theta_t))
- \delta(\tilde{x}_t,\theta_t)$ is bounded.
Therefore, using the Cauchy-Schwarz inequality, we have:
\begin{align}
  \label{eq:control-linear-fa-error}
  &\bE[\ps{\nabla J(\theta_t),(\hat{\delta}(\tilde{x}_t,\bar{\omega}_*(\theta_t))
  - \delta(\tilde{x}_t,\theta_t)) \psi_{\theta_t}(\tilde{S}_t,\tilde{A}_t)}]\nonumber\\
  &=   \bE[\ps{\nabla J(\theta_t),(\hat{\delta}(\tilde{x}_t,\bar{\omega}_*(\theta_t))
    - \delta(\tilde{x}_t,\theta_t)) (\psi_{\theta_t}(\tilde{S}_t,\tilde{A}_t)-\psi_{\theta_{t-\tau}}(\tilde{S}_t,\tilde{A}_t))}]\nonumber\\
  &+ \bE[\ps{\nabla J(\theta_t),(\hat{\delta}(\tilde{x}_t,\bar{\omega}_*(\theta_t))
      - \delta(\tilde{x}_t,\theta_t)) \psi_{\theta_{t-\tau}}(\tilde{S}_t,\tilde{A}_t)}]\nonumber\\
  &\geq -C (1+ C_V) C_{\nabla J} \bE[\|\theta_t - \theta_{t-\tau}\|]
  + \bE[\ps{\nabla J(\theta_t),( \hat{\delta}(\tilde{x}_t,\bar{\omega}_*(\theta_t)) - \delta(\tilde{x}_t,\theta_t) )\psi_{\theta_{t-\tau}}(\tilde{S}_t,\tilde{A}_t)}]\,.
\end{align}

Let us introduce for every~$\tilde x = (\tilde s, \tilde a, s, r) \in \mS \times \mA \times \mS \times [-U_R,U_R]$,\, and every~$\theta \in \bR^d$ the shorthand notation:
\[
\Delta(\tilde{x},\theta) \eqdef \ps{\nabla J(\theta),(\hat{\delta}(\tilde{x},\bar{\omega}_*(\theta))- \delta(\tilde{x},\theta)) \psi_{\theta_{t-\tau}}(\tilde{S}_t ,\tilde{A}_t)}\,.
\]
Note here that the term~$\psi_{\theta_{t-\tau}}(\tilde{S}_t ,\tilde{A}_t)$ in the notation above is fixed in adequacy with Eq.~\eqref{eq:control-linear-fa-error}. The following decomposition holds:
\begin{multline}
\Delta(\tilde{x}_t,\theta_t) = (\Delta(\tilde{x}_t,\theta_t) - \Delta(\tilde{x}_t,\theta_{t-\tau}))
+ (\Delta(\tilde{x}_t,\theta_{t-\tau}) - \Delta(\check{x}_t,\theta_{t-\tau}))\\
+ (\Delta(\check{x}_t,\theta_{t-\tau}) - \Delta(\bar{x}_t,\theta_{t-\tau}))
+ \Delta(\bar{x}_t,\theta_{t-\tau})\,.
\end{multline}

Similar derivations to the previous section allow us to control each one of the error terms.
\begin{enumerate}[{\sl (i)},leftmargin=*]
\item Using that the mappings $\nabla J,\, \theta \mapsto V_{\pi_\theta}(s)$ (for every~$s \in \mS$) and $\theta \mapsto \bar{\omega}_*(\theta)$ are
$L_{\nabla J}$(resp. $L_V,L_{\bar{\omega}_*}$)-Lipschitz continuous, we obtain:
\begin{equation}
\label{eq:Delta1}
\nonumber \Delta(\tilde{x}_t,\theta_t) - \Delta(\tilde{x}_t,\theta_{t-\tau})
\geq -C \tilde{C}_{1-\gamma} \|\theta_t - \theta_{t-\tau}\|\,,
\end{equation}
where~$\tilde{C}_{1-\gamma} \eqdef L_{\nabla J}(1+C_V) + C_{\nabla J} (L_V + L_{\bar{\omega}_*})$\,.

Using similar manipulations to the previous section, we get:
\item
\begin{equation}
\label{eq:Delta2}
\bE[\Delta(\tilde{x}_t,\theta_{t-\tau}) - \Delta(\check{x}_t,\theta_{t-\tau})]
\geq - C C_{\nabla J} (1+C_V)|\mA| \sum_{i=t-\tau}^t \bE[\|\theta_i - \theta_{t-\tau}\|]\,.
\end{equation}

\item
\begin{equation}
\label{eq:Delta3}
\bE[\Delta(\check{x}_t,\theta_{t-\tau}) - \Delta(\bar{x}_t,\theta_{t-\tau})] \geq -C C_{\nabla J} (1+C_V) \sigma^{\tau-1}\,.
\end{equation}

\item For the last term, we can write:
\begin{equation}
  \label{eq:Delta4}
\bE[\Delta(\bar{x}_t,\theta_{t-\tau})|\theta_{t-\tau}]
\geq - C \|\nabla J(\theta_{t-\tau})\| \cdot
\bE[|\hat{\delta}(\bar{x}_t,\bar{\omega}_*(\theta_{t-\tau})) - \delta(\bar{x}_t, \theta_{t-\tau})| |\theta_{t-\tau}]\,.
\end{equation}
Then, recall that $\bar{x}_t = (\bar{S}_t, \bar{A}_t, S_{t+1})$ where~$S_{t+1} \sim p(\cdot|\bar{S}_t, \bar{A}_t)$ and observe that:
\begin{multline}
\hat{\delta}(\bar{x}_t,\bar{\omega}_*(\theta_{t-\tau})) - \delta(\bar{x}_t, \theta_{t-\tau}) =
\gamma (\phi(S_{t+1})^T \bar{\omega}_*(\theta_{t-\tau}) - V_{\pi_{\theta_{t-\tau}}}(S_{t+1}))\\
+ (V_{\pi_{\theta_{t-\tau}}}(\bar{S}_t) - \phi(\bar{S}_t)^T \bar{\omega}_*(\theta_{t-\tau}))\,.
\end{multline}

Recalling that~$\tilde p = \gamma p + (1-\gamma) \rho$ and using Assumption~\ref{hyp:eps_fa}\,, one can then easily show that:
\[
\bE[|\hat{\delta}(\bar{x}_t,\bar{\omega}_*(\theta_{t-\tau})) - \delta(\bar{x}_t, \theta_{t-\tau})| |\theta_{t-\tau}] \leq  C \epsilon_{\text{FA}}\,.
\]
As a consequence, noticing that~$\nabla J$ is bounded, we obtain from~Eq.~\eqref{eq:Delta4}:
\[
\bE [\Delta(\bar{x}_t,\theta_{t-\tau})]
\geq - C C_{\nabla J} \epsilon_{\text{FA}}\,.
\]
\end{enumerate}
Combining items~(i) to~(iv) with the boundedness of the function~$\nabla J$, we conclude from this section that:
\begin{align}
  \label{eq:control_fa}
&\bE[\ps{\nabla J(\theta_t),(\hat{\delta}(\tilde{x}_t,\bar{\omega}_*(\theta_t))
- \delta(\tilde{x}_t,\theta_t)) \psi_{\theta_t}(\tilde{S}_t,\tilde{A}_t)}]\nonumber\\
&\geq - C ((1+C_V)C_{\nabla J} + \tilde{C}_{1-\gamma}) \bE[\|\theta_t - \theta_{t-\tau}\|]
- C C_{\nabla J} (1+ C_V) |\mA| \sum_{i=t-\tau}^t \bE[\|\theta_i - \theta_{t-\tau}\|]\nonumber\\
&\quad - C C_{\nabla J} (1+C_V) \sigma^{\tau-1}
- C C_{\nabla J} \epsilon_{\text{FA}}\nonumber\\
&\geq -C \left( (  ((1+C_V)C_{\nabla J} + \tilde{C}_{1-\gamma}) \tau + C_{\nabla J} (1+C_V) |\mA| \tau^2) \frac{\alpha_{t-\tau}}{1-\gamma} + C_{\nabla J} (1+ C_V) \alpha_T + C_{\nabla J} \epsilon_{\text{FA}}  \right)
\end{align}
where the last inequality has already been established in Sec.~\ref{subsec:critic_rate} with the choice of the mixing time $\tau = \tau_T$.

\subsubsection{End of the proof of Th.~\ref{th:actor_rate}}

Combining Eq.~\eqref{eq:taylor} with Eqs.~\eqref{eq:taylor_term1}, \eqref{eq:control_Gamma} and~\eqref{eq:control_fa} yields:
\begin{multline}
\bE[J(\theta_{t+1})] \geq \bE[J(\theta_t)] + \frac{\alpha_t}{1-\gamma} \bE[\|\nabla J(\theta_t)\|^2]
-C \frac{\alpha_t}{1-\gamma} \sqrt{\bE[\|\nabla J(\theta_t)\|^2]}
\sqrt{\bE[\|\omega_t - \bar{\omega}_*(\theta_t)\|^2]}\\
- C \frac{\alpha_t}{1-\gamma} ( (C_{1-\gamma}^1 \tau +  C_{1-\gamma}^2 \tau^2) \frac{\alpha_{t-\tau}}{1-\gamma} + C_{1-\gamma}^3 \alpha_T + C_{\nabla J} \epsilon_{\text{FA}})
- C L_{\nabla J}\frac{\alpha_t^2}{(1-\gamma)^2}\,,
\end{multline}
where~$C_{1-\gamma}^1 \eqdef (1+C_V)C_{\nabla J} + \tilde{C}_{1-\gamma} + C_{1-\gamma}$, $C_{1-\gamma}^2 \eqdef C_V C_{\nabla J} + C_{\nabla J} (1+ C_V) |\mA|$ and~$C_{1-\gamma}^3 \eqdef C_V C_{\nabla J} + C_{\nabla J} (1+C_V)$\,.

Rearranging and summing this inequality for $t= \tau_T$ to~$T$ lead to:
\begin{equation}
  \label{eq:ineq_rate_actor}
\frac{1}{T-\tau_T +1} \sum_{t= \tau_T}^T \bE[\|\nabla J(\theta_t)\|^2] \leq U_1(T) + U_2(T) + U_3(T) + CC_{\nabla J} \epsilon_{\text{FA}}\,,
\end{equation}
where
\begin{align}
U_1(T) &\eqdef \frac{1}{T-\tau_T +1} \sum_{t= \tau_T}^T \frac{1-\gamma}{\alpha_t}(\bE[J(\theta_{t+1})] - \bE[J(\theta_t)])\,,\\
U_2(T) &\eqdef \frac{C}{T-\tau_T +1} \sum_{t=\tau_T}^T \left((C_{1-\gamma}^1 \tau_T +  C_{1-\gamma}^2 \tau_T^2) \frac{\alpha_{t-\tau_T}}{1-\gamma} + C_{1-\gamma}^3 \alpha_T +  L_{\nabla J} \frac{\alpha_t}{1-\gamma} \right) \,,\\
U_3(T) &\eqdef \frac{C}{T-\tau_T +1} \sum_{t= \tau_T}^T \sqrt{\bE[\|\nabla J(\theta_t)\|^2]}
\sqrt{\bE[\|\omega_t - \bar{\omega}_*(\theta_t)\|^2]}\,.
\end{align}

Let us now provide estimates of each one of the quantities $U_i(T)$ for  $i = 1,2,3$\,.

\begin{enumerate}
\item Since the function~$J$ is bounded by $\frac{U_R}{1-\gamma}$ and the sequence~$(\alpha_t)$ is nonincreasing, the first term can be controlled as follows:
\begin{align}
U_1(T) &= \frac{1-\gamma}{T-\tau_T +1} \left( \frac{1}{\alpha_T} \bE[J(\theta_{T+1})] -   \frac{1}{\alpha_ {\tau_T-1}} \bE[J(\theta_{\tau_T})] + \sum_{t=\tau_T}^T \left( \frac{1}{\alpha_{t-1}} - \frac{1}{\alpha_t} \right) \bE[J(\theta_t)]\right)\nonumber\\
&\leq  \frac{U_R}{T-\tau_T +1} \left( \frac{1}{\alpha_T} + \frac{1}{\alpha_ {\tau_T-1}} + \frac{1}{\alpha_T} - \frac{1}{\alpha_ {\tau_T-1}}  \right)\nonumber\\
&\leq \frac{U_R}{T-\tau_T +1}\frac{2}{\alpha_T} \nonumber\\
&= \mO\left(T^{\alpha-1}\right)\,.
\end{align}

\item We can observe from the policy gradient that~$C_{\nabla J} = \mO((1-\gamma)^{-2}), L_V = \mO((1-\gamma)^{-2})$ and from the definition of the value function that~$C_V = \mO((1-\gamma)^{-1})$. Moreover, it follows from \cite[Lem.~4.2]{zhang-koppel-zhu-basar20} that~$L_{\nabla J} = \mO((1-\gamma)^{-3})$. As a consequence, we have that:
$$
C_{1-\gamma} = \mO((1-\gamma)^{-5}),\, \tilde{C}_{1-\gamma} =  \mO((1-\gamma)^{-4})\,; C_{1-\gamma}^1 = \mO((1-\gamma)^{-5})\,; C_{1-\gamma}^2 = \mO((1-\gamma)^{-3})\,; C_{1-\gamma}^3 = \mO((1-\gamma)^{-3})\,.
$$
Recalling that the sequence of stepsizes~$(\alpha_t)$ is nonincreasing and that~$\tau_T = \mO(\ln T)$, the second term can be estimated by the following derivations:
\begin{align}
U_2(T) &= \frac{C}{T-\tau_T +1} \left( (C_{1-\gamma}^1 \tau_T +  C_{1-\gamma}^2 \tau_T^2) \sum_{t= \tau_T}^T \frac{\alpha_{t-\tau_T}}{1-\gamma} + C_{1-\gamma}^3 (T-\tau_T + 1) \alpha_T + L_{\nabla J}\sum_{t= \tau_T}^T \frac{\alpha_t}{1-\gamma} \right)\nonumber\\
&\leq \frac{C}{T-\tau_T +1} \left( (C_{1-\gamma}^1 \tau_T +  C_{1-\gamma}^2 \tau_T^2) \sum_{t= 0}^{T-\tau_T} \frac{\alpha_{t}}{1-\gamma} + C_{1-\gamma}^3 (T-\tau_T + 1) \alpha_T + L_{\nabla J}\sum_{t= 0}^{T-\tau_T} \frac{\alpha_t}{1-\gamma} \right)\nonumber\\
&\leq \frac{C}{T-\tau_T +1} \left(\frac{(C_{1-\gamma}^1 \tau_T +  C_{1-\gamma}^2 \tau_T^2) + L_{\nabla J}}{1-\gamma} \cdot \frac{(T-\tau_T+1)^{1-\alpha}}{1-\alpha} + C_{1-\gamma}^3 (T-\tau_T + 1) \alpha_T  \right)\nonumber\\
&= \mO \left( \frac{\ln^2 T}{(1-\gamma)^6} T^{-\alpha}    \right)
\end{align}

\item Using the Cauchy-Schwarz inequality, we have:
\begin{equation}
U_3(T) \leq \frac{C}{T-\tau_T +1} \sqrt{\sum_{t= \tau_T}^T \bE[\|\nabla J(\theta_t)\|^2]}\sqrt{\sum_{t= \tau_T}^T \bE[\|\omega_t - \bar{\omega}_*(\theta_t)\|^2]}\,.
\end{equation}
\end{enumerate}

Define the quantities:
\begin{align}
F(T) &\eqdef \frac{1}{T-\tau_T +1} \sum_{t= \tau_T}^T \bE[\|\nabla J(\theta_t)\|^2] \,,\\
E(T) &\eqdef \frac{1}{T-\tau_T +1}\sum_{t= \tau_T}^T \bE[\|\omega_t - \bar{\omega}_*(\theta_t)\|^2] \,,\\
K(T) &\eqdef U_1(T) + U_2(T) + C C_{\nabla J} \epsilon_{\text{FA}}\,.
\end{align}

Using these definitions, we can rewrite Eq.~\eqref{eq:ineq_rate_actor} as follows:
\[
F(T) \leq C \sqrt{F(T)}\sqrt{E(T)} + K(T)\,.
\]
Solving this inequality yields:
\begin{equation}
  \label{eq:rate_actor}
F(T) = \mO(E(T)) + \mO(K(T))\,.
\end{equation}

We conclude the proof by remarking that items (1) to (3) above imply:
\begin{equation}
  \label{eq:rate_actor2}
K(T) = \mO\left(T^{\alpha-1}\right) + \mO\left(\frac{\ln^2 T}{(1-\gamma)^6}T^{-\alpha}\right) + \mO\left(\frac{\epsilon_{\text{FA}}}{(1-\gamma)^2}\right)\,.
\end{equation}

Eqs.~\eqref{eq:rate_actor} and~\eqref{eq:rate_actor2} combined can be explicitely written as follows:
\begin{multline*}
\frac{1}{T-\tau_T +1} \sum_{t= \tau_T}^T \bE[\|\nabla J(\theta_t)\|^2] =  \mO\left(T^{\alpha-1}\right) + \mO\left(\frac{\ln^2 T}{(1-\gamma)^6} T^{-\alpha}\right) + \mO\left(\frac{\epsilon_{\text{FA}}}{(1-\gamma)^2}\right)\\
 + \mO\left( \frac{1}{T-\tau_T +1}\sum_{t= \tau_T}^T \bE[\|\omega_t - \bar{\omega}_*(\theta_t)\|^2]\right)\,.
\end{multline*}
Thus, by combining with the result of Theorem~\ref{th:critic_rate}, we have:
\begin{multline}
\frac{1}{T-\tau_T +1} \sum_{t= \tau_T}^T \bE[\|\nabla J(\theta_t)\|^2] =  \mO\left(T^{\alpha-1}\right) + \mO\left(\frac{\ln^2 T}{(1-\gamma)^6} T^{-\alpha}\right) + \mO\left(\frac{\epsilon_{\text{FA}}}{(1-\gamma)^2}\right)\\
+ \mO(T^{\xi-1}) + \mO\left(\frac{\ln T}{1-\gamma} T^{-\beta}\right)
+ \mO\left(\frac{T^{-2(\alpha - \xi)}}{(1-\gamma)^2}\right)  + \mO(T^{-2(\xi-\beta)})\,.
\end{multline}
Then, we can write
\begin{align*}
\frac{1}{T} \sum_{t= 1}^T \bE[\|\nabla J(\theta_t)\|^2]
&=
\frac{1}{T}\left(\sum_{t=1}^{\tau_T -1} \bE[\|\nabla J(\theta_t)\|^2] + \sum_{t= \tau_T}^T \bE[\|\nabla J(\theta_t)\|^2]\right)\nonumber\\
&\leq \frac{C \ln T}{T} + \mO\left( \frac{1}{T-\tau_T + 1}\sum_{t= \tau_T}^T \bE[\|\nabla J(\theta_t)\|^2] \right) \\
&= \mO\left(T^{\alpha-1}\right) + \mO\left(\frac{\ln T}{(1-\gamma)^6} T^{-\beta} \right) +  \mO\left(\frac{T^{-2(\alpha - \xi)}}{(1-\gamma)^2}\right)  + \mO(T^{-2(\xi-\beta)}) + \mO\left(\frac{\epsilon_{\text{FA}}}{(1-\gamma)^2}\right)\,.
\end{align*}

This completes the proof.

\subsubsection{Proof of Cor.~\ref{corollary}}

The result is a consequence of combining Ths.~\ref{th:critic_rate} and~\ref{th:actor_rate} and simplifying the obtained rate using the fact that~$0 < \beta < \xi < \alpha < 1$\,.

\section{Proof of the stability result}
\label{appendix:stab}

The proof is inspired from the techniques used in \cite{konda-tsitsiklis03slowMC,lakshminarayanan-bhatnagar17}. Note though that our proof deviates from a simple application of these results. On the one hand, the approach of Konda and Tsitsiklis \cite{konda-tsitsiklis03slowMC} is not sufficient to tackle the case of our three timescales algorithms which is more involved than the standard two timescales actor-critic algorithm.
On the other hand, the result of \cite{lakshminarayanan-bhatnagar17} extending the rescaling technique of \cite{borkar-meyn00} to two timescales stochastic approximation algorithms does not handle the Markovian noise and only addresses the case of additive martingale noise.

Before proceeding with the proof, we state the stability result with all the required assumptions.

\subsection{Assumptions and stability theorem}

We first introduce a useful assumption regarding the increments of the actor iterates.

\begin{assumption}
  \label{hyp:bounded-increments}
There exists a constant~$C>0$ s.t. for every~$t \in \bN, \|\theta_{t+1} - \theta_t\| \leq \alpha_t C$\,.
\end{assumption}

In order to satisfy this assumption, one can slightly change the update rule of the actor sequence~$(\theta_t)$ of our algorithm to bound its increments. This trick was previously used in \cite[p.~80]{konda02thesis} for instance and considered later in~\cite{zhang-et-al20}.
Let~$\Gamma: \bR^m \to \bR$ be a function assumed to satisfy the following inequalities for some positive constants~$C_1 < C_2$: for every~$\omega \in \bR^m, \|\omega\|\cdot \Gamma(\omega) \in [C_1,C_2]$\,, and for every~$\omega, \omega' \in \bR^m,$ $|\Gamma(\omega) - \Gamma(\omega')| \leq \frac{C_2 \|\omega - \omega'\|}{1 + \|\omega\| + \|\omega'\|}\,.$
An example of such function as provided in \cite{konda02thesis} is for instance the function defined for every~$\omega \in \bR^m$ by:
\[
\Gamma(\omega) \eqdef \1_{\|\omega\|\leq C_0} + \frac{1+C_0}{1+\|\omega\|}\1_{\|\omega\| \geq C_0}\,,
\]
where~$C_0$ is some given positive constant.
Given such a projection-like function~$\Gamma$, we replace the update rule of the actor of our actor-critic algorithm (see Algorithm~\ref{algo}\,) by a modified update rule guaranteeing Assumption~\ref{hyp:bounded-increments} above as follows:
\[
\theta_{t+1} = \theta_t + \alpha_t \frac{1}{1-\gamma} \Gamma(\omega_t) \delta_{t+1}\psi_{\theta_t}(\tilde{S}_t,\tilde{A}_t)\,.
\]

We introduce an additional assumption on the stepsizes complementing Assumption~\ref{hyp:stepsizes}.\\

\begin{assumption}
  \label{hyp:decreasing-stepsizes}
  \begin{enumerate}[{\sl (i)},leftmargin=*,noitemsep,topsep=0pt]
    The sequences of positive stepsizes satisfy the following:
  \item The sequences~$(\beta_t), (\alpha_t)$ and~$(\xi_t)$ are nonincreasing.  \item For every~$t \in \bN, 0 < \xi_t \leq 1$\,.\\
  \end{enumerate}
\end{assumption}

\begin{theorem}
  \label{th:stability}
  Let Assumptions~\ref{hyp:grad_pi}\,, \ref{hyp:markov_chain}\,,
  \ref{hyp:stepsizes}\,, \ref{hyp:features}\,, \ref{hyp:eigenval}\,, \ref{hyp:bounded-increments} and~\ref{hyp:decreasing-stepsizes} hold true.
  Then, $\sup_k (\|\bar{\omega}_k\| + \|\omega_k\|) < \infty, a.s.$, i.e., Assumption~\ref{hyp:stability} holds true.
\end{theorem}

The proof of this result proceeds as for our convergence result: we address the faster timescale first before analyzing the slower one.

\subsection{Faster timescale analysis}
\label{subsec:stab-faster-ts}

In this section, our goal is to bound the norm of the sequence~$(\omega_t)$ evolving on the fast timescale driven by the stepsizes~$(\beta_t)$ using the norm of the sequence~$(\bar{\omega}_t)$ updated in a slower timescale defined by the stepsizes~$(\xi_t)$.
In order to use a rescaling technique inspired from~\cite{borkar-meyn00,lakshminarayanan-bhatnagar17}, we introduce a few useful notations. Define for every~$\theta \in \bR^d$ the functions~$h_\theta: \bR \times \mS^2 \to \bR^{2m}$ and~$G_\theta: \bR \times \mS^2 \to \bR^{2m \times 2m}$ for every $y = (r, \tilde{s}, s') \in \bR  \times \mS^2$ by:
\[
h_\theta(y) \eqdef
\begin{bmatrix}
 r\phi(\tilde{s}) \\ 0
\end{bmatrix}\,,
\quad
G_\theta(y) \eqdef
\begin{bmatrix}
\phi(\tilde{s}) \phi(\tilde{s})^T &
-\gamma \phi(\tilde{s})\phi(s')^T \\
0  &
0
\end{bmatrix}\,.
\]

Consider the sequences~$r_k \eqdef (\omega_k^{T}, \bar{\omega}_k^{T})^{T}$ and~$Y_{k+1} \eqdef (\tilde{S}_k,S_{k+1},R_{k+1})$.
Given the update rules of the sequences~$(\omega_k)$ and~$(\bar{\omega}_k)$ from our algorithm, we have the following decomposition:
\[
r_{k+1} = r_k + \beta_k \biggl(h_{\theta_k}(Y_{k+1}) -  G_{\theta_k}(Y_{k+1})r_k \biggr) + \beta_k M_{k+1} r_k + \beta_k \eta_{k+1}\,,
\]
where $(M_{k+1})$ is a~$2m \times 2m$-matrix valued martingale difference sequence w.r.t. the filtration~$(\cF_k)$ (where the~$\sigma$-field is generated by all the r.v.s up to time~$k$) defined for every~$k \in \bN$ by:
\[M_{k+1} \eqdef
	\begin{bmatrix}
	0 &
	\gamma \phi(\tilde{S}_k) (\phi(S_{k+1}) - \bE[\phi(S_{k+1}) | \cF_k])^T \\
	0  &
	0
	\end{bmatrix}\,,
\]
and $(\eta_{k+1})$ is a~$2m$-vector valued sequence defined for every~$k \in \bN$ by :
\[\eta_{k+1} \eqdef \frac{\xi_k}{\beta_k}
\begin{bmatrix}
0 \\
\omega_{k+1}  - \bar{\omega}_k
\end{bmatrix}\,.
\]
Consider now the functions~$\tilde{h}: \bR^d \to \bR^{2m}$ and~$\tilde{G}: \bR^d \to \bR^{2m \times 2m}$ defined for every~$\theta \in \bR^d$ by:
\[
\tilde{h}(\theta) \eqdef
\begin{bmatrix}
	h(\theta) \\ 0
\end{bmatrix}\,,
\quad
\tilde{G}(\theta) \eqdef
\begin{bmatrix}
	\bar{G}(\theta) &  - \gamma \Phi^T D_{\rho,\theta} P_{\theta} \Phi \\
	0 & 0
\end{bmatrix}\,,
\]
where we recall that~$h(\theta) = \Phi^T D_{\rho,\theta} R_\theta$ and~$\bar{G}(\theta) =\Phi^T D_{\rho,\theta} \Phi$.

Let the sequence of nonnegative integers~$(k_j^{\beta})$ be defined by:
\begin{equation}
  \label{eq:k_j_beta}
k_0^{\beta} = 0\,,\quad k_{j+1}^{\beta} = \min \left\{k > k_j^{\beta} : \sum_{l=k_j}^{k-1} \beta_l > T \right\}\,,
\end{equation}
where~$T$ is a positive constant that will be chosen appropiately later on. For notational convenience, in the rest of Section~\ref{subsec:stab-faster-ts}, we will simply use the notation~$(k_j)$ for the sequence~$(k_j^{\beta})$. The superscript~$\beta$ will be useful when considering a different timescale in the upcoming section.

Then, for any~$j \in \bN$, we can introduce the rescaled iterates~
$\hat{r}_k^j = \frac{r_k}{\max(1,\|r_{k_j}\|)}$ defined for every~$k\geq k_j$ and which satisfy the following recurrence relation:
\[
\hat{r}_{k+1}^j = \hat{r}_k^j + \beta_k \left(\frac{\tilde{h}(\theta_k)}{\max(1,\|r_{k_j}\|)} - \tilde{G}(\theta_k)\hat{r}_k^j\right) + \beta_k \hat{\epsilon}_{k+1}^j + \beta_k \frac{\eta_{k+1}}{\max(1,\|r_{k_j}\|)}\,,
\]
where for $k\geq k_j$, the term $\hat{\epsilon}_{k+1}^j$ is defined by:
\[\hat{\epsilon}_{k+1}^j \eqdef
	\biggl(\frac{h_{\theta_k}(Y_{k+1}) - \tilde{h}(\theta_k)}{\max(1,\|r_{k_j}\|)} -  (G_{\theta_k}(Y_{k+1}) - \tilde{G}(\theta_k))\hat{r}_k^j \biggr) + \beta_k M_{k+1} \hat{r}_k^j\,.
\]
We also introduce the iterates~$(r_k^j)$ defined as follows: $r_{k_j}^j = \hat{r}_{k_j}$ and
\[
r_{k+1}^j = r_k^j + \beta_k \left(\frac{\tilde{h}(\theta_k)}{\max(1,\|r_{k_j}\|)} - \tilde{G}(\theta_k)r_k^j\right)\, + \beta_k \frac{\eta_{k+1}}{\max(1,\|r_{k_j}\|)}.
\]
Observing that the sequence $r_k^j$ can be written as~$(\omega_k^j,\bar{\omega}_k^j)$ and given the update rule of~$(r_k^j)$, we have the following for every~$j \in \bN, k \geq k_j$:
\begin{equation}
  \label{eq:update}
  \begin{cases}
    \omega_{k+1}^j&= \omega_k^j + \beta_k \left(\frac{h(\theta_k)}{\max(1,\|r_{k_j}\|)} + \gamma\Phi^TD_{\rho,\theta_k}P_{\theta_k}\Phi\bar{\omega}_k^j - \bar{G}(\theta_k) \omega_k^j\right)\,, \\
    \bar{\omega}_{k+1}^j&=  \bar{\omega}_{k}^j + \xi_k(\omega_{k+1}^j - \bar{\omega}_{k}^j)\,,\,
	  \end{cases}
\end{equation}

Before proceeding, we recall two useful lemmas which we will repeatedly use in the proofs.
\begin{lemma}
	\label{lem:boundedness}
Let~$\lambda \in [0,1)$. Suppose that~$(u_k)$ and~$(\varepsilon_k)$ are nonnegative sequences satisfying~$u_{k+1} \leq \lambda u_k + \varepsilon_k$. If $\sup_k \varepsilon_k < \infty$, then~$\sup_k u_k < \infty$.
\end{lemma}

\begin{lemma}
	\label{lem:unif-pos-def}
Let~$G \in \bR^{m \times m}$ be a matrix verifying for every~$\omega \in \bR^m, \, \omega^T G \omega \geq \epsilon \|\omega\|^2$ where~$\epsilon >0$ is a constant. Then, for sufficiently small~$\gamma >0$, $\|(I- \gamma G) \omega\| \leq (1- \frac{1}{2}\gamma \epsilon) \|\omega\| \leq e^{-\frac{1}{2}\gamma\epsilon} \|\omega\|$.
\end{lemma}

\begin{lemma}
	\label{lemma9:beta}
	We have the following:
\begin{enumerate}[{\sl (i)},leftmargin=*,noitemsep,topsep=0pt]
	\item There exists a constant~$C>0$ s.t. $\sup_j \max_{k_j\leq k \leq k_{j+1}} \|r_k^j\| \leq C$.

	\item $\lim_j \max_{k_j\leq k \leq k_{j+1}} \|\hat{r}_{k}^j - r_k^j\| = 0,\, a.s.$

	\item There exists a constant~$C'>0$ s.t. $\sup_j \max_{k_j\leq k \leq k_{j+1}} \|\hat{r}_k^j\| \leq C',\, a.s.$.
\end{enumerate}
\end{lemma}

\begin{proof}
	\begin{enumerate}[{\sl (i)},leftmargin=*,noitemsep,topsep=0pt]

		\item
Let us show that there exists a positive constant~$\tilde{C}>0$ s.t. $\sup_j \max_{k_j\leq k\leq k_{j+1}}  \|\omega_{k}^j\| \leq \tilde{C}$.
For $j$ sufficiently large s.t. Lem.~\ref{lem:unif-pos-def} holds and for $k$ between $k_j$ and $k_{j+1}$, we have
\begin{align}
\nonumber\|\omega_{k+1}^j\| &\leq \|(I-\beta_k\bar{G}(\theta_k))\omega_k^j\| + \beta_k \frac{\|h(\theta_k)\|}{\max(1,\|r_{k_j}\|)} + \beta_k\|\gamma\Phi^TD_{\rho,\theta_k}P_{\theta_k}\Phi\bar{\omega}_{k}^j\|\\
\nonumber	&\leq (1-\frac{1}{2}\beta_k\epsilon)\| \omega_k^j\| + \beta_k \frac{C_1}{\max(1,\|r_{k_j}\|)} + \beta_k C_2 \|\bar{\omega}_{k}^j\|\\
\nonumber&\leq e^{- \frac{1}{2}\epsilon \sum_{i = k_j}^{k}\beta_i}\|\omega_{k_j}^j\| + \left(\sum_{i = k_j}^k\beta_i \right) \frac{C_1}{\max(1,\|r_{k_j}\|)} +  C_2\left(\sum_{i = k_j}^k\beta_i \|\bar{\omega}_{i}^j\|\right)\\
&\leq 1 + T' C_1 + C_2\left(\sum_{i = k_j}^k\beta_i \|\bar{\omega}_{i}^j\|\right)\,,
\label{eq:rec_omega_norm}
\end{align}
where $C_1, C_2$ are two positive constants, $T'$ is a positive constant (which we do not explicit) s.t.~$T'> T$, the second inequality follows from the fact that the matrix $\bar{G}(\theta)$ is~$\epsilon$-uniformly positive definite (i.e., for every~$\omega \in \bR^m,\, \omega^T \bar{G}(\theta) \omega \geq \epsilon\|\omega\|^2$) together with Lem.~\ref{lem:unif-pos-def} and the last inequality stems from the fact that~$\|\omega_{k_j}^j\| \leq 1$ by definition.\\
We now relate the term $\|\bar{\omega}_i^j\|$ to the quantity $\max_{k_j\leq l \leq i}\|\omega_{l}^j\|$. For every~$i \in \{k_j \cdots, k_{j+1}-1\},$
\begin{equation*}
\|\bar{\omega}_{i+1}^j\| 
\leq \|\bar{\omega}_{i}^j\| + \xi_i\|\omega_{i+1}^j\|
\leq \|\bar{\omega}_{k_j}^j\| + \sum_{l = k_j}^{i} \xi_l \|\omega_{l+1}^j\|
\leq 1 + T' \left(\max_{k_j\leq l \leq i} \frac{\xi_l}{\beta_l} \right) \left( \max_{k_j\leq l \leq i+1} \|\omega_{l}^j\| \right)\,.
\end{equation*}
Notice then that $\|\bar{\omega}_k^j\|$ is bounded whenever $\|\omega_{k}^j\|$ is bounded. It remains to show that the sequence~$(\omega_k^j)$ is bounded. For this purpose, combining the above inequality with Eq.~\eqref{eq:rec_omega_norm} yields
\begin{align*}
	\max_{k_j\leq k \leq k_{j+1}}\|\omega_{k}^j\| &\leq (1 + T'C_1 + C_2T') + C_2 T' \max_{k_j\leq k \leq k_{j+1}} \left(\sum_{i = k_j}^{k} \beta_i \left(\max_{k_j\leq l \leq i} \frac{\xi_l}{\beta_l} \right) \left( \max_{k_j\leq l \leq i} \|\omega_{l}^j\| \right) \right)\\
	&\leq (1 + T'C_1 + C_2T') + C_2 T'^2 \left(\max_{k_j\leq k \leq k_{j+1}} \frac{\xi_k}{\beta_k}\right) \left( \max_{k_j\leq k \leq k_{j+1}} \|\omega_{k}^j\| \right)\,.
\end{align*}
Since the sequence $(\frac{\xi_k}{\beta_k})$ converges to $0$ by Assumption~\ref{hyp:stepsizes}\,, there exists $\upsilon >0$ s.t. for $j$ sufficiently large, $C_2T'^2(\max_{k_j\leq k \leq k_{j+1}} \frac{\xi_k}{\beta_k}) \leq 1-\upsilon$. Thus,
$$\max_{k_j\leq k \leq k_{j+1}}\|\omega_{k}^j\| \leq \frac{1 + T'C_1 + C_2T'}{\upsilon},$$
which concludes the proof.\\

		\item This result is a consequence of applying \cite[Lem.~9]{konda-tsitsiklis03slowMC} to the sequence $(r_t)$.
		Note that Assumption~6 in \cite{konda-tsitsiklis03slowMC} is not needed for this result to hold since we proved item one. This means that the matrix~$\tilde{G}(\theta)$ is not required to be uniformly positive definite (see \cite[Assumption~6]{konda-tsitsiklis03slowMC}). We leave the verification of the remaining technical assumptions to the reader.\\

		\item This item follows from combining the two first items with the triangular inequality. Remark that the second item implies that the sequence $(\max_{k_j \leq k \leq k_{j+1}} \|\hat{r}_k^j - r_k^j\|)_j$ is a.s. bounded.

	\end{enumerate}
\end{proof}

Recall that for every~$\bar{\omega} \in \bR^m, \theta \in \bR^d$,
\[
\omega_*(\bar{\omega},\theta) = \bar{G}(\theta)^{-1} \left(h(\theta) + \gamma \Phi^T D_{\rho,\theta} P_\theta \Phi \bar{\omega}\right)\,.
\]
Now, we define for every~$j \in \bN$ and for every~$\bar{\omega} \in \bR^m, \theta \in \bR^d$ a rescaled version~$\tilde{\omega}^*_j(\bar{\omega},\theta)$ of~$\omega_*(\bar{\omega},\theta)$ as follows:
\begin{equation}
  \label{eq:tilde-omega-j-star}
\tilde{\omega}_j^*(\bar{\omega},\theta) := \bar{G}(\theta)^{-1} \left(\frac{h(\theta)}{\max(1,\|r_{k_{j}}\|)} + \gamma \Phi^T D_{\rho,\theta} P_\theta \Phi \bar{\omega}\right)\,.
\end{equation}

Notice that there exists a constant $C^*>0$ s.t. for every~$j \in \bN$, for every~$\bar{\omega} \in \bR^m, \theta \in \bR^d$,
\begin{equation}
\label{eq:bound_on_omega_star}
\max{(\|\omega_*(\bar{\omega},\theta)\|, \|\tilde{\omega}_j^*(\bar{\omega},\theta)\|)}
\leq C^*(1+\|\bar{\omega}\|)\,.
\end{equation}

\begin{lemma}
	\label{lem:contraction_omega}
	There exists~$j_* \in \bN, T_* > 0$ s.t for every integer~$j \geq j_*$ and~$T \geq T_*$ ($T$ as in the definition of $k_j$), if~$\|\omega_{k_{j}} - \omega_*(\bar{\omega}_{k_j},\theta_{k_j})\| > C_1 (1 + \|\bar{\omega}_{k_j}\|)$ for some constant~$C_1>0$\,, then,
	$$\|\omega_{k_{j+1}} - \omega_*(\bar{\omega}_{k_{j+1}},\theta_{k_{j+1}})\| \leq \frac{3}{4} \|\omega_{k_{j}} - \omega_*(\bar{\omega}_{k_j},\theta_{k_j})\|\,, a.s.$$
\end{lemma}
\begin{proof}
	Notice that if $\|\omega_{k_{j}} - \omega_*(\bar{\omega}_{k_j},\theta_{k_j})\| > C_1 (1 + \|\bar{\omega}_{k_j}\|)$,
	using Eq.~\eqref{eq:bound_on_omega_star}, we obtain that:
	\begin{align*}
	\|r_{k_j}\| = \|(\omega_{k_{j}},\bar{\omega}_{k_j})\| &=\sqrt{\|\omega_{k_{j}}- \omega_*(\bar{\omega}_{k_j},\theta_{k_j}) + \omega_*(\bar{\omega}_{k_j},\theta_{k_j})\|^2 + \|\bar{\omega}_{k_j}\|^2 }\\
	&\leq \sqrt{2 \|\omega_{k_{j}}- \omega_*(\bar{\omega}_{k_j},\theta_{k_j})\|^2 + 2\|\omega_*(\bar{\omega}_{k_j},\theta_{k_j})\|^2+ \|\bar{\omega}_{k_j}\|^2}\\
	&\leq \sqrt{2}\|\omega_{k_{j}}- \omega_*(\bar{\omega}_{k_j},\theta_{k_j})\| + \sqrt{2 {C^*}^2(1 + \|\bar{\omega}_{k_j}\|)^2 + \|\bar{\omega}_{k_j}\|^2}\\
	&\leq \sqrt{2}\|\omega_{k_{j}}- \omega_*(\bar{\omega}_{k_j},\theta_{k_j})\| + \sqrt{2} C^* + (\sqrt{2}C^*+1) \|\bar{\omega}_{k_j}\|)\,.
	\end{align*}
	As a consequence, we have:
	\begin{equation*}
		\frac{\|\omega_{k_{j}} - \omega_*(\bar{\omega}_{k_j},\theta_{k_j})\|}{\max(1,\|r_{k_j}\|)}
		\geq \frac{\|\omega_{k_{j}} - \omega_*(\bar{\omega}_{k_j},\theta_{k_j})\|}{ \sqrt{2}\|\omega_{k_{j}}- \omega_*(\bar{\omega}_{k_j},\theta_{k_j})\| + (\sqrt{2}C^* +1)(1 + \|\bar{\omega}_{k_j}\|) }
		\geq \frac{1}{\sqrt{2} + \frac{\sqrt{2}C^* +1}{C_1}}\,.
	\end{equation*}
	Then, setting $C_2 \eqdef \sqrt{2} + \frac{\sqrt{2}C^* +1}{C_1}$, it follows that:
	\begin{multline}
		\label{eq:inequality_for_contraction}
		\frac{\|\omega_{k_{j+1}} - \omega_*(\bar{\omega}_{k_{j+1}},\theta_{k_{j+1}})\|}{\|\omega_{k_{j}} - \omega_*(\bar{\omega}_{k_j},\theta_{k_j})\|}
		= \frac{\|\hat{\omega}_{k_{j+1}}^j - \tilde{\omega}_j^*(\hat{\bar{\omega}}_{k_{j+1}}^j,\theta_{k_{j+1}})\|}{\|\hat{\omega}_{k_{j}}^j - \tilde{\omega}_j^*(\hat{\bar{\omega}}^j_{k_j},\theta_{k_j})\|}
		\leq C_2 (\|\hat{\omega}_{k_{j+1}}^j - \omega_{k_{j+1}}^{j}\| + \| \omega_{k_{j+1}}^{j} - \tilde{\omega}_j^*(\hat{\bar{\omega}}_{k_{j+1}},\theta_{k_{j+1}})\|)
\end{multline}
	Since the first term of the right-hand side converges a.s. to zero as~$j$ goes to infinity by Lem.~\ref{lemma9:beta}\,, there exists~$j_0  \in \bN$ s.t. for every~$j \geq j_0$,
		\begin{equation}
			\label{eq:term1}
		\|\hat{\omega}_{k_{j+1}}^j - \omega_{k_{j+1}}^{j}\| \leq \frac{1}{4C_2}\,, a.s.
		\end{equation}
	We now establish a bound for the second term in Eq.~\eqref{eq:inequality_for_contraction}.
		For every $k_j\leq k < k_{j+1}$, we have that:
		\begin{align*}
		\|\omega_{k+1}^j - \tilde{\omega}_j^*(\bar{\omega}_{k+1}^j,\theta_{k+1})\| &= \| \omega_k^j - \tilde{\omega}_j^*(\bar{\omega}_{k}^j,\theta_{k}) + \beta_k (\frac{h(\theta_{k})}{\max(1,\|r_{k_j}\|)} + \gamma\Phi^TD_{\rho,\theta_k}P_{\theta_k}\Phi\bar{\omega}_{k}^j - \bar{G}(\theta_k) \omega_k^j) \\
    &\quad\quad + \tilde{\omega}_j^*(\bar{\omega}_{k}^j,\theta_{k})- \tilde{\omega}_j^*(\bar{\omega}_{k+1}^j,\theta_{k+1})\| \\
		&\leq \| \omega_{k}^j - \tilde{\omega}_j^*(\bar{\omega}_{k}^j,\theta_{k}) -\beta_k \bar{G}(\theta_{k}) (\omega_{k}^j - \tilde{\omega}_j^*(\bar{\omega}_{k}^j,\theta_{k})) \| \\ &\quad\quad+ \|\tilde{\omega}_j^*(\bar{\omega}_{k}^j,\theta_{k}) - \tilde{\omega}_j^*(\bar{\omega}_{k+1}^j,\theta_{k+1})\|\\
		&\leq \|I - \beta_k \bar{G}(\theta_k)\| \|\omega_{k}^j - \tilde{\omega}_j^*(\bar{\omega}_{k_j}^j,\theta_{k})\| + C(\xi_k +\alpha_k)
		\end{align*}
		where~$C >0$ is a constant coming from~Lem.~\ref{lemma9:beta} and the last inequality stems from the fact that the function $(\bar{\omega},\theta) \mapsto\tilde{\omega}_j^*(\bar{\omega},\theta)$ is Lipschitz continuous for every $j$ (by the same arguments as for the proof showing that the function $U$ is Lipschitz before Lemma \ref{lem:hurwitz_critic2}).
		Similarly to the proof of the first item of Lem.~\ref{lemma9:beta}\,, we have:
		\begin{align}
	\nonumber	\| \omega_{k_{j+1}}^j - \tilde{\omega}_j^*(\bar{\omega}_{k_{j+1}}^j,\theta_{k_{j+1}}) \| &\leq
		e^{-\frac{1}{2}\epsilon T } \|\omega_{k_j}^j - \tilde{\omega}_j^*(\bar{\omega}_{k_j}^j,\theta_{k_j})\| + C \sum_{k=k_j}^{k_{j+1}} (\xi_k + \alpha_k)\\
		\label{eq:rec_in_lemma_contraction} &\leq e^{-\frac{1}{2}\epsilon T } \biggl(\|\omega_{k_j}^j\| +\| \tilde{\omega}_j^*(\bar{\omega}_{k_j}^j,\theta_{k_j})\|\biggr) + C \sum_{k=k_j}^{k_{j+1}} (\xi_k + \alpha_k)\,.
		\end{align}
		By definition, $\|\omega_{k_j}^j\| \leq 1$, $\|\bar{\omega}_{k_j}^j\| \leq 1$, and it stems from Eq.~\eqref{eq:bound_on_omega_star} that  $\|\omega_{k_j}^j\| +\| \tilde{\omega}_j^*(\bar{\omega}_{k_j}^j,\theta_{k_j})\|\leq C'$ for some~$C'>0$.
		Choosing $T\geq \frac{2\ln(4C'/C_2)}{\epsilon}$, we obtain:
		$e^{-\frac{1}{2}\epsilon T } \biggl(\|\omega_{k_j}^j\| +\| \tilde{\omega}_j^*(\bar{\omega}_{k_j}^j,\theta_{k_j})\|\biggr) \leq \frac{1}{4C_2}\,.$
		We also have that for every~$j \in \bN$,\, $\sum_{k=k_j}^{k_{j+1}} (\xi_k + \alpha_k) \leq  \max_{k_j \leq k \leq k_{j+1}} \frac{\xi_k + \alpha_{k}}{\beta_{k}} T'$.
		Since $(\xi_k +\alpha_k)/\beta_k \to 0$, there exists $j_1 \in \bN$ s.t., for every $j\geq j_1$\,, $C\sum_{k=k_j}^{k_{j+1}} (\xi_k + \alpha_k) \leq \frac{1}{4C_2}\,.$
		As a consequence, Eq.~\eqref{eq:rec_in_lemma_contraction} implies that for every $j\geq \max(j_0 ,j_1)$,
		\begin{equation}
			\label{eq:term2}
		\| \omega_{k_{j+1}}^j - \tilde{\omega}_j^*(\bar{\omega}_{k_{j+1}}^j,\theta_{k_{j+1}}) \| \leq \frac{1}{2C_2} \,.
		\end{equation}
		Combining Eq.~\eqref{eq:inequality_for_contraction} with Eqs.~\eqref{eq:term1} and~\eqref{eq:term2} yields for every
		$j\geq\max(j_0 ,j_1)$,
		\[
		 \frac{\|\omega_{k_{j+1}} - \omega_*(\bar{\omega}_{k_{j+1}},\theta_{k_{j+1}})\|}{\|\omega_{k_{j}} - \omega_*(\bar{\omega}_{k_j},\theta_{k_j})\|} \leq C_2 \left(\frac{1}{4C_2} + \frac{1}{2C_2}\right) = \frac{3}{4}\,,
		 \]
		which is the desired inequality.

\end{proof}

\begin{theorem}
  \label{th:stab-omega}
	There exists a constant $C>0$ s.t. for every $j \in \bN$,
		\begin{enumerate}[{\sl (i)},leftmargin=*,noitemsep,topsep=0pt]
				\item  $\|\omega_{k_j} - \omega_*(\bar{\omega}_{k_j},\theta_{k_j})\| \leq C(1 + \|\bar{\omega}_{k_j}\|), a.s.$

				\item \label{cor:omegakj-leq-baromegakj}  $\|\omega_{k_j}\| \leq C(1 + \|\bar{\omega}_{k_j}\|), a.s.$

				\item \label{cor:bound_omega_with_omega_bar}
				$\max_{k_j \leq k \leq k_{j+1}}\|\omega_k\| \leq C(1+ \|\bar{\omega}_{k_j}\|), a.s.$
		\end{enumerate}
\end{theorem}

\begin{proof}
	\begin{enumerate}[{\sl (i)},leftmargin=*,noitemsep,topsep=0pt]
				\item The proof follows exactly the same path than the proof of \cite[Th.~7-(ii)]{lakshminarayanan-bhatnagar17}. We reproduce it here for completeness. On a set of positive probability, let us assume on the contrary that there exists a monotonically increasing sequence $(j_l)$ for which $C_{j_l} \uparrow \infty$ as $l \to \infty$ and $\|\omega_{k_{j_l}}\| \geq C_{j_l} (1 + \|\bar{\omega}_{k_{j_l}}\|)$. Now, from Lem.~\ref{lem:contraction_omega}\,, we know that if $\|\omega_{k_{j}} - \omega_*(\bar{\omega}_{k_j},\theta_{k_j})\| > C_1 (1 + \|\bar{\omega}_{k_j}\|)$, then $\|\omega_{k_i} - \omega_*(\bar{\omega}_{k_i},\theta_{k_i})\|$ for $i\geq j$ falls at an exponential rate until it is within the ball of radius $C_1(1 + \|\bar{\omega}_{k_j}\|)$. Thus, corresponding to the sequence~$(j_l)$, there must exist another sequence~$(j_l')$ s.t. $j_{l-1}\leq j_l'\leq j_l$ and $\|\omega_{k_{j_{l}'-1}} - \omega_*(\bar{\omega}_{k_{j_{l}'-1}},\theta_{k_{j_{l}'-1}})\|$ is within the ball of radius $C_1(1 + \|\bar{\omega}_{k_{j_{l}'-1}}\|)$ and  $\|\omega_{k_{l}'} - \omega_*(\bar{\omega}_{k_l'},\theta_{k_l'})\|$ is greater than $C_{j_l}(1+\|\bar{\omega}_{k_{j'_l}}\|)$. However, we know from Lem.~\ref{lemma9:beta} that the iterates can only grow by a factor of $C'$ between the time~$k_{j_l'-1}$ and~$k_{j_l'}$. This leads to a contradiction. We conclude that $\|\omega_{k_j} - \omega_*(\bar{\omega}_{k_j},\theta_{k_j})\| \leq \bar{C}(1 + \|\bar{\omega}_{k_j}\|)$ for some $\bar{C}>0$.\\

			\item The inequality is a consequence of the first item combined with Eq.~\eqref{eq:bound_on_omega_star}.\\

			\item Using the definition of the sequence~$(\hat{\omega}_k^j)$ and the third item of Lem.~\ref{lemma9:beta} (providing the constant $C'$) combined with the second item of the present theorem, we obtain the desired result as follows:
			\begin{equation*}
				\|\omega_{k}\| = \max(1,\|(\omega_{k_{j}},\bar{\omega}_{k_j})\|) \|\hat{\omega}_k^j\|
				\leq (1 + \|\omega_{k_{j}}\| + \|\bar{\omega}_{k_j}\|)C'
				\leq C(1 + \|\bar{\omega}_{k_j}\|)\,,
			\end{equation*}
			where $C \eqdef C'(1+\bar{C})$ and~$\bar{C}$ comes from the proof of the first item.

		\end{enumerate}
\end{proof}

\subsection{Slower timescale analysis}

We now turn to the analysis of the sequence~$(\bar{\omega}_t)$ evolving in a slower timescale than that of the sequence~$(\omega_t)$.
Recall the update rule of the sequence~$(\bar{\omega}_t)$:
\begin{align*}
\bar{\omega}_{k+1} &= \bar{\omega}_k + \xi_k (\omega_{k+1} - \bar{\omega}_k)\,.
\end{align*}
Given a constant $T>0$, let $(k_j^\beta)$ be defined as in Eq.~\eqref{eq:k_j_beta} and define the sequence~$(k_n^{\xi})$ (which we will sometimes simply denote $(k_n)$ in the rest of this section when unambiguous) as follows:
\[
k_0^{\xi} = 0, \quad\quad k_{n+1}^{\xi} = \min\left\{k_j^\beta > k_n\, :\, j \in \bN\,, \sum_{l=k_n}^{k_j^\beta-1} \xi_l > T\right\}\,.
\]

Since $\xi_k/\beta_k$ converges to $0$, there exists $C_\xi>0$ such that
$T \leq \sum_{l = k_n}^{k_{n+1}} \xi_l \leq C_\xi T$.

Similarly to the previous section, for every~$n \in \bN$, we define the rescaled iterates~$(\hat{\omega}_{k}^n)_k$ and~$(\hat{\bar{\omega}}_{k}^n)_k$ for every~$k \geq k_n$ as follows:
\begin{equation}
  \label{eq:rescaled1}
  \begin{cases}
    \hat{\omega}_{k_n}^n &= \frac{\omega_{k_n}}{\max(1,\|r_{k_n}\|)}\\
    \hat{\omega}_{k+1}^n &= \hat{\omega}_k^n + \beta_k \phi(\tilde{S}_k)\left(\frac{R_{k+1}}{\max(1,\|r_{k_n}\|)} + \gamma \phi(S_{k+1})^T\hat{\bar{\omega}}_k^n - \phi(\tilde{S}_k)^T \hat{\omega}_k^n\right)
	  \end{cases}
  \,;\quad
  \begin{cases}
    \hat{\bar{\omega}}_{k_n}^n &= \frac{\bar{\omega}_{k_n}}{\max(1,\|r_{k_n}\|)} \\
    \hat{\bar{\omega}}_{k+1}^n &= \hat{\bar{\omega}}_k^n + \xi_k (\hat{\omega}_{k+1}^n - \hat{\bar{\omega}}_k^n)\,,
    \end{cases}
\end{equation}

and their noiseless counterparts~$(\omega_k^n)_k$ and~$(\bar{\omega}_k^n)_k$ are defined for every~$n \in \bN, k \geq k_n$ by:
\begin{equation}
  \label{eq:rescaled3}
  \begin{cases}
    \omega_{k_n}^n &= \hat{\omega}_{k_n}^n\\
    \omega_{k+1}^n &= \omega_{k}^n + \beta_k (\frac{h(\theta_k)}{\max(1,\|r_{k_n}\|)} + \gamma \Phi^T D_{\rho,\theta_{k}} P_{\theta_k}\Phi \bar{\omega}_k^n - \bar{G}(\theta_k) \omega_{k}^n)
	  \end{cases}
  \,;\quad
  \begin{cases}
    \bar{\omega}_{k_n}^n &= \hat{\bar{\omega}}_{k_n}^n\\
    \bar{\omega}_{k+1}^n &= \bar{\omega}_k^n + \xi_k (\omega_{k+1}^n - \bar{\omega}_k^n)\,.
    \end{cases}
\end{equation}

The following lemma states the almost sure boundedness of the above rescaled and noiseless iterates.\\

\begin{lemma}
	\label{lem:boundedness_rescaled}
  \begin{enumerate}[{\sl (i)},leftmargin=*,noitemsep,topsep=0pt]
    The following assertions hold true:
  	\item \label{lem:hat_bounded}
  	$\sup_n \max_{k_n\leq k\leq k_{n+1}} (\|\hat{\bar{\omega}}_k^n\| + \|\hat{\omega}_k^n\|) < \infty\,, a.s.$
  	\item \label{lem:noiseless_bounded}
  	$\sup_n \max_{k_n\leq k\leq k_{n+1}} (\|\bar{\omega}_k^n\| + \|\omega_k^n\|)  < \infty\,, a.s.$
  \end{enumerate}
\end{lemma}

\begin{proof}

  \begin{enumerate}[{\sl (i)},leftmargin=*,noitemsep,topsep=0pt]
    \item Let~$n \in \bN$. By definition of the sequence~$(k_n)$, there exists~$j \in \bN$ s.t.~$k_n = k_{j}^{\beta}$.  There exists~$C>0$ (independent of~$n$) s.t. for every~$k \in \{k_{j}^\beta,\cdots,k_{j+1}^\beta-1\}$,  a.s.,
  	\[
  	\|\hat{\bar{\omega}}_{k+1}^n\| \leq (1-\xi_k)\|\hat{\bar{\omega}}_{k}^n\| + \xi_k \|\hat{\omega}_{k+1}^n\|
  	\leq \|\hat{\bar{\omega}}_{k}^n\| + \xi_k C(1 + \|\hat{\bar{\omega}}_{k_{j}^\beta}^n\|)\,,
  	\]
  	where we used Th.~\ref{th:stab-omega}-\ref{cor:bound_omega_with_omega_bar} for the last inequality. It follows that for every $k \in \{k_{j}^\beta,\cdots,k_{j+1}^\beta-1\}$, a.s.,
  	\[
    \|\hat{\bar{\omega}}_{k+1}^n\| \leq \left(1 + C\sum_{l=k_{j}^\beta}^{k} \xi_l\right) \|\hat{\bar{\omega}}_{k_{j}^\beta}^n\| + C\sum_{l=k_{j}^\beta}^{k} \xi_l
    \leq e^{C \sum_{l=k_{j}^\beta}^{k}\xi_l} \|\hat{\bar{\omega}}_{k_{j}^\beta}^n\|+ C\sum_{l=k_{j}^\beta}^{k} \xi_l\,.
    \]
    As a consequence, using the notation~$u_j \eqdef \sum_{l=k_{j}^\beta}^{k_{j+1}^\beta-1}\xi_l$ for every~$j \in \bN$, we obtain that a.s.,
    \begin{equation}
      \label{eq:rec_hatbaromega}
    \|\hat{\bar{\omega}}_{k_{j+1}^\beta}^n\|
    \leq e^{C u_{j}} \|\hat{\bar{\omega}}_{k_{j}^\beta}^n\|+ C u_{j}\,.
  \end{equation}

    For every~$l, p \in \bN,$ let $\mathcal{U}(l,p)$ be the set of integers~$j$ s.t.~$l \leq k_j^{\beta} \leq p$. Recall that for every~$n \in \bN$, there exist integers $j_{n+1} > j_n$ s.t.~$k_n = k_{j_n}^{\beta}$ and~$k_{n+1} = k_{j_{n+1}}^{\beta}$ by definition of the sequence~$(k_n)$.
  	Then, using Eq.~\eqref{eq:rec_hatbaromega}, we have for every~$j \in U(k_n,k_{n+1}),$ a.s,
  	\begin{align*}
  	\|\hat{\bar{\omega}}_{k_{j+1}^\beta}^n\|
    &\leq \left(\prod_{i \in \mathcal{U}(k_n,k_{j+1}^\beta-1)} e^{C u_i}\right) \|\hat{\bar{\omega}}_{k_n}^n\|
    + C \sum_{p \in \mathcal{U}(k_n,k_{j+1}^\beta-1)} \left(\prod_{i \in \mathcal{U}(k_{p+1}^\beta,k_{j+1}^\beta-1) }e^{C u_i}\right) u_p\\
  	&= e^{C \sum_{l = k_n}^{k_{j+1}^\beta-1} \xi_l} \|\hat{\bar{\omega}}_{k_n}^n\| + C \sum_{p \in \mathcal{U}(k_n,k_{j+1}^\beta-1)} e^{C \sum_{l = k_{p+1}^\beta}^{k_{j+1}^\beta-1} \xi_l} u_p\\
  	&\leq e^{CC_\xi T} +  C e^{CC_\xi T}C_\xi T\,,
  	\end{align*}
  	where the last inequality comes from the facts that $\|\hat{\bar{\omega}}_{k_n}^n\|$ is bounded by $1$ and that $\sum_{l=k_n}^{k_{n+1}} \xi_l \leq C_\xi T$. To conclude, notice that this bound also holds for any $k \in \{k_n,\cdots,k_{n+1}\}$ and use Th.~\ref{th:stab-omega}-\ref{cor:bound_omega_with_omega_bar} to bound~$\|\hat{\omega}_k^n\|$.\\

    \item The proof of this item follows a similar path to the first one. Notice that the iterates considered in this item are noiseless versions of their counterparts which were shown to be bounded in the first item.
  \end{enumerate}
\end{proof}

\begin{lemma}
	\label{lem:lemma_8_slow}
	$\lim_n \max_{k_n\leq k\leq k_{n+1}} \left\| (\hat{\omega}_k^n,\hat{\bar{\omega}}_k^n) - (\omega_{k}^n,\bar{\omega}_k^n)\right\| = 0\,.$
\end{lemma}
\begin{proof}
	Let~$n \in \bN$. Consider the shorthand notations $x_k^n \eqdef \hat{\omega}_k^n - \omega_{k}^n$ and $y_k^n \eqdef \hat{\bar{\omega}}_k^n - \bar{\omega}_k^n$ for $k\geq k_n^\xi$. Note that for every~$k\geq k_n^\xi$, the sequences $(x_k^n)_k$ and $(y_k^n)_k$ satisfy the recurrence relations:
  \begin{equation}
    \label{eq:xk-yk}
    \begin{cases}
      x_{k+1}^n &= x_k^n + \beta_k (\gamma \Phi^T D_{\rho,\theta_k}P_{\theta_k}\Phi y_k^n - \bar{G}(\theta_k) x_k^n) + \beta_k \hat{\epsilon}_k^n\,,\\
      y_{k+1}^n &= y_k^n + \xi_k(x_{k+1}^n - y_{k}^n)\,,
  	  \end{cases}
  \end{equation}
	where the Markovian noise sequence $(\hat{\epsilon}_k^n)_k$ is defined for every~$k\geq k_n^\xi$ by:
  \begin{multline}
    \label{eq:hat-eps-k-n}
  \hat{\epsilon}_k^n \eqdef \frac{1}{\max(1,\|r_{k_n^{\xi}}\|)}\biggl[\phi(\tilde{S}_k)R_{k+1} - h(\theta_k)\biggr] + \gamma \biggl[\phi(\tilde{S}_k)\phi(S_{k+1})^T - \Phi^T D_{\rho,\theta_k}P_{\theta_k} \Phi \biggr]\hat{\bar{\omega}}_k^n + \biggl[\bar{G}(\theta_k) - \phi(\tilde{S}_k) \phi(\tilde{S}_k)^T\biggr]\hat{\omega}_{k}^n \,.
\end{multline}
	It is clear that the sequence $(\hat{\epsilon}_k^n)$ is a.s. bounded using Lem.~\ref{lem:boundedness_rescaled}.
  Define the mapping~$x^*: \bR^m \times \bR^d \to \bR^m$ for every~$\theta \in \bR^d, y \in \bR^m$ by:
  \begin{equation}
    \label{eq:x-star-def}
    x^*(y,\theta) \eqdef \gamma \bar{G}(\theta)^{-1} \Phi^T D_{\rho,\theta}P_\theta \Phi y\,.
  \end{equation}
  Then, we have the following decomposition for every~$k\geq k_n^\xi$:
	\begin{align*}
	y_{k+1}^n &= y_k^n + \xi_k (x^*(y_k^n,\theta_k) - y_k^n) + \xi_k(x_{k+1}^n - x_k^n) + \xi_k(x_k^n - x^*(y_k^n,\theta_k))\\
	&= (I_m - \xi_k \bar{G}(\theta_k)^{-1}G(\theta_k)) y_k^n + \xi_k(x_{k+1}^n - x_k^n) + \xi_k(x_k^n - x^*(y_k^n,\theta_k))\,.
	\end{align*}
	Since $\bar{G}(\theta)^{-1}G(\theta)$ is uniformly (in~$\theta$) $\kappa$-positive definite (see Assumption~\ref{hyp:eigenval}), Lem.~\ref{lem:unif-pos-def} implies that there exists~$\kappa >0$ s.t. for sufficiently large $n$ and $k\in \{k_n^\xi,\cdots,k_{n+1}^\xi-1\},$
	\begin{align*}
\|y_{k+1}^n\|	&\leq e^{-\frac{1}{2}\kappa T} \|y_{k_n^{\xi}}^n\| +  \sum_{l = k_n^\xi}^k \xi_l\|x_{l+1}^n - x_l^n\| + \xi_l\|x_l^n - x^*(y_l^n,\theta_l)\|\\
&= \sum_{l = k_n^\xi}^k \xi_l\|x_{l+1}^n - x_l^n\| + \xi_l\|x_l^n - x^*(y_l^n,\theta_l)\|\\
	&\leq \sum_{l = k_n^\xi}^k \xi_l \beta_l C + C_\xi T \max_{l \in\{k_n^\xi,\cdots,k\}}\|x_l^n - x^*(y_l^n,\theta_l)\|\,,
	\end{align*}
	where the equality comes from the fact that $y_{k_n^\xi}^n = 0$ by definition and the last inequality comes from the fact that $x_{l+1}^n - x_l^n = \beta_l (\gamma \Phi^T D_{\rho,\theta_l}P_{\theta_l}\Phi y_l^n - \bar{G}(\theta_l) x_l^n + \hat{\epsilon}_l^n)$ and the a.s. boundedness of the sequences~$(x_k^n)_k$,$(y_k^n)_k$ and~$(\hat{\epsilon}_k^n)_k$ resulting from Lem.~\ref{lem:boundedness_rescaled}.
  Observe then that:
  \begin{equation}
    \sum_{l = k_n^\xi}^{k_{n+1}^{\xi}-1} \xi_l \beta_l
    =   \sum_{l = k_n^\xi}^{k_{n+1}^{\xi}-1} \frac{\xi_l}{\beta_l} \beta_l^2
    =  \sum_{l = k_{j_n}^{\beta}}^{k_{j_{n+1}}^{\beta}-1} \frac{\xi_l}{\beta_l} \beta_l^2
    \leq \max_{k_{j_n}^{\beta} \leq l \leq k_{j_{n+1}}^{\beta}} \left(\frac{\xi_l}{\beta_l}\right) \sum_{l = k_{j_n}^{\beta}}^{+\infty} \beta_l^2\,.
  \end{equation}
  Since~$\sum_n \beta_n^2 < \infty$ and~$\xi_n/\beta_n \to 0$, it follows that~$\sum_{k = k_n^\xi}^{k_{n+1}^\xi-1} \beta_k\xi_k \to 0$. Combining this result with Lem.~\ref{tec_lem:cvg_x} below yields:
	\begin{equation}\label{eq:lem_8_y}
	\lim_{n\to\infty} \max_{k_n^\xi\leq k\leq k_{n+1}^\xi} \|y_k^n\| = 0, \quad a.s.
	\end{equation}

  We now show the same result for the sequence~$(x_k^n)_k$.
  First, observing that~$x_{k_n}^n = 0$, we obtain by iterating Eq.~\eqref{eq:xk-yk} that:
  \[
  x_{k+1}^n = \sum_{l = k_n^\xi}^{k} \left[\prod_{p=l+1}^{k} (I_m - \beta_p \bar{G}(\theta_p)) \right] \beta_l \left( \gamma \Phi^T D_{\rho,\theta_l}P_{\theta_l}\Phi y_{l}^n + \hat{\epsilon}_l^n\right)\,.
  \]

	Then, similarly to the first part of the proof, there exist~$C >0$ and~$\varepsilon >0$ s.t. for sufficiently large~$n$ and~$k_n\leq k\leq k_{n+1},$
	\begin{align*}
	\|x_{k+1}^n\|
	&\leq C\sum_{l = k_n^\xi}^{k}\left[\prod_{p=l+1}^{k} (1-\frac{1}{2}\epsilon \beta_p) \right]\beta_l \|y_l^n\| + \left\|\sum_{l = k_n^\xi}^k \left[\prod_{p=l+1}^{k} (I_m - \beta_p \bar{G}(\theta_p)) \right]\beta_l\hat{\epsilon}_l^n\right\|\\
	&\leq C\frac{2}{\epsilon} \max_{k_n^\xi\leq l\leq k_{n+1}^{\xi}} \|y_l^n\| + \left\|\sum_{l = k_n^\xi}^k \left[\prod_{p=l+1}^{k} (I_m - \beta_p \bar{G}(\theta_p)) \right]\beta_l\hat{\epsilon}_l^n\right\|\,.
	\end{align*}
	where the first inequality stems from the fact that the matrix  $\bar{G}(\theta)$ is uniformly positive definite and Lem.~\ref{lem:unif-pos-def}, and the last inequality is a consequence of \cite[Lem.~12]{kal-mou-nau-tad-wai20}.
	Eq.~\eqref{eq:lem_8_y} and Lem.~\ref{tech_lem:cvg_noise} below entail together that:
  $$\lim_{n\to\infty} \max_{k_n^\xi\leq k\leq k_{n+1}^\xi} \|x_k^n\| = 0, \quad a.s. \,,$$
   which concludes the proof.
\end{proof}

\begin{lemma}
	\label{lem:rec_bar_omega_noiseless}
	There exists a sequence $(\delta_n)$ that converges to $0$ when $n \to \infty$ and a constant $C>0$ s.t. for every~$n \in \bN,$
	$$\|\bar{\omega}_{k_{n+1}}^n \| \leq e^{-\frac{1}{2}\kappa T}\|\bar{\omega}_{k_n}^n\| + \delta_n + \frac{C}{\max(1,\|r_{k_n}\|)}\,.$$
\end{lemma}
\begin{proof}
  Recall from Eq.~\eqref{eq:tilde-omega-j-star} that $ \tilde{\omega}_n^*(\bar{\omega},\theta) := \bar{G}(\theta)^{-1} \left(\frac{h(\theta)}{\max(1,\|r_{k_n}\|)} + \gamma \Phi^T D_{\rho,\theta} P_\theta \Phi \bar{\omega}\right)$ for every~$n \in \bN, \bar{\omega} \in \bR^m, \theta \in \bR^d$. It is clear that for every~$k \geq k_n$:
	$$\bar{\omega}_{k+1}^n = \bar{\omega}_k^n + \xi_k(\tilde{\omega}_n^*(\bar{\omega}_k^n,\theta_k) - \bar{\omega}_k^n) + \xi_k(\omega_{k+1}^n - \omega_{k}^n) + \xi_k(\omega_{k}^n - \tilde{\omega}_n^{*}(\bar{\omega}_{k}^n,\theta_k)) \,.$$

	Rewriting this equation using the definition of $\tilde{\omega}_n^*$ gives us:
	$$\bar{\omega}_{k+1}^n = (I - \xi_k\bar{G}(\theta_k)^{-1}G(\theta_k))\bar{\omega}_k^n + \xi_k \frac{ \bar{G}(\theta_k)^{-1} h(\theta_k)}{\max(1,\|r_{k_n}\|)}  + \xi_k(\omega_{k+1}^n - \omega_{k}^n) + \xi_k(\omega_{k}^n - \tilde{\omega}_n^*(\bar{\omega}_{k}^n,\theta_k)) \,.$$

	Remember that $\bar{G}(\theta)^{-1}G(\theta)$ is (uniformly) positive definite. Thus, for sufficiently large $n$, Lem.~\ref{lem:unif-pos-def} ensures the existence of~$\kappa > 0$ s.t.:
  \begin{equation}
    \label{eq:bar-omega-k-n-ineq}
  \| \bar{\omega}_{k+1}^n\| \leq (1-\frac{1}{2}\kappa \xi_k)\|\bar{\omega}_{k}^n\|
  + \xi_k \frac{\|\bar{G}(\theta_k)^{-1} h(\theta_k)\|}{\max(1,\|r_{k_n}\|)} + \xi_{k}\|\omega_{k+1}^n-\omega_{k}^n\| + \xi_k \|\omega_{k}^n-\tilde{\omega}_n^*(\bar{\omega}_k^n,\theta_k)\|\,.
\end{equation}

	Since the sequences~$(\omega_k^n)$, $(\bar{\omega}_k^n)$ and~$h(\theta_k)$ are bounded and $\sup_{\theta \in \bR^d}  \|\bar{G}(\theta)^{-1}\| < \infty$, there exists $C>0$ s.t. for every $k\in\{k_n^\xi,\cdots,k_{n+1}^\xi\}, \|\bar{G}(\theta_k)^{-1} h(\theta_k)\| \leq C$ and:
	$$\|\omega_{k+1}^n-\omega_{k}^n\| = \beta_k \left\| \frac{h(\theta_k)}{\max(1,\|r_{k_n}\|)} + \gamma \Phi^T D_{\rho,\theta_{k}} P_{\theta_k}\Phi \bar{\omega}_k^n - \bar{G}(\theta_k) \omega_{k}^n\right\| \leq \beta_k C\,.$$

  Therefore, for sufficiently large~$n$,
	$$\| \bar{\omega}_{k_{n+1}^\xi}^n\| \leq e^{-\frac{1}{2}\kappa T}\|\bar{\omega}_{k_n^\xi}^n\| + \frac{C C_\xi T}{\max(1,\|r_{k_n}\|)} + CC_\xi T\beta_{k_n} + \sum_{k = k_n^\xi}^{k_{n+1}^\xi}\xi_k \|\omega_{k}^n-\tilde{\omega}_n^*(\bar{\omega}_k^n,\theta_k)\|\,.$$
	It remains to show that $\sum_{k = k_n^\xi}^{k_{n+1}^\xi}\xi_k \|\omega_{k}^n-\tilde{\omega}_n^*(\bar{\omega}_k^n,\theta_k)\|$ converges to $0$ as~$n \to \infty$. For this purpose, we adopt the same strategy used for studying the sequence $(\bar{\omega}_k^n)$. First, we write for every~$k \geq k_n$,
	$$\omega_{k+1}^n-\tilde{\omega}_n^*(\bar{\omega}_{k+1}^n,\theta_{k+1}) = (I-\beta_k\bar{G}(\theta_k))(\omega_k^n - \tilde{\omega}_n^*(\bar{\omega}_{k}^n,\theta_{k})) - (\tilde{\omega}_n^*(\bar{\omega}_{k+1}^n,\theta_{k+1}) - \tilde{\omega}_n^*(\bar{\omega}_{k}^n,\theta_{k}))\,.$$
	Then, applying Lem.~\ref{lem:unif-pos-def}, for sufficiently large $n$, there exists $\epsilon >0$ s.t. for every $k \in \{k_n,\cdots,k_{n+1}\}$,
	$$\|\omega_{k+1}^n-\tilde{\omega}_n^*(\bar{\omega}_{k+1}^n,\theta_{k+1})\| \leq (1-\beta_k\frac{1}{2}\epsilon)\|\omega_k^n - \tilde{\omega}_n^*(\bar{\omega}_{k}^n,\theta_{k})\| + \|\tilde{\omega}_n^*(\bar{\omega}_{k+1}^n,\theta_{k+1}) - \tilde{\omega}_n^*(\bar{\omega}_{k}^n,\theta_{k})\|\,.$$
	We can show that, for every $n$, the function $(\bar{\omega},\theta) \mapsto \tilde{\omega}_n^*(\bar{\omega},\theta)$ is Lipschitz continuous (same arguments as the proof showing that the function~$U$ is Lipschitz before Lem.~\ref{lem:hurwitz_critic2}).
	It follows that there exists positive constants $C$ and $C'$ s.t. for every $k\in\{k_n^\xi,\cdots,k_{n+1}^\xi\},$
   $$\|\tilde{\omega}_n^*(\bar{\omega}_{k+1}^n,\theta_{k+1}) - \tilde{\omega}_n^*(\bar{\omega}_{k}^n,\theta_{k})\| \leq C\xi_k\|\omega_{k+1}^n - \bar{\omega}_{k}^n\| + C\alpha_k\leq C'\xi_k\,,$$
	where the last inequality comes from the boundedness of the sequences $\omega_k^n$ and $\bar{\omega}_k^n$ for $k \in\{k_n^\xi,\cdots,k_{n+1}^\xi\}$, and the fact that there exists $C>0$ s.t. for every $k$, $\alpha_k \leq C\xi_k$.
	Therefore, noticing that there exists $C>0$ s.t. $\|\omega_{k_n^\xi}^n-\tilde{\omega}_n^*(\bar{\omega}_{k_n^\xi}^n,\theta_{k_n^\xi})\| \leq C$, it is easy to check that
	$$\|\omega_{k+1}^n-\tilde{\omega}_n^*(\bar{\omega}_{k+1}^n,\theta_{k+1})\| \leq e^{-\frac{1}{2}\epsilon \sum_{l=k_n^\xi}^{k}\beta_l} C + C' \sum_{l=k_n^\xi}^{k} e^{-\frac{1}{2}\epsilon\sum_{p = l+1}^{k}\beta_p}\xi_l\,.$$
	To conclude the proof, it is sufficient to show that:
	$$\lim_{n\to\infty}\,\sum_{k = k_n^\xi}^{k_{n+1}^\xi}\xi_k \biggl(e^{-\frac{1}{2}\epsilon \sum_{l=k_n^\xi}^{k-1}\beta_l} + \sum_{l=k_n^\xi}^{k-1} e^{-\frac{1}{2}\epsilon\sum_{p = l+1}^{k-1}\beta_p}\xi_l\biggr) = 0\,.$$
	The proof of this technical result is deferred to Lem.~\ref{tech_lem:bar_omega_rec} below.
\end{proof}

\begin{theorem}
  \label{th:stab-bar-omega}
We have the following:
\begin{enumerate}[{\sl (i)},leftmargin=*,noitemsep,topsep=0pt]
	\item $\sup_n \|\bar{\omega}_{k_n}\| < \infty, a.s.$
  \item $\sup_n \max_{k_n \leq k \leq k_{n+1}} \|\bar{\omega}_k^n\| < \infty, a.s.$
  \item $\sup_k \|\bar{\omega}_k\| < \infty, a.s.$
\end{enumerate}
\end{theorem}

\begin{proof}
  \begin{enumerate}[{\sl (i)},leftmargin=*,noitemsep,topsep=0pt]
      \item Combining Lem.~\ref{lem:rec_bar_omega_noiseless} with Lem.~\ref{lem:lemma_8_slow} implies the existence of a sequence~$(\hat{\delta}_n)$ converging to zero a.s. s.t. for sufficiently large~$n$,
    	$$\|\hat{\bar{\omega}}_{k_{n+1}}^n \| \leq e^{-\frac{1}{2}\kappa T}\|\hat{\bar{\omega}}_{k_n}^n\| + \hat{\delta}_n + \frac{C}{\max(1,\|r_{k_n}\|)}\,.$$
      Multiplying both sides by $\max(1,\|r_{k_n}\|)$ and using the fact that a.s.:
      \[
      \max(1,\|r_{k_n}\|)
      \leq 1 + \|\omega_{k_n}\| + \|\bar{\omega}_{k_n}\|
      \leq (1 + C')( 1 + \|\bar{\omega}_{k_n}\|)\,,
      \]
      where~$C'>0$ in the last inequality is a constant stemming from Th.~\ref{th:stab-omega}-\ref{cor:bound_omega_with_omega_bar}, we obtain a.s.:
      \[
      \|\bar{\omega}_{k_{n+1}}\| \leq (e^{-\frac{1}{2}\kappa T} + (1+C')\hat{\delta}_n)\|\bar{\omega}_{k_n}\| + (1 + C')\hat{\delta}_n + C\,.
      \]
    	The result follows from Lem.~\ref{lem:boundedness}.\\

      \item This result can be proven following similar arguments to the first item by exploiting Eq.~\eqref{eq:bar-omega-k-n-ineq} in the proof of Lem.~\ref{lem:rec_bar_omega_noiseless} and the results therein.\\

      \item First, using the definition of~$(\hat{\bar{\omega}}_k^{n})$, observe that:
\begin{align}
	\label{eq:proof-boundedness}
	\sup_k \|\bar{\omega}_k\| &= \sup_n \max_{k_n \leq k \leq k_{n+1}} \|\bar{\omega}_k\| \nonumber\\
					&= \sup_n \max_{k_n \leq k \leq k_{n+1}}
					\{ \max(1, \|(\omega_{k_n},\bar{\omega}_{k_n})\|) \cdot \|\hat{\bar{\omega}}_k^{n}\| \}\,.
\end{align}
Then, using that~$\max(a,b) \leq a+b$ for any nonnegative reals~$a,b$, together with the triangular inequality, it follows  from Eq.~\eqref{eq:proof-boundedness} that:
\begin{equation}
\sup_k \|\bar{\omega}_k\| \leq
\sup_n (1 + \|\omega_{k_n}\| + \|\bar{\omega}_{k_n}\|) (\max_{k_n \leq k \leq k_{n+1}} \|\bar{\omega}_k^n\|
+ \max_{k_n \leq k \leq k_{n+1}} \|\hat{\bar{\omega}}_k^n - \bar{\omega}_k^n\|)\,.
\end{equation}

Given Th.~\ref{th:stab-omega}-\ref{cor:omegakj-leq-baromegakj}, there exists a constant~$\tilde{C} >0$ s.t. a.s.:
\begin{equation}
\sup_k \|\bar{\omega}_k\| \leq
\sup_n \tilde{C}(1 + \|\bar{\omega}_{k_n}\|) (\max_{k_n \leq k \leq k_{n+1}} \|\bar{\omega}_k^n\|
+ \max_{k_n \leq k \leq k_{n+1}} \|\hat{\bar{\omega}}_k^n - \bar{\omega}_k^n\|)\,.
\end{equation}

The result follows from the boundedness of the sequences~$(\bar{\omega}_{k_n})$ (see the first item) and~$(\bar{\omega}_k^n)$ (see the second item) and Lem.~\ref{lem:lemma_8_slow}.
  \end{enumerate}
\end{proof}

\subsection{Technical lemmas}

\begin{lemma}
	\label{tec_lem:cvg_x}
	With $(x_k^n)$ and $(y_k^n)$ defined as in the proof of Lem.~\ref{lem:lemma_8_slow}, it holds that:
	$$\lim_{n\to\infty}\max_{k_n^{\xi} \leq k \leq k_{n+1}^{\xi}} \|x_k^n - x^*(y_k^n,\theta_k)\| = 0, \quad a.s.\,,$$
  where we recall that for every~$y \in \bR^m, \theta \in \bR^d,\, x^*(y,\theta) = \gamma \bar{G}(\theta)^{-1} \Phi^T D_{\rho,\theta} P_{\theta} \Phi y$ as previously defined in Eq.~\eqref{eq:x-star-def}.
\end{lemma}

\begin{proof}
  Recall that $x_{k_n^\xi}^n = y_{k_n^\xi}^n = 0$.
	Throughout this proof, we will use the shorthand notation~$v_k^n \eqdef x_k^n - x^*(y_k^n,\theta_k)$. Recall that $(x_k^n)$ and $(y_k^n)$ are bounded sequences in the sense of Lem.~\ref{lem:boundedness_rescaled} and so is the sequence $(v_k^n)$. Using Eq.~\eqref{eq:xk-yk}, it is easy to check that the sequence $(v_k^n)$ satisfies for every~$k \geq k_n^\xi$ the recurrence relation:
	$$v_{k+1}^n = (I_m - \beta_k \bar{G}(\theta_k)) v_k^n + (x^*(y_k^n,\theta_k)-x^*(y_{k+1}^n,\theta_{k+1})) + \beta_k \hat{\epsilon}_k^n\,.$$
	Iterating this equality for $k\geq k_n^\xi$ and observing that~$v_{k_n^\xi}^n = 0$ leads to the identity:
	\begin{equation*}
	v_{k+1}^n
	= \sum_{p = k_n^\xi}^{k} \left[\prod_{l = p+1}^{k} \left(I_m - \beta_l\bar{G}(\theta_l)\right)\right] ((x^*(y_p^n,\theta_p)-x^*(y_{p+1}^n,\theta_{p+1})) + \beta_p \hat{\epsilon}_p^n)\,.
\end{equation*}

	It can be shown that the function $(\bar{\omega},\theta) \mapsto x^*(\bar{\omega},\theta)$ is $L$-Lipschitz continuous for some $L>0$ (using the same arguments as for the proof showing that the function~$U$ is Lipschitz before Lem.~\ref{lem:hurwitz_critic2}).
	Furthermore, since $\bar{G}(\theta)$ is uniformly positive definite, applying Lem.~\ref{lem:unif-pos-def} yields the existence of~$\epsilon >0$ s.t. for sufficiently large~$n$ and for~$k \geq k_n^\xi$,
	$$\|v_{k+1}^n\| \leq L\sum_{p=k_n^\xi}^{k} e^{-\frac{1}{2}\epsilon \sum_{l = p+1}^k \beta_l}\|(y_p^n,\theta_p)-(y_{p+1}^n,\theta_{p+1})\|
  +\left\|\sum_{p=k_n^\xi}^{k} \left[\prod_{l = p+1}^{k} (I_m - \beta_l\bar{G}(\theta_l))\right] \beta_p \hat{\epsilon}_p^n\right\|\,.$$

	It can be easily checked that there exist~$C>0$ and~$C'>0$ s.t. for every $k \in \{k_n^\xi,\cdots,k_{n+1}^\xi -1\},$ $\|(y_k^n,\theta_k)-(y_{k+1}^n,\theta_{k+1})\| \leq C(\xi_k + \alpha_k) \leq C'\xi_k$. As a consequence, we obtain for every~$k \in \{k_n^\xi,\cdots,k_{n+1}^\xi - 1\},$
	\begin{align}
	\label{eq:rec_two_term}
	\|v_{k+1}^n\| \leq LC'\sum_{p=k_n^\xi}^{k} e^{-\frac{1}{2}\epsilon \sum_{l = p+1}^k \beta_l}\xi_p +\left\|\sum_{p=k_n^\xi}^{k} \left[\prod_{l = p+1}^{k} (I_m - \beta_l\bar{G}(\theta_l))\right] \beta_p \hat{\epsilon}_p^n\right\|\,.
	\end{align}
	To prove the lemma, it is sufficient to show that both terms on the r.h.s. of the above inequality converge a.s. to $0$.
	For this, recall first from the definition of the sequence $(k_n^\xi)$ that there exist $j_n, j_{n+1} \in \bN$ s.t.~$ k_n^\xi = k_{j_n}^\beta $ and~$k_{n+1}^\xi = k_{j_{n+1}}^\beta$. Observe also that for every~$k \in \{k_n^\xi,\cdots,k_{n+1}^\xi - 1\}$, there exists $i_k \in \{j_n,\cdots,j_{n+1}-1\}$ s.t.~$k\in\{k_{i_k}^\beta,\cdots,k_{i_k+1}^\beta-1\}$. Then, we can rewrite the first term in the above inequality as follows:
	\begin{equation*}
	\sum_{p=k_n^\xi}^{k} e^{-\frac{1}{2}\epsilon \sum_{l = p+1}^k \beta_l}\xi_p =
	\sum_{i = j_n}^{i_k-1} \sum_{p = k_{i}^\beta}^{k_{i+1}^\beta}e^{-\frac{1}{2}\epsilon \sum_{l = p+1}^{k}\beta_l}\xi_p + \sum_{p = k_{i_k}^\beta}^{k}e^{-\frac{1}{2}\epsilon \sum_{l = p+1}^{k}\beta_l}\xi_p\,.
\end{equation*}
	The second term on the r.h.s. of the above equation can be easily upperbounded by $\sum_{p=k_{i_k}^\beta}^{k_{i_k+1}^\beta-1}\xi_p \leq (T + \beta_{k_{i_k+1}^\beta-1}) \max_{k_n^\xi \leq p \leq k_{n+1}^\xi} \frac{\xi_p}{\beta_p}$.
  Now, for the first term of the above equation, notice that
	for $i \in\{j_n,\cdots,i_k-1\}$, $p \in \{k_{i}^\beta,\cdots,k_{i+1}^\beta-1\}$ and $k \in \{k_{i_k}^\beta,\cdots,k_{i_k+1}^\beta\}$, $\sum_{l = p+1}^k \beta_l \geq \sum_{l = k_{i+1}^\beta}^{k_{i_k}^\beta} \beta_l \geq T(i_k-i-1)$ and this implies:
	\begin{equation*}
	\sum_{i = j_n}^{i_k-1} \sum_{p = k_{i}^\beta}^{k_{i+1}^\beta-1}e^{-\frac{1}{2}\epsilon \sum_{l = p+1}^{k}\beta_l}\xi_p
  \leq \sum_{i=j_n}^{i_k-1} e^{-\frac{1}{2}\epsilon T(i_k-1-i)} \sum_{p = k_i^\beta}^{k_{i+1}^\beta-1}\xi_p
	\leq \frac{C_\xi T}{1-e^{-\frac 12 \epsilon T}} \max_{k_n^\xi \leq p \leq k_{n+1}^\xi} \frac{\xi_p}{\beta_p}\,.
\end{equation*}
	We conclude from the above derivations that there exists~$C>0$ (independent of~$n$) s.t. for every~$k \in \{k_n^\xi,\cdots,k_{n+1}^\xi - 1\},$
	$$\sum_{p=k_n^\xi}^{k} e^{-\frac{1}{2}\epsilon \sum_{l = p+1}^k \beta_l}\xi_p \leq C \max_{k_n^\xi \leq p \leq k_{n+1}^\xi} \frac{\xi_p}{\beta_p}\,.$$
	Given Assumption~\ref{hyp:stepsizes}, we deduce from this inequality that the first term on the r.h.s. of Eq.~\eqref{eq:rec_two_term} converges to~$0$, i.e.,
	$$\lim_{n\to\infty} \max_{k_n^\xi \leq k \leq k_{n+1}^\xi}\sum_{p=k_n^\xi}^{k} e^{-\frac{1}{2}\epsilon \sum_{l = p+1}^k \beta_l}\xi_p = 0\,.$$
  As for the second term on the r.h.s. of Eq.~\eqref{eq:rec_two_term}, we control it in the following lemma (Lem.~\ref{tech_lem:cvg_noise}).
\end{proof}

\begin{lemma}
	\label{tech_lem:cvg_noise}
	$\lim_{n\to\infty} \max_{k_n^\xi \leq k \leq k_{n+1}^\xi} \left\|\sum_{p=k_n^\xi}^{k} \left[\prod_{l = p+1}^{k} (I_m - \beta_l\bar{G}(\theta_l))\right] \beta_p \hat{\epsilon}_p^n\right\| = 0, a.s.$
\end{lemma}

\begin{proof}
	Let $\bar{G}_{p+1:k} \eqdef \prod_{l = p+1}^{k} (I_m - \beta_l\bar{G}(\theta_l))$ for every~$p \in \{k_n^{\xi}, \cdots, k_{n+1}^{\xi}\}$ and~$p \leq k-1$.
	As in the proof of Lem.~\ref{tec_lem:cvg_x}, we begin by the observation that there exist $j_n$ and $j_{n+1}$ s.t.~$k_{j_n}^\beta = k_n^\xi, k_{j_{n+1}}^\beta = k_{n+1}^\xi$ and that for every~$k \in \{k_n^\xi,\cdots,k_{n+1}^\xi - 1\}$, there exists $i_k \in \{j_n,\cdots,j_{n+1}-1\}$ s.t.~$k\in\{k_{i_k}^\beta,\cdots,k_{i_k+1}^\beta-1\}$.
  Then, we can write for every~$k \in \{k_n^\xi,\cdots,k_{n+1}^\xi - 1\},$
	$$
  \left\|\sum_{p=k_n^\xi}^{k} \bar{G}_{p+1:k} \beta_p \hat{\epsilon}_p^n \right\|
  \leq \left\|\sum_{i=j_n}^{i_k - 1}\sum_{p=k_{i}^\beta}^{k_{i+1}^\beta} \bar{G}_{p+1:k} \beta_p \hat{\epsilon}_p^n\right\|
  + \left\|\sum_{p=k_{i_k}^\beta}^{k} \bar{G}_{p+1:k} \beta_p \hat{\epsilon}_p^n\right\|\,.
  $$
	We will show that the first term on the r.h.s. of the above inequality converges to $0$ a.s. when $n\to\infty$. A slight change in the following proof will establish the convergence to zero of the second term.
	Notice that for~$k \in \{k_n^\xi,\cdots,k_{n+1}^\xi - 1\},$
	\begin{align}
    \label{eq:all_noise}
\left\|\sum_{i=j_n}^{i_k - 1}\sum_{p=k_{i}^\beta}^{k_{i+1}^\beta} \bar{G}_{p+1:k} \beta_p \hat{\epsilon}_p^n\right\|
	\leq  \sum_{i=j_n}^{i_k-1}\left\|\bar{G}_{k_{i+1}^\beta:k}\right\| \left\|\sum_{p=k_{i}^\beta}^{k_{i+1}^\beta-1} \bar{G}_{p+1:k_{i+1}^\beta-1} \beta_p \hat{\epsilon}_p^n\right\|.
	\end{align}
	Lem~\ref{lem:unif-pos-def} implies that for sufficiently large $n$, for $k\in\{k_{i_k}^\beta,\cdots,k_{i_k+1}^\beta\}\subset \{k_n^\xi,\cdots,k_{n+1}^\xi\}$ and $i \in \{j_n,\cdots,i_k-1\}$
	\begin{equation}
	\left\|\bar{G}_{k_{i+1}^\beta:k}\right\| \leq e^{-\frac{1}{2}\epsilon \sum_{l = k_{i+1}^\beta}^{k}\beta_l}\leq e^{-\frac{1}{2}\epsilon T (i_k -1 - i) }\,.
	\label{eq:bound_G_bar}
	\end{equation}
	Recall now from Eq.~\eqref{eq:hat-eps-k-n} the definition of $\hat{\epsilon}_p^n$ for $p\geq k_{n}^\xi,$
	$$\hat{\epsilon}_p^n \eqdef \frac{1}{\max(1,\|r_{k_n^{\xi}}\|)} \biggl[\phi(\tilde{S}_p)R_{p+1} - h(\theta_p)\biggr] + \gamma \biggl[\phi(\tilde{S}_p)\phi(S_{p+1})^T - \Phi^T D_{\rho,\theta_p}P_{\theta_p} \Phi \biggr]\hat{\bar{\omega}}_p^n + \biggl[\bar{G}(\theta_p) - \phi(\tilde{S}_p) \phi(\tilde{S}_p)^T\biggr]\hat{\omega}_{p}^n \,.$$
  In the following, we control this Markovian noise using the decomposition technique of \cite{ben-met-pri90} which was also used in \cite{konda-tsitsiklis03slowMC}. We use similar notations to those of the proof of \cite[Lem.~8]{konda-tsitsiklis03slowMC}. Define the Markov chain $Y_{p+1} \eqdef (\tilde{S}_p, \tilde{A}_p)$.
	The perturbation $\hat{\epsilon}_p^n$ is of the form
	$$F_{\theta_p}(\hat{\omega}_p^n,\hat{\bar{\omega}}_p^n,Y_{p+1}) - \bar{F}_{\theta_p}(\hat{\omega}_p^n,\hat{\bar{\omega}}_p^n)
  + M_{p+1}^{(1)} \hat{\bar{\omega}}_p^n
  + M_{p+1}^{(2)}\,,$$
	where $\bar{F}_{\theta}(\omega,\bar{\omega})$ is the steady state expectation of $F_\theta(\omega,\bar{\omega},(\bar{S}_p,\bar{A}_p))$, where $\bar{S}_p$ is a Markov chain with transition kernel~$P_\theta$, and where $M_{p+1}^{(i)}$ for~$i = 1,2$ are martingale difference sequences. For every~$\theta \in \bR^d, \omega, \bar{\omega} \in \bR^m,$ there exists a solution~$\hat{F}_{\theta}(\omega,\bar{\omega})$ to the so-called Poisson equation:
	$$ F_\theta(\omega,\bar{\omega},y) - \bar{F}_\theta(\omega,\bar{\omega}) = \hat{F}_\theta(\omega,\bar{\omega},y) - (P_\theta \hat{F}_\theta)(\omega,\bar{\omega},y)\,.$$
	Using this equation, the perturbation can be decomposed as follows for any fixed~$n \in \bN$ and~$p \geq k_n$,
	\begin{align}
	\nonumber\hat{\epsilon}_p^n &= M_{p+1}^{(1)} \hat{\bar{\omega}}_p^n + M_{p+1}^{(2)} + F_{\theta_p}(\hat{\omega}_p^n,\hat{\bar{\omega}}_p^n,Y_{p+1}) - \bar{F}_{\theta_p}(\hat{\omega}_p^n,\hat{\bar{\omega}}_p^n) \\
	\nonumber&= M_{p+1}^{(1)} \hat{\bar{\omega}}_p^n + M_{p+1}^{(2)} + \hat{F}_{\theta_p}(\hat{\omega}_p^n,\hat{\bar{\omega}}_p^n,Y_{p+1}) - (P_{\theta_p}\hat{F}_{\theta_p})(\hat{\omega}_p^n,\hat{\bar{\omega}}_p^n,Y_{p+1}) \\
	&= (M_{p+1}^{(1)}\hat{\bar{\omega}}_p^n + M_{p+1}^{(2)} + (\hat{F}_{\theta_p}(\hat{\omega}_p^n,\hat{\bar{\omega}}_p^n,Y_{p+1})) - (P_{\theta_p}\hat{F}_{\theta_p})(\hat{\omega}_p^n,\hat{\bar{\omega}}_p^n,Y_p))\\
  &\quad+ ((P_{\theta_{p-1}}\hat{F}_{\theta_{p-1}})(\hat{\omega}_{p-1}^n,\hat{\bar{\omega}}_{p-1}^n,Y_{p}) - (P_{\theta_{p}}\hat{F}_{\theta_{p}})(\hat{\omega}_{p}^n,\hat{\bar{\omega}}_{p}^n,Y_{p+1}) )\\
 &\quad + (P_{\theta_p}\hat{F}_{\theta_p})(\hat{\omega}_{p}^n,\hat{\bar{\omega}}_{p}^n,Y_p) - (P_{\theta_p}\hat{F}_{\theta_p})(\hat{\omega}_{p-1}^n,\hat{\bar{\omega}}_{p}^n,Y_p)\\
 &\quad + (P_{\theta_p}\hat{F}_{\theta_p})(\hat{\omega}_{p-1}^n,\hat{\bar{\omega}}_{p}^n,Y_p) - (P_{\theta_p}\hat{F}_{\theta_p})(\hat{\omega}_{p-1}^n,\hat{\bar{\omega}}_{p-1}^n,Y_p) \\
 &\quad + (P_{\theta_p}\hat{F}_{\theta_p})(\hat{\omega}_{p-1}^n,\hat{\bar{\omega}}_{p-1}^n,Y_p) -(P_{\theta_{p-1}}\hat{F}_{\theta_{p-1}})(\hat{\omega}_{p-1}^n,\hat{\bar{\omega}}_{p-1}^n,Y_p)\,.
	\label{eq:noise_mart}
	\end{align}

	Eqs.~\eqref{eq:all_noise},\eqref{eq:bound_G_bar} and \eqref{eq:noise_mart} imply that the proof is complete if we show that:
	$$\lim_{n\to\infty} \max_{k_n^{\xi} \leq k \leq k_{n+1}^{\xi}} \sum_{i=j_n}^{i_k-1}e^{-\frac{1}{2}\epsilon T(i_k -1 -i)} \left\|\sum_{p=k_{i}^\beta}^{k_{i+1}^\beta-1} \bar{G}_{p+1:k_{i+1}^\beta-1} \beta_p \hat{\epsilon}_p^n\right\| = 0\,, \quad a.s.$$
	For this, it is sufficient to prove the following inequality:
	\begin{equation}
	\label{eq:lem_8}\mathbb{E} \left[ \max_{j_n\leq i \leq j_{n+1}-1}\left\|\sum_{p=k_{i}^\beta}^{k_{i+1}^\beta-1} \bar{G}_{p+1:k_{i+1}^\beta-1} \beta_p \hat{\epsilon}_p^n\right\|^2\right] \leq C  \sum_{p = k_n^\xi}^{k_{n+1}^\xi-1} \beta_p^2\,.
	\end{equation}
	Indeed, the Chebyshev inequality implies that for every~$\delta>0$,
	$$\mathbb{P} \left(\max_{j_n\leq i\leq j_{n+1}-1}\left\|\sum_{p=k_{i}^\beta}^{k_{i+1}^\beta-1} \bar{G}_{p+1:k_{i+1}^\beta-1} \beta_p \hat{\epsilon}_p^n\right\| \geq \delta\right) \leq \frac{C}{\delta^2}\sum_{p = k_n^\xi}^{k_{n+1}^\xi-1} \beta_p^2 \,,$$
	and applying the Borel-Cantelli lemma with the summability of the series $\sum_k \beta_k^2$ yields:
	$$ \lim_{n \to \infty} \max_{j_n\leq i\leq j_{n+1}-1}\left\|\sum_{p=k_{i}^\beta}^{k_{i+1}^\beta-1} \bar{G}_{p+1:k_{i+1}^\beta-1} \beta_p \hat{\epsilon}_p^n\right\|= 0\,, \quad a.s.$$

	To prove that Ineq.~\ref{eq:lem_8} holds, it is sufficient to show that the desired inequality holds when $\hat{\epsilon}_p^n$ is replaced by each one
	of the terms of its decomposition. For the first term which is a martingale difference with bounded second moment, we establish the sought-after inequality in Lem.~\ref{tech_lem:mart_diff}. The last three terms are of the order~$O(\beta_p)$, $O(\xi_p)$ and~$O(\alpha_p)$, respectively. The remaining term is the summand of a telescopic series with bounded moment and we address its particular case in Lem.~\ref{tech_lem:telescopic} below.
\end{proof}


\begin{lemma}
	\label{tech_lem:mart_diff}
	There exists $C>0$ s.t. for every~$n \in \bN,$
	$$\mathbb{E}\left[ \max_{j_n\leq i\leq j_{n+1}-1} \left\| \sum_{p = k_{i}^\beta}^{k_{i+1}^\beta-1} \bar{G}_{p+1:k_{i+1}^\beta - 1}\beta_p Z_{p+1}^n \right\|^2\right]\leq C \sum_{p = k_n^\xi}^{k_{n+1}^\xi-1} \beta_p^2\,,$$
  where for every~$p \geq k_n,$
    $
    Z_{p+1}^n \eqdef M_{p+1}^{(1)}\hat{\bar{\omega}}_p^n
    + M_{p+1}^{(2)}+ (\hat{F}_{\theta_p}(\hat{\omega}_p^n,\hat{\bar{\omega}}_p^n,Y_{p+1}) - (P_{\theta_p}\hat{F}_{\theta_p})(\hat{\omega}_p^n,\hat{\bar{\omega}}_p^n,Y_p))\,.
    $
\end{lemma}

\begin{proof}
  In this proof, we suppress the superscript~$n$ of~$Z_{p+1}^n$ to simplify notation. Note that~$n$ is fixed throughout the proof.
	Define~$M_{k_i^\beta}^k := \sum_{l = k_i^\beta}^{k}\beta_lZ_{l+1}$ for every~$i \in \bN, k > k_i^{\beta}$. This is a zero mean, square integrable martingale for $k\in\{k_n^\xi+1,\cdots,k_{n+1}^\xi\}$. By summation by part, we have for every $j_n \leq i \leq j_{n+1}-1$,
	$$\sum_{p=k_{i}^\beta}^{k_{i+1}^\beta-1}\bar{G}_{p+1:k_{i+1}^\beta-1}\beta_p Z_p = M_{k_{i}^\beta}^{k_{i+1}^\beta-1} -  \sum_{p = k_{i}^\beta}^{k_{i+1}^\beta-2}(\bar{G}_{p+1:k_{i+1}^\beta-1} - \bar{G}_{p:k_{i+1}^\beta-1})M_{k_{i}^\beta}^p\,.$$
	Notice that $\bar{G}_{p+1:k_{i+1}^\beta-1} - \bar{G}_{p:k_{i+1}^\beta-1} = \beta_p  \bar{G}_{p+1:k_{i+1}^\beta-1}\bar{G}(\theta_p)$. Hence, bounding the max by the sum, we obtain the following inequality:
	\begin{align}
	\max_{j_n\leq i \leq j_{n+1}-1} \left\|\sum_{p=k_{i}^\beta}^{k_{i+1}^\beta-1}\bar{G}_{p+1:k_{i+1}^\beta-1}\beta_p Z_{p+1}\right\|^2
	&\leq 2\sum_{i=j_n}^{j_{n+1}-1}\left\|M_{k_{i}^\beta}^{k_{i+1}^\beta-1}\right\|^2 + 2 \sum_{i=j_n}^{j_{n+1}-1} \left\|\sum_{p = k_i^\beta}^{k_{i+1}^\beta-1} \beta_p \bar{G}_{p+1:k_{i+1}^\beta-1}\bar{G}(\theta_p)M_{k_{i}^\beta}^p\right\|^2
	\label{eq:sum_mart_square}
	\end{align}
	We have that~$\sup_{\theta \in \bR^d} \|\bar{G}(\theta)\|< \infty$. Moreover, using Lem.~\ref{lem:unif-pos-def}\,, one can show that there exists~$C>0$ s.t. for every integers $q>p$, $\|\bar{G}_{p:q}\| \leq C$.
	Thus, we obtain the following upper bound using the triangle inequality:
	\begin{equation*}
	\left\|\sum_{p = k_i^\beta}^{k_{i+1}^\beta-1} \beta_p \bar{G}_{p+1:k_{i+1}^\beta-1}\bar{G}(\theta_p)M_{k_{i}^\beta}^p\right\|^2
	\leq C^2\left(\sum_{p=k_i^\beta}^{k_{i+1}^\beta-1} \beta_p \left\|M_{k_i^\beta}^{p}\right\|\right)^2
	\leq C^2 T'^2 \left(\max_{k_i^\beta \leq p \leq k_{i+1}^\beta-1} \|M_{k_i^\beta}^p\|\right)^2\,.
\end{equation*}
	Taking the expectation in Eq.~\eqref{eq:sum_mart_square}  and using Doob's inequality yields:
	\begin{align*}\mathbb{E}\left[\max_{j_n\leq i \leq j_{n+1}-1} \left\|\sum_{p=k_{i}^\beta}^{k_{i+1}^\beta-1}\bar{G}_{p+1:k_{i+1}^\beta-1}\beta_p Z_{p+1} \right\|^2\right]
	&\leq \left(2 + 8C^2T'^2\right) \sum_{i = j_n}^{j_{n+1}-1} \mathbb{E}\left[\left\|M_{k_{i}^\beta}^{k_{i+1}^\beta-1}\right\|^2\right] \\
	& \leq \left(2 + 8C^2T'^2\right) C_Z\sum_{i=j_n}^{j_{n+1}-1} \sum_{p = k_i^\beta}^{k_{i+1}^\beta-1} \beta_p^2 \\
	&= \left(2 + 8C^2T'^2\right) C_Z\sum_{p = k_n^\xi}^{k_{n+1}^\xi-1} \beta_p^2\,,
	\end{align*}
	where the last inequality comes from the bounded second moment of $Z_{p+1}$.
\end{proof}

\begin{lemma}
	\label{tech_lem:telescopic}
	Let~$(X_k)$ be an $\mathbb{R}^m$-valued random sequence with bounded second moment. Then, there exists~$C>0$ s.t.:
	$$\mathbb{E}\left[\max_{j_n\leq i \leq j_{n+1}-1}\left\|\sum_{p=k_{i}^\beta}^{k_{i+1}^\beta-1} \bar{G}_{p+1:k_{i+1}^\beta-1} \beta_p (X_{p+1} - X_p)\right\|^2\right] \leq C \sum_{p = k_{n}^\xi}^{k_{n+1}^\xi-1} \beta_p^2\,.$$
\end{lemma}
\begin{proof}
	Summation by parts yields for~$j_n\leq i \leq j_{n+1}-1$,
	\begin{multline}
    \sum_{p=k_{i}^\beta}^{k_{i+1}^\beta-1} \bar{G}_{p+1:k_{i+1}^\beta-1} \beta_p(X_{p+1}-X_p) = \beta_{k_{i+1}^\beta-1} X_{k_{i+1}^\beta} - \beta_{k_{i}^\beta} \bar{G}_{k_{i}^\beta+1:k_{i+1}^\beta-1}X_{k_{i}^\beta}\\
     + \sum_{p=k_{i}^\beta+1}^{k_{i+1}^\beta-1}(\beta_p\bar{G}_{p+1:k_{i+1}^\beta-1} - \beta_{p-1}\bar{G}_{p:k_{i+1}^\beta-1})X_p\,.
	\label{eq:sum_G_bar_tel}
\end{multline}
	Notice that $\beta_p\bar{G}_{p+1:k_{i+1}^\beta-1} - \beta_{p-1}\bar{G}_{p:k_{i+1}^\beta-1}=(\beta_{p}-\beta_{p-1})\bar{G}_{p+1:k_{i+1}^\beta-1} + \beta_{p-1}\beta_p \bar{G}_{p+1:k_{i+1}^\beta-1}\bar{G}(\theta_p)$. Then, similarly to the proof of the previous lemma, recall that~$\sup_{\theta \in \bR^d} \|\bar{G}(\theta)\|< \infty$ and that Lem.~\ref{lem:unif-pos-def} entails the existence of a constant~$C_G>0$ s.t. for sufficiently large~$p$ and for every integers~$q>p$, $\max(\|\bar{G}(\theta_p)\bar{G}_{p+1:q}\|,\|\bar{G}_{p:q}\|) \leq C_G$.
  Using the previous remarks with Eq.~\eqref{eq:sum_G_bar_tel} yields for $j_n\leq i \leq j_{n+1}-1,$
	\begin{multline}
    \label{eq:ineq-four-terms}
	\left\|\sum_{p=k_{i}^\beta}^{k_{i+1}^\beta-1} \bar{G}_{p+1:k_{i+1}^\beta-1} \beta_p (X_{p+1} - X_{p})\right\|^2
	\leq 4\beta_{k_{i+1}^\beta-1}^2 \left\|X_{k_{i+1}^\beta}\right\|^2 + 4C_G^2\beta_{k_{i}^\beta}^2 \left\|X_{k_{i}^\beta}\right\|^2\\
	+4\left\|\sum_{p=k_{i}^\beta+1}^{k_{i+1}^\beta-1}(\beta_{p}-\beta_{p-1})\bar{G}_{p+1:k_{i+1}^\beta-1}X_p\right\|^2
  + 4\left\|\sum_{p=k_{i}^\beta+1}^{k_{i+1}^\beta-1} \beta_{p-1}\beta_p \bar{G}_{p+1:k_{i+1}^\beta-1}\bar{G}(\theta_p)X_p\right\|^2\,.
\end{multline}
	To prove the lemma, it is sufficient to show that the desired inequality holds when the l.h.s. is replaced by each of the terms on the r.h.s. of the above equation. Consider the first term:
	\begin{equation}
	\mathbb{E}\left[ \max_{j_n\leq i \leq j_{n+1}-1} \beta_{k_{i+1}^\beta-1}^2 \|X_{k_{i+1}^\beta}\|^2\right]
  \leq \sum_{i = j_n}^{j_{n+1}-1} \beta_{k_{i+1}^\beta-1}^2\mathbb{E}\left[ \|X_{k_{i+1}^\beta}\|^2\right]
	\leq C_X \sum_{i = j_n}^{j_{n+1}-1} \beta_{k_{i+1}^\beta-1}^2 \leq C \sum_{p = k_n^\xi}^{k_{n+1}^\xi-1}\beta_p^2\,,
	\label{eq:max_beta_x}
  \end{equation}
	where the constant $C_X>0$ bounds the second moment of $X_k$ (i.e., $\sup_k \mathbb{E} \|X_k\|^2 \leq C_X$) and~$C$ is also a positive constant independent of~$p$ and~$n$. The second term is treated analogously.\\
	Let us consider now the third term. Using the triangle inequality combined with the boundedness of~$\|\bar{G}_{p:q}\|$ for~$q>p$ yields for~$n$ sufficiently large and~$j_n\leq i \leq j_{n+1}-1,$
	\[
		\left\|\sum_{p = k_i^\beta+1}^{k_{i+1}^\beta-1} (\beta_{p} - \beta_{p-1})\bar{G}_{p+1:k_{i+1}^\beta-1}X_p\right\|^2
    \leq  C_G^2 \left(\sum_{p=k_{i}^\beta+1}^{k_{i+1}^\beta-1} |\beta_{p-1}-\beta_{p}| \cdot \|X_p\| \right)^2\,.
	\]
	Then, it follows that:
	\[
	\max_{j_n \leq i \leq j_{n+1}-1} \left\|\sum_{p = k_i^\beta+1}^{k_{i+1}^\beta-1} (\beta_{p} - \beta_{p-1})\bar{G}_{p+1:k_{i+1}^\beta-1}X_p\right\|^2
	\leq C_G^2 \sum_{i= j_n}^{j_{n+1}-1} \left(\sum_{p=k_{i}^\beta+1}^{k_{i+1}^\beta-1} |\beta_{p-1}-\beta_{p}| \cdot \|X_p\| \right)^2\,.
	\]
	We obtain the desired inequality by taking the expectation and using the boundedness of the second moment of the r.v.~$X_k$:
	\begin{equation*}
	\mathbb{E}\left[ \max_{j_n \leq i \leq j_{n+1}-1} \left\|\sum_{p = k_i^\beta+1}^{k_{i+1}^\beta-1} (\beta_{p-1} - \beta_{p})\bar{G}_{p+1:k_{i+1}^\beta-1}X_p\right\|^2 \right]
	\leq C' \sum_{i = j_n}^{j_{n+1}-1} \left(\sum_{p = k_i^\beta+1}^{k_{i+1}^\beta-1} (\beta_{p-1} - \beta_p) \right)^2
	\leq C' \sum_{p = k_n^\xi}^{k_{n+1}^\xi-1} \beta_p^2\,,
\end{equation*}
	where $C' \eqdef C_G^2 C_X$.
	It only remains to show that the desired inequality also holds for the fourth term in Ineq.~\eqref{eq:ineq-four-terms}. Using similar manipulations as above, we have:
	$$\max_{j_n\leq i \leq j_{n+1}}\left\|\sum_{p=k_{i}^\beta+1}^{k_{i+1}^\beta-1} \beta_{p-1}\beta_p \bar{G}_{p+1:k_{i+1}^\beta-1}\bar{G}(\theta_p)X_p\right\|^2 \leq
	C_G^2\sum_{i = j_n}^{j_{n+1}-1}\sum_{p=k_{i}^\beta+1}^{k_{i+1}^\beta-1}\sum_{q=k_{i}^\beta+1}^{k_{i+1}^\beta-1} \beta_p\beta_{p-1}\beta_q\beta_{q-1} \|X_p\| \|X_q\|\,,$$
	and taking the expectation implies:
	\begin{equation*}
	\mathbb{E}\left[\max_{j_n\leq i \leq j_{n+1}}\left\|\sum_{p=k_{i}^\beta+1}^{k_{i+1}^\beta-1} \beta_{p-1}\beta_p \bar{G}_{p+1:k_{i+1}^\beta-1}\bar{G}(\theta_p)X_p\right\|^2\right]
  \leq C' \left(\sum_{p = k_n^\xi}^{k_{n+1}^\xi-1} \beta_p^2 \right)^2
	\leq \tilde{C} \sum_{p = k_n^\xi}^{k_{n+1}^\xi-1} \beta_p^2\,,
\end{equation*}
	where $\tilde{C} := C' \sum_{k=0}^\infty \beta_k^2$. Thus, the lemma holds for $C\geq 4C_X + 4C_X C_G^2 + 4C' + 4\tilde{C}$.
\end{proof}

\begin{lemma}
	\label{tech_lem:bar_omega_rec}
$\lim_{n\to\infty}\,\sum_{k = k_n}^{k_{n+1}}\xi_k \biggl(e^{-\frac{1}{2}\epsilon \sum_{l=k_n}^{k}\beta_l} + \sum_{l=k_n}^{k} e^{-\frac{1}{2}\epsilon\sum_{m = l+1}^{k}\beta_m}\xi_l\biggr) = 0\,.$
\end{lemma}

\begin{proof}
	We have already proved that $\lim_{n \to \infty} \sum_{k = k_n}^{k_{n+1}}\xi_k e^{-\frac{1}{2}\epsilon \sum_{l=k_n}^{k}\beta_l} = 0$ (see the proof of Lem.~\ref{tec_lem:cvg_x}).
	The convergence of the second term in the lemma is proven in the same manner.
\end{proof}

\end{document}